\documentclass[9pt]{extarticle}

\usepackage{microtype}
\usepackage{graphicx}
\usepackage{subfigure}
\usepackage{booktabs} 

\usepackage{comment}

\usepackage{notations}
\usepackage{amsmath}
\usepackage{amssymb}
\usepackage{bbm}
\usepackage{amsthm}
\usepackage{bm}
\usepackage[dvipsnames]{xcolor}

\usepackage{soul}

\makeatletter
\newenvironment{sqcases}{%
  \matrix@check\sqcases\env@sqcases
}{%
  \endarray\right.%
}
\def\env@sqcases{%
  \let\@ifnextchar\new@ifnextchar
  \left\lbrack
  \def\arraystretch{1.2}%
  \array{@{}l@{\quad}l@{}}%
}
\makeatother

\newenvironment{talign}
 {\align}
 {\endalign}
\newenvironment{talign*}
 {\csname align*\endcsname}
 {\endalign}

\usepackage{hyperref}

\usepackage[accepted]{arxiv2025_2}

\usepackage{amsmath}
\usepackage{amssymb}
\usepackage{mathtools}
\usepackage{amsthm}

\usepackage[capitalize,noabbrev]{cleveref}

\theoremstyle{plain}
\newtheorem{theorem}{Theorem}[section]

\newtheorem{result}[theorem]{Result}

\theoremstyle{definition}

\theoremstyle{remark}
\newtheorem{remark}[theorem]{Remark}

\usepackage[textsize=tiny]{todonotes}

\arxivtitlerunning{Statistical mechanics of extensive-width Bayesian neural networks near interpolation}

\begin{document}

\twocolumn[
\arxivtitle{Statistical mechanics of extensive-width Bayesian neural networks near interpolation}

\arxivsetsymbol{equal}{*}

\begin{arxivauthorlist}
\arxivauthor{Jean Barbier}{equal,ictp}
\arxivauthor{Francesco Camilli}{equal,ictp}
\arxivauthor{Minh-Toan Nguyen}{equal,ictp}
\arxivauthor{Mauro Pastore}{equal,ictp}
\arxivauthor{Rudy Skerk}{equal,sissa}
\end{arxivauthorlist}

\arxivaffiliation{ictp}{The Abdus Salam International Centre for Theoretical Physics (ICTP), Strada Costiera 11, 34151 Trieste, Italy}
\arxivaffiliation{sissa}{International School for Advanced Studies (SISSA), Via Bonomea 265, 34136 Trieste, Italy}

\arxivcorrespondingauthor{Mauro Pastore}{mpastore@ictp.it}

\arxivkeywords{Bayesian neural networks | Optimal generalisation performance | Statistical physics | Replica method | Random matrix theory  | Learning phase transitions | Statistical-to-computational gap}

\vskip 0.3in
]



\printAffiliationsAndNotice{\arxivEqualContribution} 

\begin{abstract}
For three decades statistical mechanics has been providing a framework to analyse neural networks. However, the theoretically tractable models, e.g., perceptrons, random features models and kernel machines, or multi-index models and committee machines with few neurons, remained simple compared to those used in applications. In this paper we help reducing the gap between practical networks and their theoretical understanding through a statistical physics analysis of the supervised learning of a two-layer fully connected network with generic weight distribution and activation function, whose hidden layer is large but remains proportional to the inputs dimension. This makes it more realistic than infinitely wide networks where no feature learning occurs, but also more expressive than narrow ones or with fixed inner weights. We focus on the Bayes-optimal learning in the teacher-student scenario, i.e., with a dataset generated by another network with the same architecture. We operate around interpolation, where the number of trainable parameters and of data are comparable and feature learning emerges. Our analysis uncovers a rich phenomenology with various learning transitions as the number of data increases. In particular, the more strongly the features (i.e., hidden neurons of the target) contribute to the observed responses, the less data is needed to learn them.
Moreover, when the data is scarce, the model only learns non-linear combinations of the teacher weights, rather than ``specialising'' by aligning its weights with the teacher's.
Specialisation occurs only when enough data becomes available, but it can be hard to find for practical training algorithms, possibly due to statistical-to-computational~gaps.
\end{abstract}

\section{Introduction}

Understanding the expressive power and generalisation capabilities of neural networks is not only a stimulating intellectual activity, producing surprising results that seem to defy established common sense in statistics and optimisation~\citep{bartlett2021}, but has important practical implications in cost-benefit planning whenever a model is deployed.
E.g., from a fruitful research line that spanned three decades, we now know that deep fully connected Bayesian neural networks with $O(1)$ readout weights and $L_2$ regularisation behave as kernel machines (the so-called Neural Network Gaussian processes, NNGPs) in the heavily overparametrised, infinite-width regime~\citep{neal1996,williams1996,lee2018gaussian,matthews2018gaussian,hanin2023infinite}, and so suffer from these models' limitations. Indeed, kernel machines infer the decision rule by first embedding the data in a fixed a priori feature space, the renowned \emph{kernel trick}, then operating linear regression/classification over the features. In this respect, they do not learn features (in the sense of statistics relevant for the decision rule) from the data, so they need larger and larger feature spaces and training sets to fit their higher order statistics~\citep{yoon1998poly,dietrich1999svm,gerace2021,bordelon2021kernel,canatar2021spectral,xiao2022precise}.

Many efforts have been devoted to studying Bayesian neural networks beyond this regime. In the so-called proportional regime, when the width is large and proportional to the training set size, recent studies showed how a limited amount of feature learning makes the network equivalent to optimally regularised kernels~\citep{li2021,pacelli2023,camilli2023fundamental,cui2023bayes,baglioni2024,camilli2025inforeduction}. This could be a consequence of the fully connected architecture, as, e.g., convolutional neural networks learn more informative features~\citep{naveh2021,seroussi2023,aiudi2023,bassetti2024}. 
Another scenario is the mean-field scaling, i.e., 
when the readout weights are small:
in this case too a Bayesian network can learn features in the proportional regime~\citep{rubin2024unified,vanmeegen2024}.

Here instead we analyse a fully connected two-layer Bayesian network trained end-to-end \emph{near the interpolation threshold}, when the sample size $n$ is scaling like the number of trainable parameters: for input dimension $d$ and width $k$, both large and proportional, $n= \Theta(d^2) =\Theta(kd)$, a regime where non-trivial feature learning can happen. We consider i.i.d. Gaussian input vectors with labels generated by a teacher network with matching architecture, in order to study the Bayes-optimal learning 
of this neural network target function. Our results thus provide a benchmark for the performance of \emph{any} model trained on the same dataset.

\section{Setting and main results}

\subsection{Teacher-student setting} 

We consider supervised learning with a shallow neural network in the classical teacher-student setup~\citep{gardner1989teacher}. The data-generating model, i.e., the teacher (or target function), is thus a two-layer neural network itself, with readout weights $\bv^0\in\R^k$ and internal weights $\bW^0\in\R^{k\times d}$, drawn entrywise i.i.d. from $P_v^0$ and $P^0_W$, respectively; we assume $P^0_W$ to be centred while $P^0_v$ has mean $\bar{v}$, and both priors have unit second moment. We denote the whole set of parameters of the target as $\btheta^0=(\bv^0,\bW^0)$. The inputs are i.i.d. standard Gaussian vectors $\bx_\mu\in\R^{d}$ for $\mu\le n$. The responses/labels $y_\mu$ are drawn from a kernel $P^0_{\rm out}$:
\begin{talign} \label{eq:teacher}
    y_\mu\sim P^0_{\rm out}(\,\cdot\mid \lambda^0_\mu), \quad \lambda^0_\mu:=\frac1{\sqrt k}{\bv^{0\intercal}}\sigma(\frac1{\sqrt d}{\bW^0\bx_\mu}).
\end{talign}
The kernel can be stochastic or model a deterministic rule if $P^0_{\rm out}(y\mid\lambda)=\delta(y-\mathsf{f}^0(\lambda))$ for some outer non-linearity $\mathsf{f}^0$. The activation function $\sigma$ is applied entrywise to vectors and is required to admit an expansion in Hermite polynomials with Hermite coefficients  $(\mu_\ell)_{\ell\geq0}$, see App.~\ref{app:hermite}: $ \sigma(x) = \sum_{\ell \ge 0 } \frac{\mu_\ell}{\ell !}\He_\ell (x)$. We assume it has vanishing 0th Hermite coefficient, i.e., that it is centred $\EE_{z\sim\mathcal{N}(0,1)}\sigma(z)=0$; in App.~\ref{app:non-centered} we relax this assumption.
The input/output pairs $\mathcal{D}=\{(\bx_\mu,y_\mu)\}_{\mu \leq n}$ form the training set for a student network with matching architecture. 

Notice that the readouts $\bv^0$ are only $k$ unknowns in the target compared to the $kd=\Theta(k^2)$ inner weights $\bW^0$. Therefore, they can be equivalently considered quenched, i.e., either given and thus fixed in the student network defined below, or unknown and thus learnable, without changing the leading order of the information-theoretic quantities we aim for. E.g., in terms of mutual information per parameter $\frac1{kd+k}I((\bW^0,\bv^0);\mathcal{D})=\frac1{kd}I(\bW^0;\mathcal{D}\mid \bv^0)+o_d(1)$.
Without loss of generality, we thus consider $\bv^0$ quenched and denote it $\bv$ from now on. This equivalence holds at leading order and at equilibrium only, but not at the dynamical level, the study of which is left for future work.

The Bayesian student learns via the posterior distribution of the weights $\bW$ given the training data (and $\bv$), defined by
\begin{talign*}
&dP(\bW \mid\mathcal{D}):=\mathcal{Z}(\mathcal{D})^{-1}dP_W(\bW)\prod_{\mu\le n}P_{\rm out}\big(y_\mu\mid \lambda_\mu(\bW)\big) 
\end{talign*}
with post-activation $\lambda_\mu(\bW):=\frac1{\sqrt k}\bv^{\intercal}\sigma(\frac1{\sqrt d}{\bW\bx_\mu})$, the posterior normalisation constant $\mathcal{Z}(\mathcal{D})$ called the partition function, and $P_W$ is the prior assumed by the student. From now on, we focus on the Bayes-optimal case $P_W=P_W^0$ and $P_{\rm out}=P_{\rm out}^0$, but the approach can be extended to account for a mismatch.

We aim at evaluating the expected generalisation error of the student. Let $(\bx_{\rm test}, y_{\rm test}\sim P_{\rm out}(\,\cdot\mid \lambda^0_{\rm test}))$ be a fresh sample ({not present in $\mathcal{D}$}) drawn using the teacher, where $\lambda_{\rm test}^0$ is defined as in \eqref{eq:teacher} with $\bx_\mu$ replaced by $\bx_{\rm test}$ (and similarly for $\lambda_{\rm test}(\bW)$). Given any prediction function $\mathsf{f}$, the Bayes estimator for the test response reads $\hat{y}^{\mathsf{f}}(\bx_{\rm test},\calD)
    :=\langle \mathsf{f}(\lambda_{\rm test}(\bW)) \rangle$, where the expectation $\langle \,\cdot\, \rangle :=\EE[ \,\cdot \mid \mathcal{D}]$ is w.r.t. the posterior $dP(\bW \mid\mathcal{D})$. Then, for a
performance measure $\mathcal{C}:\mathbb{R}\times \mathbb{R}\mapsto\mathbb{R}_{\ge 0}$ the Bayes generalisation error is
\begin{align}
\varepsilon^{\mathcal{C},\mathsf{f}}:=\EE_{\btheta^0,\calD,\bx_{\rm test},y_{\rm test}}\mathcal{C}\big(y_{\rm test}, \big\langle \mathsf{f}(\lambda_{\rm test}(\bW)) \big\rangle\big).\label{eq:Bayes_error_def}
\end{align}
An important case is the square loss $\mathcal{C}(y,\hat y)=(y-\hat y)^2$ with the choice $\mathsf{f}(\lambda)=\int dy\, y\, P_{\rm out}( y\mid \lambda)=:\EE[y\mid \lambda]$. 
The Bayes-optimal mean-square generalisation error follows:
\begin{align}\label{eq:gen_error_def}
    \varepsilon^{\rm opt} &:= \EE_{\btheta^0,\calD,\bx_{\rm test},y_{\rm test}}\big(y_{\rm test} - \big\langle\EE[y\mid \lambda_{\rm test}(\bW)]\big\rangle\big)^2.
\end{align}
Our main example will be the case of \emph{linear readout} with Gaussian label noise: $P_{\rm out}(y\mid\lambda) =\exp(-\frac1{2\Delta}(y-\lambda)^2)/\sqrt{2\pi\Delta}$.
In this case, the generalisation error $\varepsilon^{\rm opt}$ takes a simpler form for numerical evaluation than \eqref{eq:gen_error_def}, thanks to the concentration of ``overlaps'' entering it, see App.~\ref{app:gen_err}.

We study the challenging \emph{extensive-width regime with quadratically many samples}, i.e., a large size limit 
\begin{align}
 d,k,n\to+\infty \quad \text{with} \quad  k/d\to\gamma, \quad n/d^2\to\alpha  .\label{thermolim}
\end{align}
We denote this joint $d,k,n$ limit with these rates by ``${\lim}$''.

In order to access $\varepsilon^{\mathcal{C},\mathsf{f}},\varepsilon^{\rm opt}$ and other relevant quantities, one can tackle the computation of the average log-partition function, or free entropy in statistical physics language: 
\begin{talign}
f_n:=\frac1n\EE_{\btheta^0,\mathcal{D}}\ln\mathcal{Z}(\mathcal{D}).
\label{eq:free_entropy_def}
\end{talign}
The mutual information between teacher weights and the data is related to the free entropy $f_n$, see App.~\ref{app:mutual_info}. E.g., in the case of linear readout with Gaussian label noise we have $\lim \frac1{kd}I(\bW^0;\mathcal{D}\mid \bv) = -\frac{\alpha}{\gamma} \lim f_n - \frac{\alpha}{2\gamma} \ln(2\pi e \Delta)$. Considering the mutual information per parameter allows us to interpret $\alpha$ as a sort of signal-to-noise ratio, so that the mutual information defined in this way increases with it.

\vspace{3pt}

\emph{Notations:} Bold is for vectors and matrices; $d$ is the input dimension, $k$ the width of the hidden layer, $n$ the size of the training set $\mathcal{D}$, with asymptotic ratios given by \eqref{thermolim}; $\bA^{\circ \ell}$ is the Hadamard power of a matrix; for a vector $\bv$, $(\bv)$ is the diagonal matrix $\diag(\bv)$;  $(\mu_\ell)$ are the Hermite coefficients of the activation function $\sigma(x) = \sum_{\ell \ge 0} \frac{\mu_\ell}{\ell !}\He_\ell (x)$; the norm $\|\,\cdot\,\|$ for vectors and matrices is the Frobenius norm.

\subsection{Main results}\label{sec:result}

The aforementioned setting is related to the recent paper \cite{maillard2024bayes}, with two major differences: said work considers Gaussian distributed weights and quadratic activation. These hypotheses allow numerous simplifications in the analysis, exploited in a series of works \cite{du2018power,soltanolkotabi2018theoretical,venturi2019spurious,sarao2020optimization,gamarnik2024stationary,martin2024impact,arjevani2025geometry}. Thanks to this, \cite{maillard2024bayes} maps the learning task onto a generalised \emph{linear} model (GLM) where the goal is to infer a Wishart matrix from linear observations, which is analysable using known results on the GLM \cite{barbier2019glm} and matrix denoising \cite{barbier2022statistical,maillard2022perturbative,matrix_inference_Barbier,semerjian2024}.

Our main contribution is a statistical mechanics framework for characterising the prediction performance of shallow Bayesian neural networks, able to handle arbitrary activation functions and different distributions of i.i.d. weights, both ingredients playing an important role for the phenomenology. 

The theory we derive draws a rich picture with various learning transitions when tuning the sample rate $\alpha\approx  n/d^2$. For low $\alpha$, feature learning occurs because the student tunes its weights to match non-linear combinations of the teacher's, rather than aligning to those weights themselves. This phase is \emph{universal} in the (centred, with unit variance) law of the i.i.d. teacher inner weights: our numerics obtained both with binary and Gaussian inner weights match well the theory, which does not depend on this prior here. When increasing $\alpha$, strong feature learning emerges through \emph{specialisation phase transitions}, where the student aligns some of its weights with the actual teacher's ones. In particular, when the readouts $\bv$ in the target function have a non-trivial distribution, a whole sequence of specialisation  transitions occurs as $\alpha$ grows, for the following intuitive reason. Different features in the data are related to the weights of the teacher neurons, $(\bW^0_j\in \mathbb{R}^d)_{j\le k}$. The strength with which the responses $(y_\mu)$ depend on the feature $\bW_j^0$ is tuned by the corresponding readout through $|v_j|$, which plays the role of a feature-dependent ``signal-to-noise ratio''. Therefore, features/hidden neurons $j\in[k]$ corresponding to the largest readout amplitude $\max\{|v_j|\}$ are learnt first by the student when increasing $\alpha$ (in the sense that the teacher-student overlap $\bW^\intercal_j\bW^0_j/d>o_d(1)$), then features with the second largest amplitude are, and so on. If the readouts are continuous, an infinite sequence of specialisation transitions emerges in the limit \eqref{thermolim}. On the contrary, if the readouts are homogeneous (i.e. take a unique value), then a single transition occurs where almost all neurons of the student 
specialise jointly (possibly up to a vanishing fraction). We predict specialisation transitions to occur for binary inner weights and generic activation, or for Gaussian ones and more-than-quadratic activation. We provide a theoretical description of these learning transitions and identify the order parameters (sufficient statistics) needed to deduce the generalisation error through scalar equations.

The picture that emerges is connected to recent findings in the context of extensive-rank matrix denoising \cite{barbier2024phase}. In that model, a recovery transition was also identified, separating a 
universal phase (i.e., independent of the signal prior), from a factorisation phase akin to specialisation in the present context. We believe that this picture and the one found in the present paper are not just similar, but a manifestation of the same fundamental mechanism inherent to the extensive-rank of the matrices involved. Indeed, matrix denoising and neural networks share features with both matrix models \cite{kazakov2000solvable,brezin2016random,anninos2020notes} and planted mean-field spin glasses \cite{nishimori2001statistical,zdeborova2016statistical}. This mixed nature requires blending techniques from both fields to tackle them.
Consequently, the approach developed in Sec.~\ref{sec:theory} based on the replica method \cite{mezard1987spin} is non-standard, as it crucially relies on the Harish Chandra--Itzykson--Zuber (HCIZ), or ``spherical'', integral used in matrix models \cite{itzykson1980planar,matytsin1994large,guionnet2002large}. Mixing spherical integration and the replica method has been previously attempted in \cite{thesis_schmidt,barbier2022statistical} for matrix denoising, both papers yielding promising but quantitatively inaccurate or non-computable results. Another attempt to exploit a mean-field technique for matrix denoising (in that case a high-temperature expansion) is \cite{maillard2022perturbative}, which suffers from similar limitations. The more quantitative answer from \cite{barbier2024phase} was made possible precisely thanks to the understanding that the problem behaves more as a matrix model or as a planted mean-field spin glass depending on the phase in which it lives. The two phases could then be treated separately and then joined using an appropriate criterion to locate the transition. 

It would be desirable to derive a unified theory able to describe the whole phase diagram based on a single formalism. This is what the present paper provides through a principled combination of spherical integration and the replica method, yielding predictive formulas that are easy to evaluate. It is important to notice that the presence of the HCIZ integral, which is a high-dimensional matrix integral, in the replica formula presented in Result~\ref{res:free_entropy} suggests that effective one-body problems are not enough to capture alone the physics of the problem, as it is usually the case in standard mean-field inference and spin glass models. Indeed, the appearance of effective one-body problems to describe complex statistical models is usually related to the asymptotic decoupling of the finite marginals of the variables in the problem at hand in terms of products of the single-variable marginals. Therefore, we do \emph{not} expect a standard cavity (or leave-one-out) approach based on single-variable extraction to be exact, while it is usually showed that the replica and cavity approaches are equivalent in mean-field models \cite{mezard1987spin}. This may explain why the approximate message-passing algorithms proposed in \cite{parker2014bilinear,krzakala2013phase,kabashima2016phase} are, as stated by the authors, not properly converging nor able to match their corresponding theoretical predictions based on the cavity method. Algorithms for extensive-rank systems should therefore combine ingredients from matrix denoising and standard message-passing, reflecting their hybrid mean-field/matrix model nature. 

In order to face this, we adapt the GAMP-RIE (generalised approximate message-passing with rotational invariant estimator) introduced in \cite{maillard2024bayes} for the special case of quadratic activation, to accommodate a generic activation function $\sigma$. By construction, the resulting algorithm described in App.~\ref{app:GAMP} \emph{cannot} find the \emph{specialisation solution}, i.e., a solution where at least $\Theta(k)$ neurons align with the teacher's. Nevertheless, it matches the  performance associated with the so-called \emph{universal solution/branch} of our theory for all $\alpha$, which describes a solution with overlap $\bW^\intercal_j\bW^0_j/d>o_d(1)$ for at most $o(k)$ neurons. As a side investigation, we show empirically that the specialisation solution is potentially hard to reach with popular algorithms for some target functions: the algorithms we tested either fail to find it and instead get stuck in a sub-optimal glassy phase (Metropolis-Hastings sampling for the case of binary inner weights), or may find it but in a training time increasing exponentially with $d$ (ADAM~\cite{ADAM} and Hamiltonian Monte Carlo (HMC) for the case of Gaussian weights). It would thus be interesting to settle whether GAMP-RIE has the best prediction performance achievable by a polynomial-time learner when $n=\Theta(d^2)$ for such targets. For specific choices of the distribution of the readout weights, the evidence of hardness is not conclusive and requires further investigation.

\paragraph*{Replica free entropy}

Our first result is a tractable approximation for the free entropy. To state it, let us introduce two functions $\mathcal{Q}_W(\mathsf{v}),\hat{\mathcal{Q}}_W(\mathsf{v})\in[0,1]$ for $\mathsf{v}\in {\rm Supp}(P_v)$, which are non-decreasing in $|\mathsf{v}|$. Let (see \eqref{eq:g_func} in appendix for a more explicit expression of $g$) 
\begin{talign*}
        &g(x) := \sum_{\ell \ge  3} x^\ell{\mu_{\ell}^2}/{\ell !},\nonumber \\
        &q_K(x,\mathcal{Q}_W) := \mu_1^2 + {\mu_2^2} \,x/2 + \EE_{v\sim P_v}[v^2 g(\mathcal{Q}_W(v))],\\
        &r_K := \mu_1^2+ {\mu_2^2}(1 + \gamma \bar{v}^2)/2 + g(1),\nonumber
\end{talign*} 
   and the auxiliary potentials
    \begin{talign*}
        &\psi_{P_W}(x) := \mathbb{E}_{w^0,\xi} \ln\mathbb{E}_w \exp(-\frac{1}{2}xw^2 + x w^0 w + \sqrt{x} \xi w),\\
        &\psi_{P_{\rm out}}(x;r) := \int dy\, \mathbb{E}_{\xi,u^0}P_{\rm out}(y\mid \xi\sqrt{x}+u^0\sqrt{r-x})\\
        &\qquad\qquad\qquad \qquad\times\ln \mathbb{E}_{u}P_{\rm out}(y\mid \xi\sqrt{x}+u\sqrt{r-x}),\\
        &\iota(x):=\frac{1}{8}+\frac{1}{2} \int \ln|t-s|\,d\mu_{\bY(x)}(t)d\mu_{\bY(x)}(s),
    \end{talign*}
    where $w^0,w \sim P_W$ and $\xi,u_0,u \sim \calN(0,1)$ all independent. Moreover, $\mu_{\bY(x)}$ is the limiting (in $d\to\infty$) spectral density of data $\bY(x)=\sqrt{x/(kd)}\,\bS^0+\bZ$ in the denoising problem of the matrix $\bS^0:=\bW^{0\intercal}(\bv)\bW^0\in\mathbb{R}^{d\times d}$, with $\bZ$ a standard GOE matrix (a symmetric matrix whose upper triangular part has i.i.d. entries from $\mathcal{N}(0,(1+\delta_{ij})/d)$). Denote the minimum mean-square error associated with this denoising problem as ${\rm mmse}_S(x)=\lim_{d\to \infty}d^{-2}\EE\|\bS^0-\EE[\bS^0\mid\bY(x)]\|^2$ (whose explicit definition is given in App.~\ref{app:entropic_contribution}) and its functional inverse by ${\rm mmse}_S^{-1}$ (which exists by monotonicity).
\begin{result}[Replica symmetric free entropy] \label{res:free_entropy}
Let the functional $\tau(\mathcal{Q}_W) := {\rm mmse}_S^{-1}(1 - \EE_{v\sim P_v}[v^2 \mathcal{Q}_W(v)^2])$. Given $(\alpha,\gamma)$, the replica symmetric (RS) free entropy approximating ${\lim}\,f_n$ in the scaling limit \eqref{thermolim} is ${\rm extr}\, f_{\rm RS}^{\alpha,\gamma}$
with RS potential $f^{\alpha,\gamma}_{\rm RS}=f^{\alpha,\gamma}_{\rm RS}( q_2,\hat{q}_2,\mathcal{Q}_W,\hat{\mathcal{Q}}_W)$ given by
\begin{talign}
f^{\alpha,\gamma}_{\rm RS}&:= \psi_{P_{\text{out}}}(q_K(q_2,\mathcal{Q}_W);r_K) + \frac{1}{4\alpha}(1 + \gamma \bar{v}^2-q_2) \Hat{q}_2\nonumber
    \\
    &\qquad + \frac{\gamma}{\alpha}\EE_{ v\sim P_v}\big[
     \psi_{P_W}(\Hat{\mathcal{Q}}_W(v))-\frac{1}{2}\mathcal{Q}_W(v) \Hat{\mathcal{Q}}_W(v) \big]\nonumber \\
     &\qquad +\frac{1}\alpha\big[\iota(\tau(\mathcal{Q}_W)) - \iota(\hat q_2 + \tau(\mathcal{Q}_W))\big].\label{eq:fRS}
\end{talign}
The extremisation operation in ${\rm extr}\,f^{\alpha,\gamma}_{\rm RS}$ selects a solution $( q_2^*,\hat{q}_2^*,\mathcal{Q}_W^*,\hat{\mathcal{Q}}_W^*)$ of the saddle point equations, obtained from $\nabla f^{\alpha,\gamma}_{\rm RS}=\mathbf{0}$, which maximises the RS potential.
\end{result}
The extremisation of $f_{\rm RS}^{\alpha,\gamma}$ yields the system~\eqref{eq:NSB_equations_generic_ch} in the appendix, solved numerically in a standard way (see provided code). 

\begin{figure*}[t!!]
  \centering
  \setlength{\tabcolsep}{4pt} 
  \begin{tabular}{ccc}
    \includegraphics[width=0.33\textwidth,trim={0 0 0.2cm 0.2cm},clip]{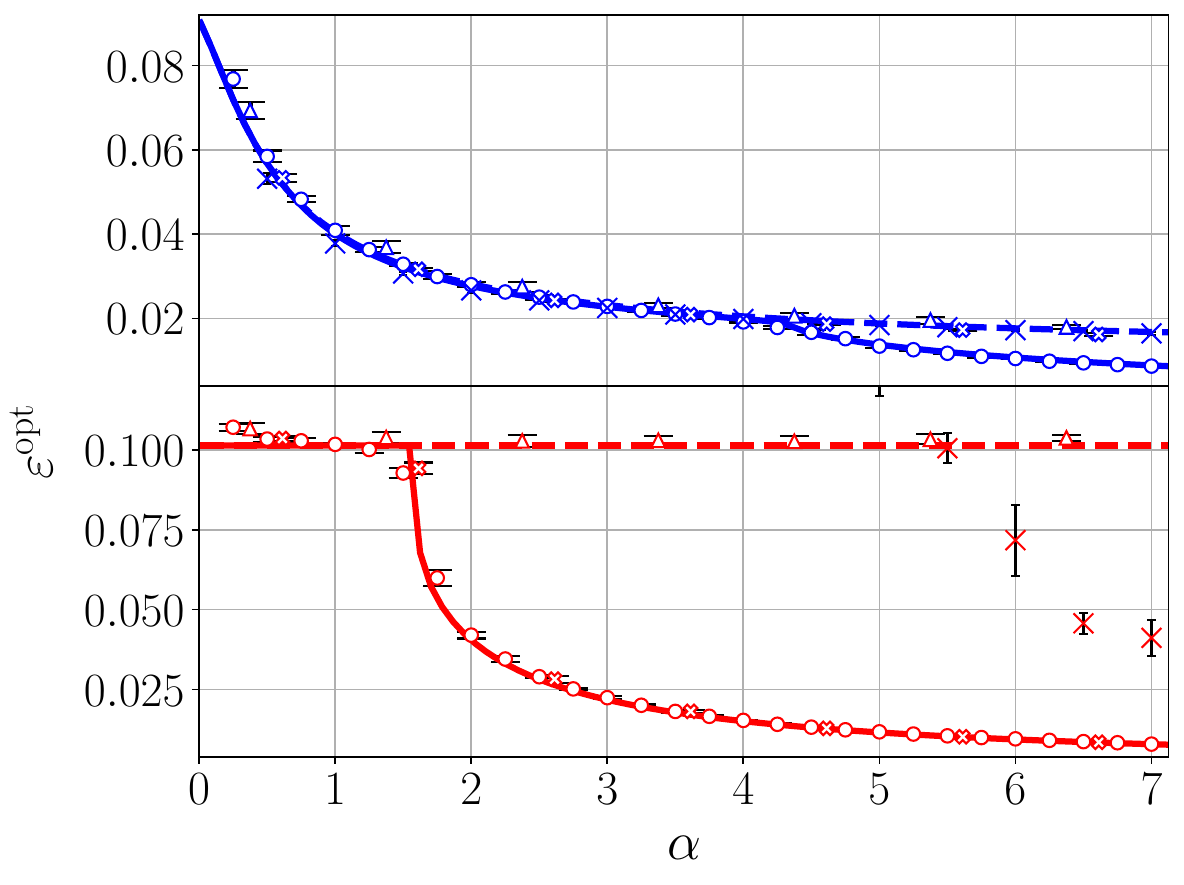} &
    \includegraphics[width=0.29\textwidth,trim={1.2cm 0 0.2cm 0.2cm},clip]{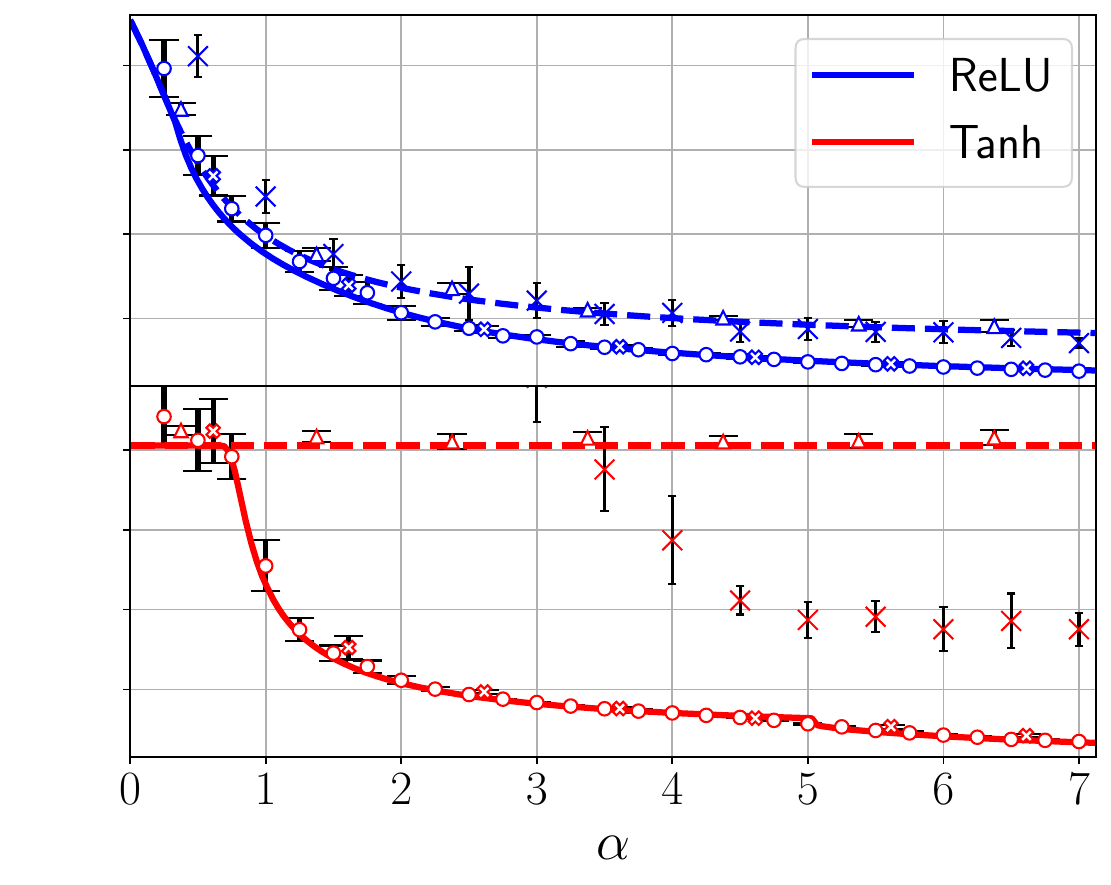} &
    \includegraphics[width=0.3\textwidth,trim={1.1cm 0 0.2cm 0,2cm},clip]{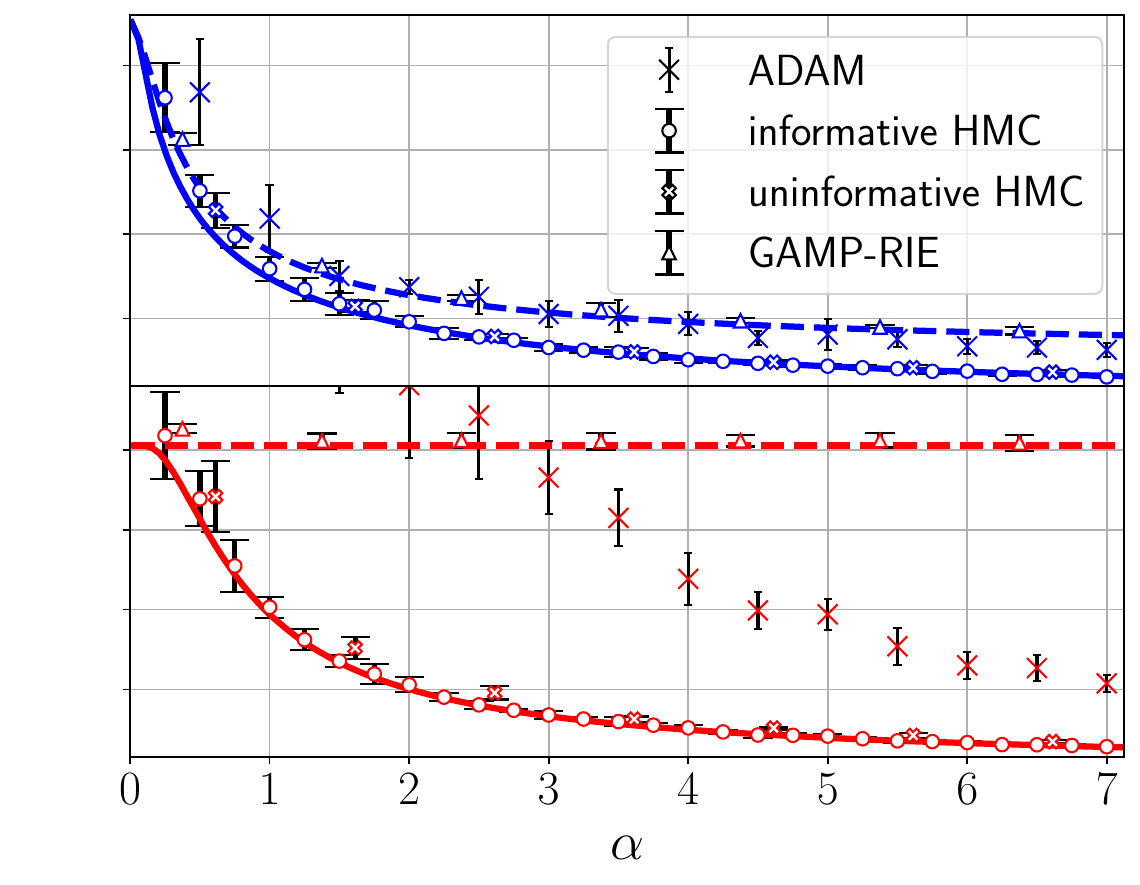} 
  \end{tabular}
  \vspace{-10pt}

  \caption{Theoretical prediction (solid curves) of the Bayes-optimal mean-square generalisation error for \emph{Gaussian inner weights} with ReLU(x) activation (blue curves) and Tanh(2x) activation (red curves), $d=150, \gamma=0.5$, with linear readout with Gaussian label noise of variance $\Delta=0.1$ and different $P_v$ laws. The dashed lines are the theoretical predictions associated with the universal solution, obtained by plugging $\calQ_W(\mathsf{v}) = 0 \ \forall \ \mathsf{v}$ in~\eqref{eq:fRS} and extremising w.r.t. $(q_2,\hat q_2)$ (the curve coincides with the optimal one before the transition $\alpha_{\rm sp}(\gamma)$). The numerical points are obtained with Hamiltonian Monte Carlo (HMC) with informative initialisation on the target (empty circles), uninformative, random, initialisation (empty crosses), and ADAM (thin crosses). Triangles are the error of GAMP-RIE \citep{maillard2024bayes} extended to generic activation, obtained by plugging estimator \eqref{eq:output_GAMP_RIE} in \eqref{eq:gen_error_def} in appendix. Each point has been averaged over 10 instances of the teacher and training set. Error bars are the standard deviation over instances. The generalisation error for a given training set is evaluated as $\frac{1}{2} \EE_{\bx_{\rm test} \sim \mathcal N(0, I_d)} (\lambda_{\rm test}(\bW)-\lambda_{\rm test}^0)^2 $, using a single sample $\bW$ from the posterior for HMC. For ADAM, with batch size fixed to $n/5$ and initial learning rate $0.05$, the error corresponds to the lowest one reached during training, i.e., we use early stopping based on the minimum test loss over all gradient updates. Its generalisation error is then evaluated at this point and divided by two (for comparison with the theory). The average over $\bx_{\rm test}$ is computed empirically from $10^5$ i.i.d. test samples. We exploit that, for typical posterior samples, the Gibbs error $\varepsilon^{\rm Gibbs}$ defined in \eqref{eq:Gibbs_error} in App.~\ref{app:gen_err} is linked to the Bayes-optimal error as $(\varepsilon^{\rm Gibbs} - \Delta)/2 = \varepsilon^{\rm opt} - \Delta$, see \eqref{eq:Gibbs_v_Bayes_error} in appendix. To use this formula, we are assuming the concentration of the Gibbs error w.r.t. the posterior distribution, in order to evaluate it from a single sample per instance. \textbf{Left}: Homogeneous readouts $P_v=\delta_{1}$. \textbf{Centre}: 4-points readouts $P_v = \frac{1}{4}(\delta_{-3/\sqrt{5}} + \delta_{-1/\sqrt{5}} + \delta_{1/\sqrt{5}} +\delta_{3/\sqrt{5}})$. \textbf{Right}: Gaussian readouts $P_v = \mathcal{N}(0,1)$.}
  \label{fig:gen_error_gauss}
  \vskip -0.1in
\end{figure*}

The order parameters $q_2^*$ and $\mathcal{Q}_W^*$ have a precise physical meaning that will be clear from the discussion in Sec.~\ref{sec:theory}. In particular, $q_2^*$ is measuring the alignment of the student's combination of weights $\bW^{\intercal}(\bv)\bW/\sqrt{k}$ with the corresponding teacher's $\bW^{0\intercal}(\bv)\bW^0/\sqrt{k}$, which is non trivial with $n=\Theta(d^2)$ data even when the student is not able to reconstruct $\bW^0$ itself (i.e., to specialise). On the other hand, $\mathcal{Q}_W^*(\mathsf{v})$ measures the overlap between weights $\{\bW_{i}^{0/\cdot}\mid v_i = \mathsf{v} \}$ (a different treatment for weights connected to different $\mathsf{v}$'s is needed because, as discussed earlier, the student will learn first --with less data-- weights connected to larger readouts). A non-trivial $\mathcal{Q}_W^*(\mathsf{v})\neq 0 $ signals that the student learns something about $\bW^0$. Thus, the \emph{specialisation transitions} are naturally defined, based on the extremiser of $f_{\rm RS}^{\alpha,\gamma}$ in the result above, as $\alpha_{\rm sp, \mathsf{v}}(\gamma):=\sup
\,\{\alpha \mid \mathcal{Q}^*_W(\mathsf{v}) =0\}$.
For non-homogeneous readouts, we call \emph{the} specialisation transition $\alpha_{\rm sp}(\gamma):=\min_{\mathsf{v}}\alpha_{\rm sp, \mathsf{v}}(\gamma)$. In this article, we report cases where the inner weights are discrete or Gaussian distributed. For activations different than a pure quadratic, $\sigma(x)\neq x^2$, we predict the transition to occur in both cases (see Fig.~\ref{fig:gen_error_gauss} and~\ref{fig:gen_errors_univ_spec}). Then, $\alpha < \alpha_{\rm sp}$ corresponds to the \emph{universal phase}, where the free entropy is independent of the choice of the prior over the inner weights. Instead, $\alpha > \alpha_{\rm sp}$ is the \emph{specialisation phase} where the prior $P_W$ matters, and the student aligns a finite fraction of its weights $(\bW_j)_{j\le k}$ with those of the teacher, which lowers the generalisation error. 

Let us comment on why the special case $\sigma(x)= x^2$ with $P_W=\mathcal{N}(0,1)$ could be treated exactly with known techniques (spherical integration) in \cite{maillard2024bayes,xu2025fundamental}. With $\sigma(x)= x^2$ the responses $(y_\mu)$ depend on $\bW^{0\intercal}(\bv)\bW^0$ only. If $\bv$ has finite fractions of equal entries, a large invariance group prevents learning $\bW^0$ and thus specialisation. Take as example $\bv=(1,\ldots,1,-1,\ldots,-1)$ with the first half filled with ones. Then, the responses are indistinguishable from those obtained using a modified matrix $\bW^{0\intercal}\bU^\intercal(\bv)\bU\bW^0$ where $\bU=((\bU_{1},\mathbf{0}_{d/2})^\intercal,(\mathbf{0}_{d/2},\bU_{2})^\intercal)$ is block diagonal with $d/2\times d/2$ orthogonal $\bU_{1},\bU_{2}$ and zeros on off-diagonal blocks. The Gaussian prior $P_W$ is rotationally invariant and, thus, does not break any invariance, so $\bU_{1},\bU_{2}$ are arbitrary. The resulting invariance group has an $\Theta(d^2)$ entropy (the logarithm of its volume), which is comparable to the leading order of the free entropy. Therefore, it cannot be broken using infinitesimal perturbations (or ``side information'') and, consequently, prevents specialisation. This reasoning can be extended to $P_v$ with a continuous support, as long as we can discretise it with a finite (possibly large) number of bins, take the limit \eqref{thermolim} first, and then take the continuum limit of the binning afterwards. However, the picture changes if the prior breaks rotational invariance; e.g., with Rademacher $P_W$, only signed permutation invariances survive, a symmetry with negligible entropy $o(d^2)$ which, consequently, does not change the limiting thermodynamic (information-theoretic) quantities. The large rotational invariance group is the reason why $\sigma(x)=x^2$ with $P_W=\mathcal{N}(0,1)$ can be treated using the HCIZ integral alone. Even when $P_W=\mathcal{N}(0,1)$, the presence of any other term in the series expansion of $\sigma$ breaks invariances with large entropy: specialisation can then occur, thus requiring our theory. We mention that our theory seems inexact\footnote{When solving the extremisation of \eqref{eq:fRS} for $\sigma(x)=x^2$ with $P_W=\mathcal{N}(0,1)$, we noticed that the difference between the RS free entropy of the correct universal solution, $\mathcal{Q}_W(\mathsf{v})=0$, and the maximiser, predicting $\mathcal{Q}_W(\mathsf{v})>0$, does not exceed $\approx 1\%$: the RS potential is very flat as a function of $\mathcal{Q}_W$. We thus cannot discard that the true maximiser of the potential is at $\mathcal{Q}_W(\mathsf{v})=0$, and that we observe otherwise due to numerical errors. Indeed, evaluating the spherical integrals $\iota(\,\cdot\,)$ in $f^{\alpha,\gamma}_{\rm RS}$ is challenging, in particular when $\gamma$ is small. Actually, for $\gamma \gtrsim 1$ we do get that  $\mathcal{Q}_W(\mathsf{v})=0$ is always the maximiser for $\sigma(x)=x^2$ with $P_W=\mathcal{N}(0,1)$.} for $\sigma(x)= x^2$ with $P_W=\mathcal{N}(0,1)$ if applied naively, as it predicts $\calQ_W(\mathsf{v})>0$ and therefore does not recover the rigorous result of \cite{xu2025fundamental} (yet, it predicts a free entropy less than $1\%$ away from the truth). 
Nevertheless, the solution of \cite{maillard2024bayes,xu2025fundamental} is recovered from our equations by enforcing a vanishing overlap $\mathcal{Q}_W(\mathsf{v})=0$, i.e., via its universal branch.

\paragraph*{Bayes generalisation error} 

Another main result is an approximate formula for the generalisation error. Let 
$(\bW^a)_{a\ge 1}$ be i.i.d.\ samples from the posterior $dP(\,\cdot \mid\mathcal{D})$ and $\bW^0$ the teacher's weights. Assuming that the joint law of $(\lambda_{\rm test}(\bW^a,\bx_{\rm test}))_{a\ge 0}=:(\lambda^a)_{a\ge 0}$ for a common test input $\bx_{\rm test}  \notin \mathcal{D}$ is a centred Gaussian, our framework predicts its covariance. Our approximation for the Bayes error follows.
\begin{result}[Bayes generalisation error]
\label{res:gen_error}
Let $q_K^*=q_K(q_2^*,\mathcal{Q}_W^*)$ where $(q_2^*,\hat q_2^*,\mathcal{Q}_W^*,\Hat{\mathcal{Q}}_W^*)$ is an extremiser of $f_{\rm RS}^{\alpha,\gamma}$ as in Result~\ref{res:free_entropy}. Assuming joint Gaussianity of the post-activations $(\lambda^a)_{a\ge 0}$, in the scaling limit \eqref{thermolim} their mean is zero and their covariance is approximated by $\EE \lambda^a \lambda^{b} = q_K^*+(r_K-q_K^*)\delta_{ab}=:(\mathbf{\Gamma})_{ab}$, see App.~\ref{app:gen_err}. 

Assume $\mathcal{C}$ has the series expansion $\mathcal{C}(y,\hat y)=\sum_{i\ge 0} c_i(y)\hat y^i$. The Bayes error $\smash{\lim\,\varepsilon^{\mathcal{C},\mathsf{f}}}$ is approximated by
\begin{talign*}
   \EE_{(\lambda^a)\sim \mathcal{N}(\mathbf{0},\mathbf{\Gamma})}\EE_{y_{\rm test}\sim  P_{\rm out}(\,\cdot\mid \lambda^0)} \sum_{i\ge 0} c_i(y_{\rm test}(\lambda^0))\prod_{a=1}^i\mathsf{f}(\lambda^a).
\end{talign*}
Letting $\EE[\,\cdot\mid \lambda]=\int dy \,(\,\cdot\,)\, P_{\rm out}(y\mid \lambda)$, the Bayes-optimal mean-square generalisation error $\smash{\lim\, \varepsilon^{\rm opt}}$ is approximated by
  \begin{talign}
\EE_{\lambda^0,\lambda^1}\big(\EE[y^2\mid \lambda^0] - \EE[y\mid \lambda^0]\EE[y\mid \lambda^1] \big).\label{eq:gen_err_result}
  \end{talign}
\end{result}
This result assumed that $\mu_0=0$; see App.~\ref{app:non-centered} if this is not the case. Results~\ref{res:free_entropy} and \ref{res:gen_error} provide an effective theory for the generalisation capabilities of Bayesian shallow networks with generic activation. We call these ``results'' because, despite their excellent match with numerics, we do not expect these formulas to be exact: their derivation is based on an unconventional mix of spin glass techniques and spherical integrals, and require approximations in order to deal with the fact that the degrees of freedom to integrate are large matrices of extensive rank. This is in contrast with simpler (vector) models (perceptrons, multi-index models, etc) where replica formulas are routinely proved correct, see e.g. \cite{barbier2019adaptive,barbier2019glm,aubin2018committee}.

\section{Theoretical predictions and numerical experiments}

\begin{figure}[t!]
\begin{center}
\centerline{\includegraphics[width=1\linewidth,trim={0cm 0cm 0cm 0.4cm},clip]{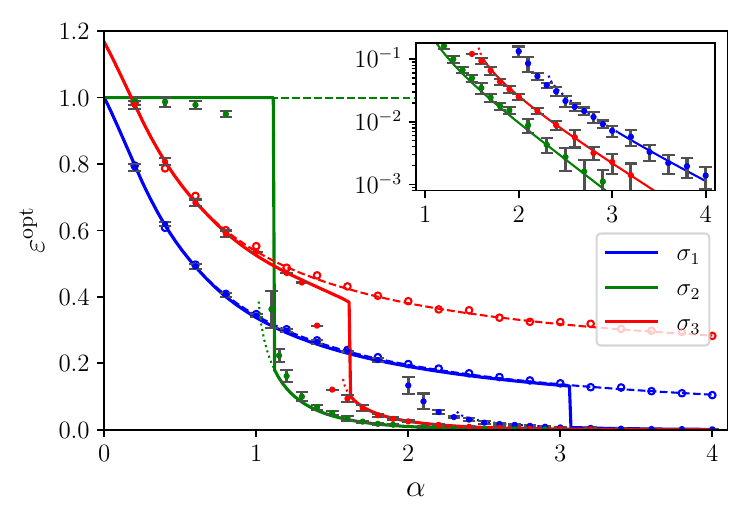}}
\vspace{-10pt}
    \caption{Theoretical prediction (solid curves) of the Bayes-optimal mean-square generalisation error for \emph{binary inner weights} and polynomial activations:
    $\sigma_1 = \He_2/\sqrt 2$, $\sigma_2 = \He_3/\sqrt 6$, $\sigma_3 = \He_2/\sqrt 2 + \He_3/6$,
    with $\gamma = 0.5$, $d=150$, linear readout with Gaussian label noise with $\Delta=1.25$, and homogeneous readouts $\bv= \mathbf{1}$. 
    Dots are optimal errors computed via Gibbs errors (see Fig.~\ref{fig:gen_error_gauss}) by running a Metropolis-Hastings MCMC initialised near the teacher.
    Circles are the error of GAMP-RIE
    \citep{maillard2024bayes} extended to generic activation, see App.~\ref{app:GAMP}.
    Points are averaged over 16 data instances. Error bars for MCMC are the standard deviation over instances (omitted for GAMP-RIE, but of the same order). Dashed and dotted lines denote, respectively, the universal (i.e. the $\mathcal{Q}_W(\mathsf{v})=0 \ \forall \ \mathsf{v}$ solution of the saddle point equations) and the specialisation branches where they are metastable (i.e., a local maximiser of the RS potential but not the global one).}
    \label{fig:gen_errors_univ_spec}
\end{center}
\vskip -0.4in
\end{figure}

Let us compare our theoretical predictions with simulations. In Fig.~\ref{fig:gen_error_gauss} and \ref{fig:gen_errors_univ_spec}, we report the theoretical curves from Result~\ref{res:gen_error}, focusing on the optimal mean-square generalisation error for networks with different $\sigma$, with linear readout with Gaussian noise variance $\Delta$. The Gibbs error divided by $2$ is used to compute the optimal error, see Remark~\ref{rem:Gibbs_error} in App.~\ref{app:gen_err} for a justification. In what follows, the error attained by ADAM is also divided by two, only for the purpose of comparison.

Figure~\ref{fig:gen_error_gauss} focuses on networks with Gaussian inner weights, various readout laws, for $\sigma(x)=\relu(x)$ and $\Tanh(2x)$. Informative (i.e., on the teacher) and uninformative (random) initialisations are used when sampling the posterior by HMC. We also run ADAM, always selecting its best performance over all epochs, and implemented an extension of the GAMP-RIE of~\cite{maillard2024bayes} for generic activation (see App.~\ref{app:GAMP}). It can be shown analytically that GAMP-RIE's generalisation error asymptotically (in $d$) matches the prediction of the universal branch of our theory (i.e., associated with $\mathcal{Q}_W(\mathsf{v}) = 0 \ \forall \ \mathsf{v}$).

For ReLU activation and homogeneous readouts (left panel), informed HMC  follows the specialisation branch (the solution of the saddle point equations with $\mathcal{Q}_W(\mathsf{v})\neq0$ for at least one $\mathsf{v}$), while with uninformative initialisation it sticks to the universal branch, thus suggesting algorithmic hardness. We shall be back to this matter in the following. We note that the error attained by ADAM (divided by 2), is close to the performance associated with the universal branch, which suggests that ADAM is an effective Gibbs estimator for this $\sigma$. For Tanh and homogeneous readouts, both the uninformative and informative points lie on the specialisation branch, while ADAM attains an error greater than twice the posterior sample's generalisation error. 

For non-homogeneous readouts (centre and right panels)
the points associated with the informative initialisation lie consistently on the specialisation branch, for both $\relu$ and Tanh, while the uninformatively initialised samples have a slightly worse performance for Tanh. 
Non-homogeneous readouts improves the ADAM performance: for Gaussian readouts and high sampling ratio its half-generalisation error is consistently below the error associated with the universal branch of the theory.

Figure~\ref{fig:gen_errors_univ_spec} concerns networks with Rademacher weights and homogeneous readout. The numerical points are of two kinds: the dots, obtained from Metropolis–Hastings sampling of the weight posterior, and the circles, obtained from the GAMP-RIE (App.~\ref{app:GAMP}). We report analogous simulations for $\relu$ and $\elu$ activations in \figurename~\ref{fig:GAMP-RIE_eLU-ReLU},  App.~\ref{app:GAMP}. The remarkable agreement between theoretical curves and experimental points in both phases supports the assumptions used in Sec.~\ref{sec:theory}.

\begin{figure}[t!!]
  \centering
  \includegraphics[width=1\linewidth,trim={0.3cm 0 0 0},clip]{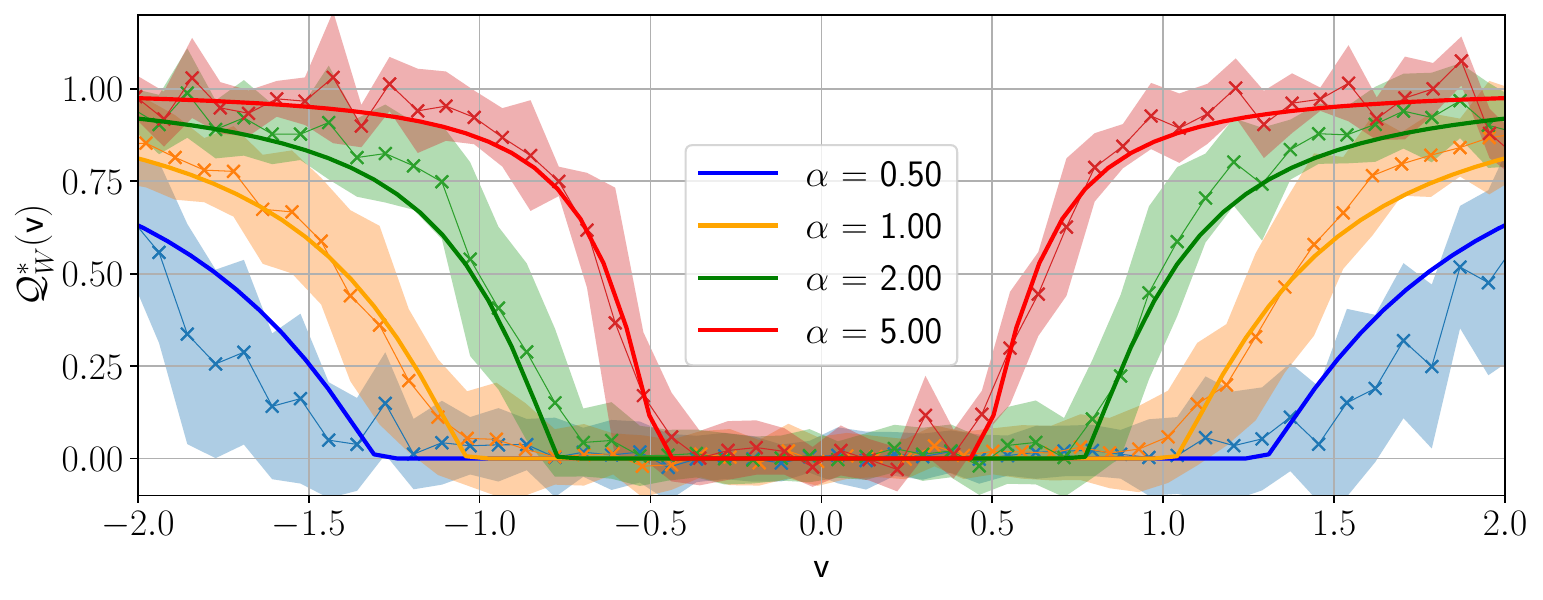} \\
  \vspace{-3pt}
  \includegraphics[width=0.955\linewidth,trim={0 0 0.2cm 0},clip]{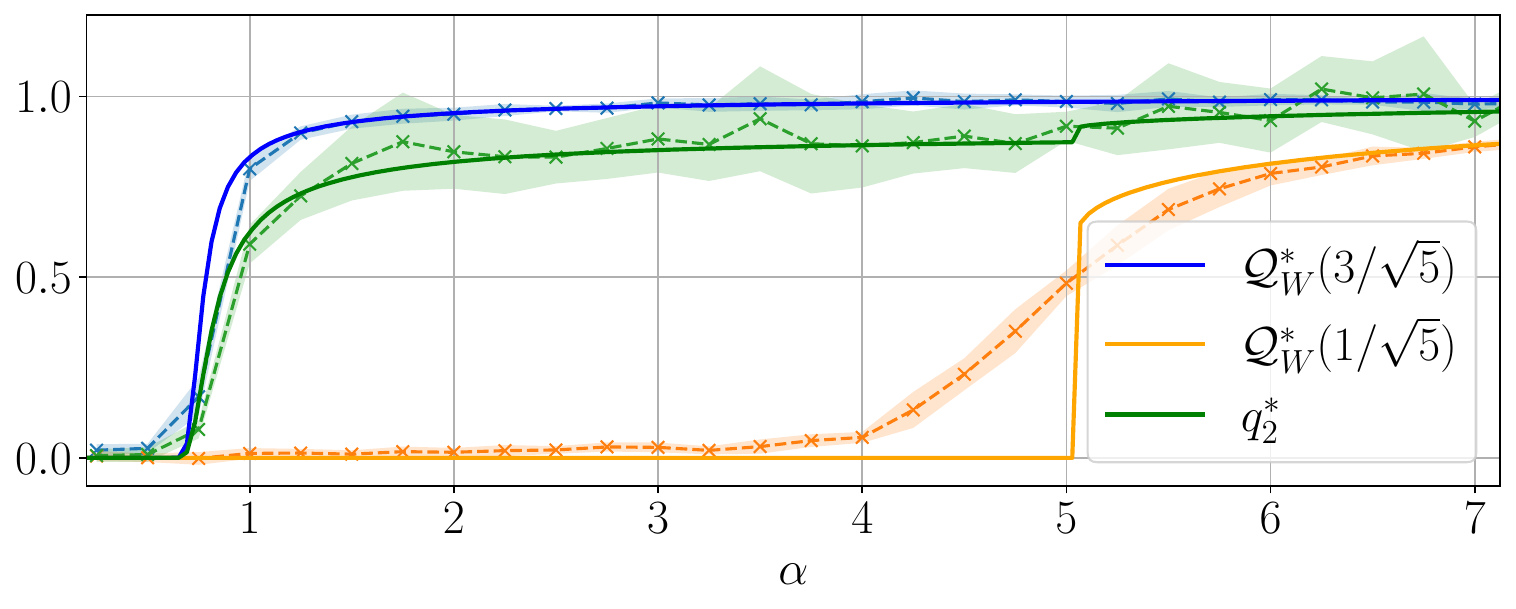}\hspace*{-0.1cm}
  \vspace{-5pt}
  \caption{\textbf{Top}: Theoretical prediction (solid curves) of the overlap function $\mathcal{Q}_W(\mathsf{v})$ for different sampling ratios $\alpha$ for \emph{Gaussian inner weights}, ReLU(x) activation, $d=150, \gamma=0.5$, linear readout with $\Delta=0.1$ and $P_v = \mathcal{N}(0,1)$. The shaded curves were obtained from HMC initialised informatively. Using a single sample $\bW^a$ from the posterior,  $\mathcal{Q}_W(\mathsf{v})$ has been evaluated numerically by dividing the interval $[-2,2]$ into 50 bins and by computing the value of the overlap associated with each bin. Each point has been averaged over 50 instances of the training set, and shaded regions around them correspond to one standard deviation. \textbf{Bottom}: Theoretical prediction (solid curves) of the overlaps as function of the sampling ratio $\alpha$ for \emph{Gaussian inner weights}, Tanh(2x) activation, $d=150, \gamma=0.5$, linear readout with $\Delta=0.1$ and $P_v = \frac{1}{4}(\delta_{-3/\sqrt{5}} + \delta_{-1/\sqrt{5}} + \delta_{1/\sqrt{5}} +\delta_{3/\sqrt{5}})$. The shaded curves were obtained from informed HMC. Each point has been averaged over 10 instances of the training set, with one standard deviation depicted.}
  \label{fig:v_non-constant}
\end{figure}

Figure~\ref{fig:v_non-constant} illustrates the learning mechanism for models with Gaussian weights and non-homogeneous readouts, revealing a sequence of phase transitions as $\alpha$ increases. Top panel shows the overlap function $\mathcal{Q}_W(\mathsf{v})$ in the case of Gaussian readouts for four different sample rates $\alpha$. In the bottom panel the readout assumes four different values with equal probabilities; the figure shows the evolution of the two relevant overlaps associated with the symmetric readout values $\pm 3/\sqrt{5}$ and $\pm 1/\sqrt{5}$. As $\alpha$ increases, the student weights start aligning with the teacher weights associated with the highest readout amplitude, marking the first phase transition. As these alignments strengthen when $\alpha$ further increases, the second transition occurs when the weights corresponding to the next largest readout amplitude are learnt, and so on. In this way, continuous readouts produce an infinite sequence of learning transitions, as supported by the upper part of \figurename~\ref{fig:v_non-constant}.

Even when dominating the posterior measure, we observe in simulations that the specialisation solution can be algorithmically hard to reach. With a discrete distribution of readouts (such as $P_v= \delta_1$ or Rademacher), simulations for binary inner weights exhibit it only when sampling with informative initialisation (i.e., the MCMC runs to sample $\bW$ are initialised in the vicinity of $\bW^0$). Moreover, even in cases where algorithms (such as ADAM or HMC for Gaussian inner weights) are able to find the specialisation solution, they manage to do so only after a training time increasing exponentially with $d$, and for relatively small values of the label noise $\Delta$,
see discussion in App.~\ref{app:hardness}.
For the case of the continuous distribution of readouts $P_v=\calN(0,1)$, our numerical results are inconclusive on hardness, and deserve numerical investigation at a larger scale.

The universal phase is superseded at $\alpha_{\rm sp}$ by a specialisation phase, where the student's inner weights start aligning with the teacher ones. This transition occurs for both binary and Gaussian priors over the inner weights, and it is different in nature w.r.t. the perfect recovery threshold identified in~\cite{maillard2024bayes}, which is the point where the student with Gaussian weights learns perfectly $\bW^{0\intercal}(\bv)\bW^0$ (but \emph{not} $\bW^{0}$) and thus attains perfect generalisation in the case of purely quadratic activation and noiseless labels. For large $\alpha$, the student somehow realises that the higher order terms of the activation's Hermite decomposition are not label noise, but are informative on the decision rule. The two identified phases are akin to those recently described in \cite{barbier2024phase} for matrix denoising. The model we consider is also a matrix model in $\bW$, with the amount of data scaling as the number of matrix elements. When data are scarce, the student cannot break the numerous symmetries of the problem, resulting in an ``effective rotational invariance'' at the source of the prior universality, with posterior samples having a vanishing overlap with $\bW^0$. On the other hand, when data are sufficiently abundant, $\alpha>\alpha_{\rm sp}$, there is a ``synchronisation'' of the student's samples with the teacher. 

The phenomenology observed depends on the activation function selected. 
In particular, by expanding $\sigma$ in the Hermite basis we realise that the way the first three terms enter information theoretical quantities is completely described by order 0, 1 and 2 tensors later defined in \eqref{eq:def_S}, that give rise to combinations of the inner and readout weights. In the regime of quadratically many data, order 0 and 1 tensors are recovered exactly by the student because of the overwhelming abundance of data compared to their dimension. The challenge is thus to learn the second order tensor. On the contrary, we claim that learning any higher order tensors can only happen when the student aligns its weights with $\bW^0$: before this ``synchronisation'', they play the role of an effective noise. This is the mechanism behind the specialisation transition.
For odd activation ($\Tanh$ in \figurename~\ref{fig:gen_error_gauss}, $\sigma_3$ in \figurename~\ref{fig:gen_errors_univ_spec}), where $\mu_2=0$, the aforementioned order-2 tensor does not contribute any more to the learning. Indeed, we observe numerically that the generalisation error sticks to a constant value for $\alpha<\alpha_{\rm sp}$, whereas at the phase transition it suddenly drops. This is because the learning of the order-2 tensor is skipped entirely, and the only chance to perform better is to learn all the other higher-order tensors through specialisation.

By extrapolating universality results to generic activations, we are able to use the GAMP-RIE of~\cite{maillard2024bayes}, publicly available at~\cite{maillard2024github}, to obtain a polynomial-time predictor for test data. Its generalisation error follows our universal theoretical curve even in the $\alpha$ regime where MCMC sampling experiences a computationally hard phase with worse performance (for binary weights), and in particular after $\alpha_{\rm sp}$ (see Fig.~\ref{fig:gen_errors_univ_spec}, circles). Extending this algorithm, initially proposed for quadratic activation, to a generic one is possible thanks to the identification of an \emph{effective} GLM onto which the learning problem can be mapped (while the mapping is exact when $\sigma(x)=x^2$ as exploited by \cite{maillard2024bayes}), see App.~\ref{app:GAMP}. The key observation is that our effective GLM representation holds not only from a theoretical perspective when describing the universal phase, but also algorithmically.

Finally, we emphasise that our theory is consistent with \cite{cui2023bayes}, which considers the simpler strongly over-parametrised regime $n=\Theta(d)$ rather than the interpolation one $n=\Theta(d^2)$: our generalisation curves at $\alpha\to0$ match theirs at $\alpha_1:=n/d\to\infty$, which is when the student learns perfectly the combinations $\bv^{0\intercal}\bW^0/\sqrt{k}$ (but nothing more).

\section{Accessing the free entropy and generalisation error: replica method and spherical integration combined}\label{sec:theory}
The goal is to compute the asymptotic free entropy by the replica method \cite{mezard1987spin}, a powerful heuristic from spin glasses also used in machine learning \cite{engel2001statistical}, combined with the HCIZ integral. Our derivation is based on a Gaussian ansatz on the replicated post-activations of the hidden layer, which generalises Conjecture 3.1 of~\cite{cui2023bayes}, now proved in~\cite{camilli2025inforeduction}, where it is specialised to the case of linearly many data ($n=\Theta(d)$). To obtain this generalisation, we will write the kernel arising from the covariance of the aforementioned post-activations as an infinite series of scalar order parameters derived from the expansion of the activation function in the Hermite basis, following an approach recently devised in~\cite{aguirre2024RF} in the context of the random features model (see also \cite{hu2024asymptotics} and \cite{10.1214/20-AOS1990}). Another key ingredient of our analysis will be a generalisation of an ansatz used in the replica method by \cite{sakata2013} for dictionary learning.

\subsection{Replicated system and order parameters}
The starting point in the replica method to tackle the data average is the replica trick:
\begin{talign*}
 {\lim}\,\frac1n\EE \ln \calZ(\mathcal{D}) ={\lim} { \lim\limits_{\,\,s\to 0^+}}\!\frac{1}{ns}\ln\EE\mathcal{Z}^s = \lim\limits_{\,\,s\to 0^+}\!{\lim}\,\frac{1}{ns} \ln\EE\mathcal{Z}^s  
\end{talign*}
assuming the limits commute. Recall $\bW^0$ are the teacher weights. Consider first $s\in \mathbb{N}^+$. Let the ``replicas'' of the post-activation $ \{\lambda^a(\bW^a):=\frac1{\sqrt k}{\bv^{\intercal}}\sigma(\frac1{\sqrt d}{\bW^a\bx})\}_{a=0,\ldots,s}$.
We then directly obtain
\begin{talign*}
\EE\mathcal{Z}^s
=\EE_{\bv}\int \prod\limits_{a}\limits^{0,s}dP_W(\bW^a)\big[\EE_\bx\int dy \prod\limits_{a}\limits^{0,s} P_{\rm out}(y\mid \lambda^a(\bW^a))\big]^n.
\end{talign*}
The key is to identify the law of the replicas $\{\lambda^a\}_{a=0,\ldots,s}$,
which are dependent random variables due to the common random Gaussian input $\bx$, conditionally on $(\bW^a)$. Our key \emph{hypothesis} is that $\{\lambda^a\}$ is jointly Gaussian, an ansatz we cannot prove 
but that we validate a posteriori thanks to the excellent match between our theory and the empirical generalisation curves, see Sec.~\ref{sec:result}. Similar Gaussian assumptions have been the crux of a whole line of recent works on the analysis of neural networks, and are now known under the name of ``Gaussian equivalence'' \citep{goldt2020modeling,hastie2022surprises,mei2022generalization,goldt2022gaussian,hu2022universality}. This can also sometimes be heuristically justified based on Breuer–Major Theorems \citep{nourdin2011quantitative,pacelli2023}. 

Given two replica indices $a,b\in\{0,\ldots,s\}$ we define the neuron-neuron overlap matrix $\Omega^{ab}_{ij}:={\bW_{i}^{a\intercal}\bW^b_{j}}/d$ with $i,j\in[k]$. Recalling the Hermite expansion of $\sigma$, by using Mehler's formula, see App.~\ref{app:hermite}, the post-activations covariance $K^{ab}:=\EE\lambda^a\lambda^b$ reads
\begin{talign}
K^{ab}&=\sum_{\ell\ge 1}^\infty\frac{\mu^2_\ell}{\ell!} Q_\ell^{ab} \ \  \text{with}\ \  Q_\ell^{ab}:=\frac{1}k\sum_{i,j\le k} v_iv_j(\Omega^{ab}_{ij})^\ell .
\label{eq:K}
\end{talign}
This covariance $\bK$ is complicated but, as we argue hereby, simplifications occur as $d\to\infty$. In particular, the first two overlaps $Q_1^{ab},Q_2^{ab}$ are special.
We claim that higher-order overlaps $(Q_\ell^{ab})_{\ell\ge 3}$ can be simplified as functions of simpler order parameters.

\begin{figure}[t!!]
    \centering
    \includegraphics[width=0.48\linewidth]{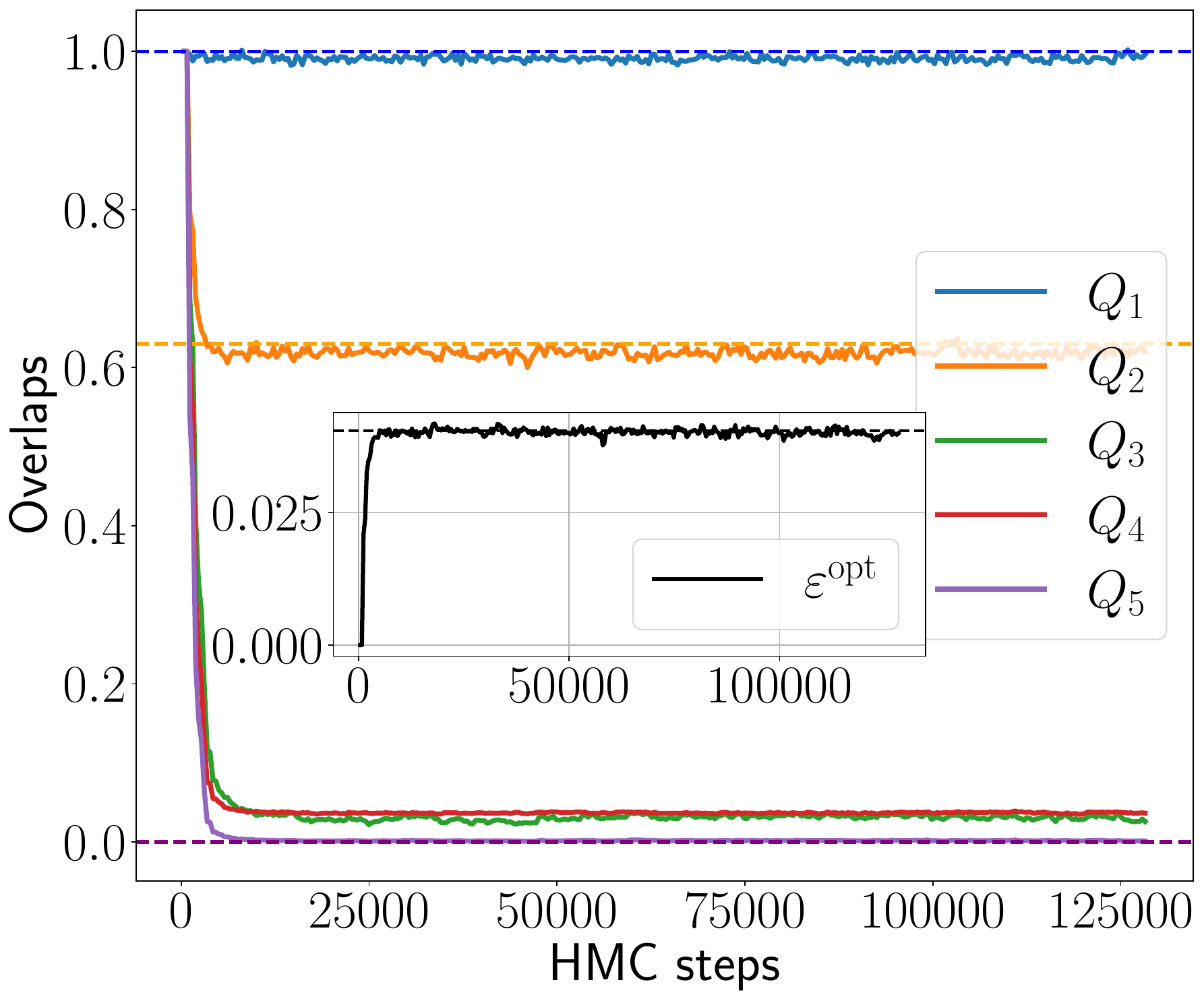}
    \hfill
    \includegraphics[width=0.48\linewidth]{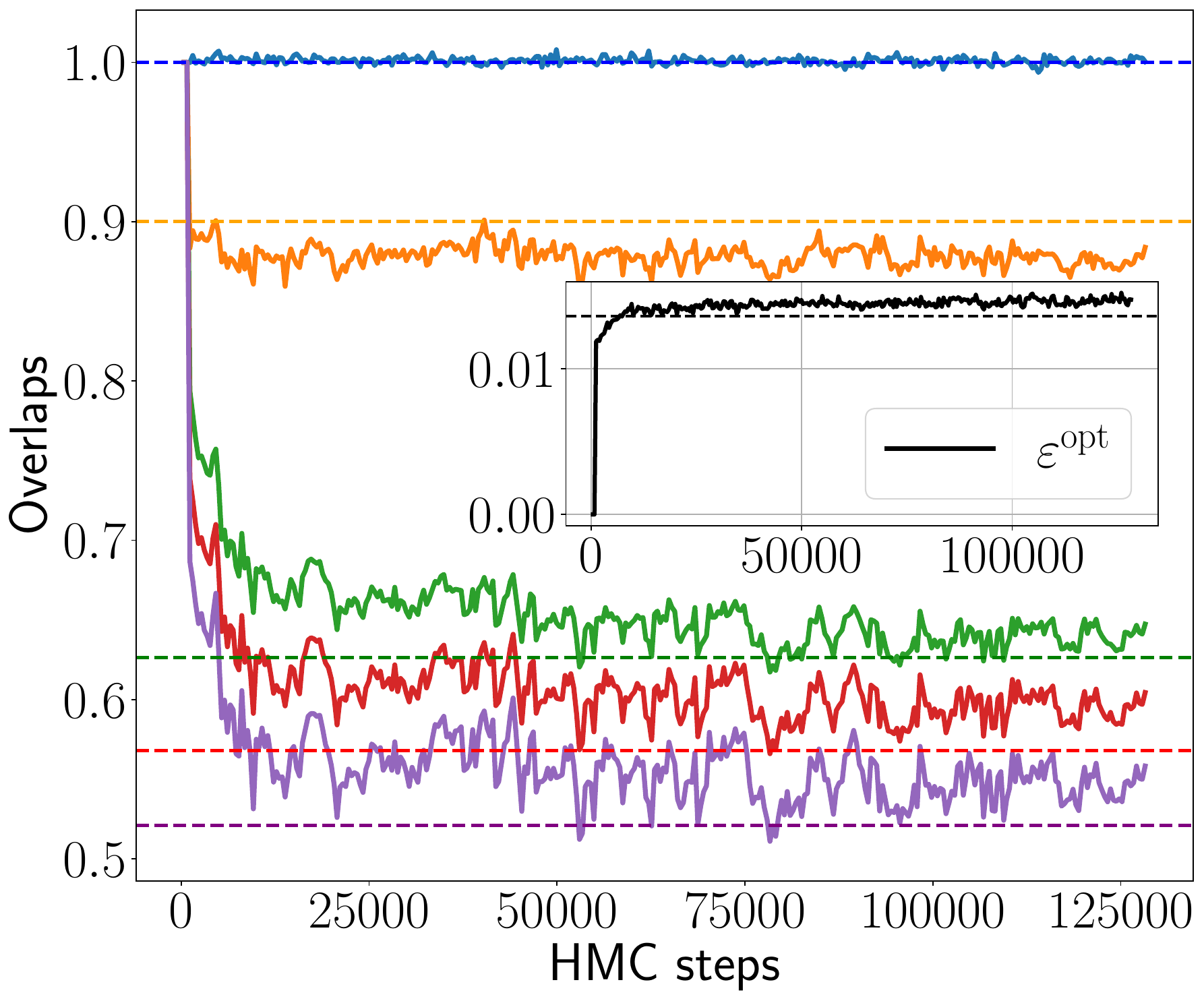}
    \vspace{-5pt}
    \caption{Hamiltonian Monte Carlo dynamics of the overlaps $Q_\ell=Q_\ell^{01}$ between student and teacher weights for $\ell \in [5]$, with activation function ReLU(x), $d=200$, $\gamma=0.5$, linear readout with $\Delta=0.1$ and two choices of sample rates and readouts: $\alpha=1.0$ with $P_v = \delta_{1}$ (\textbf{Left}) and $\alpha=3.0$ with $P_v = \mathcal{N}(0,1)$ (\textbf{Right}). The teacher weights $\bW^0$ are Gaussian. The dynamics is initialised informatively, i.e., on $\bW^0$. The overlap $Q_1$ always fluctuates around 1. \textbf{Left}: The overlaps $Q_\ell$ for $\ell  \ge 3$ at equilibrium converge to 0, while $Q_2$ is well estimated by the theory (orange dashed line). \textbf{Right}: At higher sample rate $\alpha$, also the $Q_\ell$ for $\ell \ge 3$ are non zero and agree with their theoretical prediction (dashed lines). Insets show the mean-square generalisation error and the theoretical prediction.}
    \label{fig:overlaps}
\end{figure}

\subsection{Simplifying the order parameters}
In this section we show how to drastically reduce the number of order parameters to track. Assume at the moment that the readout prior $P_v$ has discrete support $\mathsf{V}=\{ \mathsf{v} \}$; this can be relaxed by binning a continuous support, as mentioned in Sec.~\ref{sec:result}. The overlaps in \eqref{eq:K} can be written as
\begin{talign}
    Q_\ell^{ab} = \frac1k\sum_{\mathsf{v},\mathsf{v}'\in \mathsf{V}} \mathsf{v} \,\mathsf{v}'  \sum_{\{i,j\le k\mid v_i =\mathsf{v}, v_j= \mathsf{v}'  \}} (\Omega_{ij}^{ab})^\ell.
\end{talign}
In the following, for $\ell \ge 3$ we discard the terms $\mathsf{v}\neq \mathsf{v}'$ in the above sum, assuming they are suppressed w.r.t. the diagonal ones. In other words, a neuron $\bW^a_i$ of a student (replica) with a readout value $v_i=\mathsf{v}$ is assumed to possibly align only with neurons of the teacher (or, by Bayes-optimality, of another replica) with the same readout. Moreover, in the resulting sum over the neurons indices $\{i,j\mid v_i = v_j = \mathsf{v} \}$, we assume that, for each $i$, a single index $j=\pi_i$, with $\pi$ a permutation, contributes at leading order. The model is symmetric under permutations of hidden neurons. We thus take $\pi$ to be the identity without loss of generality. 

We now assume that for Hadamard powers $\ell \ge 3$, the off-diagonal of the overlap $(\bOmega^{ab})^{\circ \ell}$, obtained from typical weight matrices sampled from the posterior, is  sufficiently small to consider it diagonal in any quadratic form. Moreover, by exchangeability among neurons with the same readout value, we further assume that all diagonal elements $\{\Omega_{ii}^{ab}\mid i \in \mathcal{I}_\mathsf{v}\}$ concentrate onto the constant $\mathcal{Q}_W^{ab}(\mathsf{v})$, where $\mathcal{I}_\mathsf{v}:=\{i\le k\mid v_i=\mathsf{v}\}$:
\begin{talign}\label{eq:Omega_ansatz}    
(\Omega_{ij}^{ab})^\ell = (\frac1d \bW_{i}^{a\intercal} \bW^b_{j})^\ell\approx\delta_{ij}\mathcal{Q}_W^{ab}(\mathsf{v})^\ell 
\end{talign}
if $\ell\geq 3$, $i\ \text{or}\ j\in \mathcal{I}_\mathsf{v}$.
Approximate equality here is up to a matrix with $o_d(1)$ norm. The same happens, e.g., for a standard Wishart matrix: its eigenvectors and the ones of its square Hadamard power are delocalised, while for higher Hadamard powers $\ell\ge 3$ its eigenvectors are strongly localised; this is why $Q_2^{ab}$ will require a separate treatment. With these simplifications we can write 
\begin{talign}
\label{eq:Qell_NSB}
   Q_\ell^{ab}=\EE_{v\sim P_v}[v^2 \mathcal{Q}_{W}^{ab}(v)^\ell] +o_d(1) \  \text{for}\  \ell\geq 3.
\end{talign}
This is is verified numerically a posteriori as follows. Identity \eqref{eq:Qell_NSB} is true (without $o_d(1)$) for the predicted theoretical values of the order parameters by construction of our theory. Fig.~\ref{fig:v_non-constant} verified the good agreement between the theoretical and experimental overlap profiles $\mathcal{Q}^{01}_W(\mathsf{v})$ for all $\mathsf{v}\in\mathsf{V}$ (which is statistically the same as  $\smash{\mathcal{Q}^{ab}_W(\mathsf{v})}$ for any $a\neq b$ by the so-called Nishimori identity following from Bayes-optimality, see App.~\ref{app:nishiID}), while Fig.~\ref{fig:overlaps} verifies the agreement at the level of $(Q_\ell^{ab})$. Consequently, \eqref{eq:Qell_NSB} is also  true for the experimental overlaps.

It is convenient to define the symmetric tensors $\bS_\ell^a$ with entries
\begin{talign}   
S^a_{\ell;\alpha_1\ldots\alpha_\ell} := \frac{1}{\sqrt{k}}\sum_{i\le k} v_i W^a_{i\alpha_1} \cdots W^a_{i\alpha_\ell} .   \label{eq:def_S}
\end{talign}
Indeed, the generic $\ell$-th term of the series~\eqref{eq:K} can be written as the overlap $Q^{ab}_\ell=\langle \bS^a_\ell,\bS^b_\ell\rangle/d^\ell$ of these tensors (where $\langle\,,\,\rangle$ is the inner product), e.g., $Q_2^{ab} =  \Tr \,\bS_{2}^{a}\bS_{2}^{b}/d^2$. Given that the number of data $n=\Theta(d^2)$ and that $(\bS_1^a)$ are only $d$-dimensional, they are reconstructed perfectly (the same argument was used to argue that readouts $\bv$ can be quenched). We thus assume right away that at equilibrium the overlaps $Q_1^{ab}=1$ (or saturate to their maximum value; if tracked, the corresponding saddle point equations end up being trivial and do fix this). In other words, in the quadratic data regime, the $\mu_1$ contribution in the Hermite decomposition of $\sigma$ for the target is perfectly learnable, while higher order ones play a non-trivial role. In contrast, \cite{cui2023bayes} study the regime $n=\Theta(d)$ where $\mu_1$ is the only learnable term.

Then, the average replicated partition function reads $\EE\mathcal{Z}^s = \int d\bQ_2 d\bm{\mathcal{Q}}_W \exp(F_S\!+\! nF_E)$
where $F_E, F_S$ depend~on~$\bQ_2=(Q_{2}^{ab})$ and $\bm{\mathcal{Q}}_W:=\{\mathcal{Q}_W^{ab} \mid a\le b\}$, where $\mathcal{Q}_W^{ab}:=\{\mathcal{Q}_W^{ab}(\mathsf{v})\mid \mathsf{v}\in\mathsf{V}\}$. 

The ``energetic potential'' is defined as
\begin{talign}
    &e^{nF_E}
    :=\big(\int dyd\blambda\frac{\exp(-\frac{1}{2}\blambda^\intercal\bK^{-1}\blambda)}{((2\pi)^{s+1}\det \bK)^{1/2}}\prod_{a}^{0,s} P_{\rm out}(y\mid \lambda^a)\big)^n. \label{eq:FE}
    \end{talign}
   It takes this form due to our Gaussian assumption on the replicated post-activations and is thus easily computed, see App.~\ref{app:energetic_potential}. 
   
   The ``entropic potential'' $F_S$ taking into account the degeneracy of the order parameters is obtained by averaging delta functions fixing their definitions w.r.t. the ``microscopic degrees of freedom'' $(\bW^a)$. It can be written compactly using the following conditional law over the tensors $(\bS_2^a)$:
\begin{talign}    &P((\bS_2^a)\mid \bm{\mathcal{Q}}_W):=
    V_W^{kd}(\bm{\mathcal{Q}}_W)^{-1}\int \prod_a^{0,s}dP_W(\bW^a)\nonumber\\
    &\qquad\times\prod_{a\leq b}^{0,s}\prod_{\mathsf{v}\in \mathsf{V}}\prod_{i\in \mathcal{I}_\mathsf{v}}\delta(d\,\mathcal{Q}_W^{ab}(\mathsf{v})-{\bW^{a\intercal}_i\bW_i^{b}})\nonumber\\
    &\qquad \times\prod_{a}^{0,s}\delta(\bS^a_2-
    \bW^{a\intercal}(\bv)\bW^a/\sqrt{k}),\label{eq:truePS2}
\end{talign}
with the normalisation
\begin{talign*}
    &V_W^{kd}:=\int \prod_{a}dP_W(\bW^a)\prod_{a\le b,\mathsf{v},i\in \mathcal{I}_\mathsf{v}}\delta(d\,\mathcal{Q}_W^{ab}(\mathsf{v})-{\bW^{a\intercal}_i\bW_i^{b}}).
\end{talign*}
The entropy, which is the challenging term to compute, then reads   
\begin{talign*}
    e^{F_S}:= 
    V_W^{kd}(\bm{\mathcal{Q}}_W) \int dP((\bS_2^a)\mid\bm{\mathcal{Q}}_W) \prod\limits_{a\leq b}\limits^{0,s}\delta(d^2Q_2^{ab}-{\Tr \,\bS_{2}^{a} \bS_{2}^b}).
\end{talign*}

\subsection{Tackling the entropy: measure simplification by moment matching}
The delta functions above fixing $Q_2^{ab}$ induce \emph{quartic} constraints between the weights degrees of freedom $(W_{i\alpha}^a)$ instead of quadratic as in standard settings. A direct computation thus seems out of reach. However, we will exploit the fact that the constraints are quadratic in the matrices $(\bS_2^a)$. Consequently, shifting our focus towards $(\bS_2^a)$ as the basic degrees of freedom to integrate rather than $(W_{i\alpha}^a)$ will allow us to move forward by simplifying their measure \eqref{eq:truePS2}. Note that while $(W_{i\alpha}^a)$ are i.i.d. under their prior measure, $(\bS_2^a)$ have coupled entries, even for a fixed replica index $a$. This can be taken into account as follows.

Define $P_{S}$ as the probability density of a generalised Wishart random matrix, i.e., of $\tilde \bW^{\intercal}(\bv)\tilde\bW/\sqrt{k}$ where $\tilde \bW \in \mathbb{R}^{k\times d}$ is made of i.i.d. standard \emph{Gaussian} entries. The simplification we consider consists in replacing \eqref{eq:truePS2} by the effective measure
\begin{talign}
\label{eq:effectivePS2}
 \tilde P((\bS_2^a)\mid \bm{\mathcal{Q}}_W) := \frac1{\tilde V_W^{kd}} \prod\limits_a\limits^{0,s} P_{S}(\bS_2^a) \prod\limits_{a<b}\limits^{0,s} e^{\frac 12 \tau(\mathcal{Q}_W^{ab}) \Tr\,\bS^a_2\bS^b_2}
\end{talign}
where $\tilde V_W^{kd}=\tilde V_W^{kd}(\bm{\mathcal{Q}}_W)$ is the proper normalisation constant, and 
\begin{talign}
\tau(\mathcal{Q}_W^{ab}):= \text{mmse}_S^{-1}(1 - \EE_{v\sim P_v}[v^2 \mathcal{Q}^{ab}_W(v)^2]) . \label{eq:tauDef}  
\end{talign}

The rationale behind this choice goes as follows. The matrices $(\bS_2^a)$ are, under the measure \eqref{eq:truePS2}, $(i)$ generalised Wishart matrices, constructed from $(ii)$ non-Gaussian factors $(\bW^a)$, which $(iii)$ are coupled between different replicas, thus inducing a coupling among replicas $(\bS^a)$. The proposed simplified measure captures all three aspects while remaining tractable, as we explain now. The first assumption is that in the measure \eqref{eq:truePS2} the details of the (centred, unit variance) prior $P_W$ enter only through $\bm{\mathcal{Q}}_W$ at leading order. Due to the conditioning, we can thus relax it to Gaussian (with the same two first moments) by universality, as is often the case in random matrix theory. $P_W$ will instead explicitly enter in the entropy of $\bm{\mathcal{Q}}_W$ related to $V_W^{kd}$. Point $(ii)$ is thus taken care by the conditioning. Then, the generalised Wishart prior $P_{S}$ encodes $(i)$ and, finally, the exponential tilt in $\tilde P$ induces the replica couplings of point $(iii)$. It remains to capture the dependence of measure \eqref{eq:truePS2} on $\bm{\mathcal{Q}}_W$. This is done by realising that 
\begin{talign*}
\int dP((\bS_2^a)\mid \bm{\mathcal{Q}}_W) \frac 1{d^2}\Tr\bS_2^a \bS_2^b=    \EE_{v\sim P_v}[v^2 \mathcal{Q}_W^{ab}(v)^2]+\gamma \bar{v}^2.
\end{talign*}
It is shown in App.~\ref{app:second_moment_P(S2|QW)}. The Lagrange multiplier $\tau(\mathcal{Q}_W^{ab})$ to plug in $\tilde P$ enforcing this moment matching condition between true and simplified measures as $s\to0^+$ is \eqref{eq:tauDef}, see App.~\ref{app:entropic_contribution}. For completeness, we provide in App.~\ref{app:alternatives} alternatives to the simplification \eqref{eq:effectivePS2}, whose analysis are left for future~work.

\subsection{Final steps and spherical integration}
Combining all our findings, the average replicated partition function is simplified as
\begin{talign*}
& \EE\mathcal{Z}^s = \int d\bQ_2 d\bm{\mathcal{Q}}_W e^{nF_E+kd \ln V_W(\bm{\mathcal{Q}}_W)-kd\ln \tilde V_W(\bm{\mathcal{Q}}_W)}\\
&\,\,\,\times \!\prod\limits_a\limits^{0,s}\! P_{S}(\bS_2^a) \prod\limits_{a<b}\limits^{0,s} e^{\frac 12 \tau(\mathcal{Q}_W^{ab}) \Tr\,\bS^a_2\bS^b_2}\prod\limits_{a\le b}\limits^{0,s}\delta(d^2Q_2^{ab}\!-\!{\Tr \,\bS_{2}^{a} \bS_{2}^b}).
\end{talign*}
The equality should be interpreted as holding at leading exponential order $\exp (\Theta(n))$, assuming the validity of our previous measure simplification. All remaining steps but the last are standard:

$(i)$ Express the delta functions fixing $\bm{\mathcal{Q}}_W$ and $\bQ_2$ in exponential form using their Fourier representation; this introduces additional Fourier conjugate order parameters $\hat \bQ_2,\hat{\bm{\mathcal{Q}}}_W$ of same dimensions. 

$(ii)$ Once this is done, the terms coupling different replicas of $(\bW^a)$ or of $(\bS^a)$ are all quadratic. Using the Hubbard–Stratonovich transformation (i.e.,  $\E_{\bZ} \exp(\frac d 2 \Tr\,\bM \bZ) = \exp(\frac d 4 \Tr\,\bM^2)$
for a $d\times d$ symmetric matrix $\bM$ with $\bZ$ a standard GOE matrix) therefore allows us to linearise all replica-replica coupling terms, at the price of introducing new Gaussian fields interacting with all replicas. 

$(iii)$ After these manipulations, we identify at leading exponential order an effective action $\mathcal{S}$ depending on the order parameters only, which allows a saddle point integration w.r.t. them as $n\to \infty$: 
\begin{talign*}
    \lim \frac{1}{ns}\ln\E \mathcal{Z}^s  \!=\! \lim  \frac{1}{ns}\ln \int   d\bQ_2 d\hat\bQ_2 d\bm{\mathcal{Q}}_W d\hat{\bm{\mathcal{Q}}}_W  e^{n \mathcal{S}} \!= \! \frac 1s {\rm extr} \,\mathcal{S}.
\end{talign*}

$(iv)$ Next, the replica limit $s\to 0^+$ of the previously obtained expression has to be considered. To do so, we make a replica symmetric assumption, i.e., we consider that at the saddle point, all order parameters entering the action $\mathcal{S}$, and thus $K^{ab}$ too, take a simple form of the type $R^{ab} = r \delta_{ab} + q (1-\delta_{ab})$. Replica symmetry is rigorously known to be correct in general settings of Bayes-optimal learning and is thus justified here, see \cite{barbier2022strong,barbier2019adaptive}. 

$(v)$ After all these steps, the resulting expression still includes two high-dimensional integrals related to the $\bS_2$'s matrices. They can be recognised as corresponding to the free entropies associated with the Bayes-optimal denoising of a generalised Wishart matrix, as described just above Result~\ref{res:free_entropy}, for two different signal-to-noise ratios. The last step consists in dealing with these integrals using the HCIZ integral whose form is tractable in this case, see \cite{maillard2022perturbative,matrix_inference_Barbier}. These free entropies yield the two last terms $\iota(\,\cdot\,)$ in $f_{\rm RS}^{\alpha,\gamma}$, \eqref{eq:fRS}.

The complete derivation is in App.~\ref{app:replicas} and gives Result~\ref{res:free_entropy}. From the physical meaning of the order parameters, this analysis also yields the post-activations covariance $\bK$ and thus Result~\ref{res:gen_error}.

As a final remark, we emphasise a key difference between our approach and earlier works on extensive-rank systems. If, instead of taking the generalised Wishart $P_S$ as the base measure over the matrices $(\bS_2^a)$ in the simplified $\tilde P$ with moment matching, one takes a factorised Gaussian measure, thus entirely forgetting the dependencies among $\bS_2^a$ entries, this mimics the Sakata-Kabashima replica method \cite{sakata2013}. Our ansatz thus captures important correlations that were previously neglected in \cite{sakata2013,krzakala2013phase,kabashima2016phase,barbier2024phase} in the context of extensive-rank matrix inference. For completeness, we show in App.~\ref{app:alternatives} that our ansatz indeed greatly improves the prediction compared to these earlier approaches.

\section{Conclusion and perspectives}

We have provided an effective, quantitatively accurate, description of the optimal generalisation capability of a fully-trained two-layer neural network of extensive width with generic activation when the sample size scales with the number of parameters. This setting has resisted for a long time to mean-field approaches used, e.g., for committee machines \cite{barkai1992broken,engel1992storage,schwarze1992generalization,schwarze1993generalization,mato1992generalization,monasson1995weight,aubin2018committee,baldassi2019properties}.

A natural extension is to consider non Bayes-optimal models, e.g., trained 
through empirical risk minimisation to learn a mismatched target function. 
The formalism we provide here can be extended to these cases, by keeping track of additional order parameters. 
The extension to deeper architectures is also possible, in the vein of \cite{cui2023bayes,pacelli2023} who analysed the over-parametrised proportional regime. Accounting for structured inputs is another direction: data with a covariance~\citep{monasson1992,loureiro2021real}, mixture models~\citep{delgiudice1989,loureiro2021gmm}, hidden manifolds~\citep{goldt2020modeling}, object manifolds and simplexes~\citep{chung2018manifold,rotondo2020beyond}, etc.

Phase transitions in supervised learning are known in the statistical mechanics literature at least since~\cite{gyorgyi1990first}, when the theory was limited to linear models. It would be interesting to connect the picture we have drawn here with Grokking, a sudden drop in generalisation error occurring during the training of neural nets close to interpolation, see \cite{power2022grokking,rubin2024grokking}.

A more systematic analysis on the computational hardness of the problem (as carried out for multi-index models in~\cite{troiani2025fundamental}) is an important step towards a full characterisation of the class of target functions that are fundamentally hard to learn.

A key novelty of our approach is to blend matrix models and spin glass techniques in a unified formalism. A limitation is then linked to the restricted class of solvable matrix models (see \cite{kazakov2000solvable,anninos2020notes} for a list). Indeed, as explained in App.~\ref{app:alternatives}, possible improvements to our approach need additional finer order parameters than those appearing in Results~\ref{res:free_entropy}, \ref{res:gen_error} (at least for inhomogeneous readouts $\bv$). Taking them into account yields matrix models when computing their entropy which, to the best of our knowledge, are not currently solvable. We believe that obtaining asymptotically exact formulas for the log-partition function and generalisation error in the current setting and its relatives will require some major breakthrough in the field of multi-matrix models. This is an exciting direction to pursue at the crossroad of the fields of matrix models and high-dimensional inference and learning of extensive-rank matrices.

\section*{Software and data}

Experiments with ADAM/HMC were performed through standard implementations in PyTorch/TensorFlow/NumPyro; the Metropolis-Hastings and GAMP-RIE routines were coded from scratch (the latter was inspired by~\cite{maillard2024bayes}). GitHub repository to reproduce the results: \href{https://github.com/Minh-Toan/extensive-width-NN}{https://github.com/Minh-Toan/extensive-width-NN}

\section*{Acknowledgements}
J.B., F.C., M.-T.N. and M.P. were funded by the European Union (ERC, CHORAL, project number 101039794). Views and opinions expressed are however those of the authors only and do not necessarily reflect those of the European Union or the European Research Council. Neither the European Union nor the granting authority can be held responsible for them. M.P. thanks Vittorio Erba and Pietro Rotondo for interesting discussions and suggestions.

\bibliography{main.bib}
\bibliographystyle{arxiv2025}

\newpage
\appendix
\onecolumn

\section{Hermite basis and Mehler's formula\label{app:hermite}}
Recall the Hermite expansion of the activation:
\begin{equation}
    \sigma(x) = \sum_{\ell = 0}^{\infty} \frac{\mu_\ell}{\ell !}\He_\ell (x).
    \label{eq:sigma_hermite}
\end{equation}
We are expressing it on the basis of the probabilist's Hermite polynomials, generated through
\begin{equation}
    \He_\ell(z) = \frac{\diff^\ell}{{\diff t}^\ell} \exp\big(t z - t^2/2 \big)\big|_{t=0}.
    \label{eq:hermite_G}
\end{equation}
The Hermite basis has the property of being orthogonal with respect to the standard Gaussian measure, which is the distribution of the input data:
\begin{equation}
    \int Dz\, \He_k(z) \He_\ell(z) = \ell!\,\delta_{k\ell} ,
\end{equation}
where $Dz := \diff z \exp(-z^2/2)/\sqrt{2\pi} $. By orthogonality, the coefficients of the expansions can be obtained as
\begin{equation}
    \mu_\ell = \int Dz \He_\ell(z)\sigma(z).
\end{equation}
Moreover,
\begin{equation}
    \label{eq:avg_sigma}
    \E[\sigma(z)^2] = \int D z\, \sigma(z)^2 = \sum_{\ell=0}^{\infty} \frac{\mu_\ell^2}{\ell!}.
\end{equation}
These coefficients for some popular choices of $\sigma$ are reported in Table~\ref{tab:Hermite} for reference.
\begin{table}[b]
    \caption{First Hermite coefficients of some activation functions reported in the figues. $\theta$ is the Heaviside step function.}
    \label{tab:Hermite}
    \vskip 0.15in
    \begin{center}
    \begin{small}
    \begin{tabular}{l|ccccccc}
    \toprule
        $\sigma(z)$ & $\mu_0$ & $\mu_1$ & $\mu_2$ & $\mu_3$ & $\mu_4$ & $\cdots$ & $\E_z[\sigma(z)^2]$  \\
        \midrule
        $\relu(z) = z \theta(z)$ & $1/\sqrt{2 \pi }$ & $1/2$ & $1/\sqrt{2 \pi }$  & 0 & $-1/\sqrt{2 \pi }$& $\cdots$ & 1/2 \\
        $\elu(z) = z\theta(z) + (e^{z} - 1)\theta(-z) $ & 0.16052 & 0.76158 & 0.26158 & -0.13736 & -0.13736 & $\cdots$ & 0.64494 \\
        $\Tanh(2z)$ & 0 & 0.72948 & 0 & -0.61398 & 0 & $\cdots$ & 0.63526 \\
        \bottomrule
    \end{tabular}
    \end{small}
    \end{center}
    \vskip -0.1in
\end{table}
The Hermite basis can be generalised to an orthogonal basis with respect to the Gaussian measure with generic variance:
\begin{equation}
    \He_\ell^{[r]}(z) = \frac{\diff^\ell}{\diff t^\ell}  \exp\big(tz -  t^2 r/2\big)\big|_{t=0},
\end{equation}
so that, with $D_r z := \diff z \exp(-z^2/2r)/\sqrt{2\pi r} $, we have
\begin{equation}
    \int D_r z\, \He_k^{[r]}(z) \He_\ell^{[r]}(z) =  \ell! \,r^\ell\delta_{k\ell}.
\end{equation}

From Mehler's formula
\begin{equation}
    \frac{1}{2\pi\sqrt{r^2-q^2}} \exp\!\Big[-\frac{1}{2} (u,v) \begin{pmatrix}
        r & q \\ q & r
    \end{pmatrix}^{-1} \begin{pmatrix}
        u\\v
    \end{pmatrix} \Big] = \frac{e^{-\frac{u^2}{2r}}}{\sqrt{2\pi r}} \frac{e^{-\frac{v^2}{2r}}}{\sqrt{2\pi r}} \sum_{\ell = 0}^{+\infty} \frac{q^{\ell}}{\ell! r^{2\ell}} \He_{\ell}^{[r]}(u) \He_{\ell}^{[r]}(v),
    \label{eq:mehler}
\end{equation}
and by orthogonality of the Hermite basis, \eqref{eq:K} readily follows by noticing that the variables 
$(h_i^a = (\bW^a \bx)_i/\sqrt{d})_{i,a}
$
at given $(\bW^a)$ are Gaussian with covariances $\Omega^{ab}_{ij}=\bW_i^{a\intercal}\bW^b_j/d$, so that
\begin{equation}
    \EE [\sigma(h_{i}^a)\sigma(h_{j}^b)] = \sum_{\ell=0}^{\infty} \frac{(\mu_\ell^{[r]})^2}{\ell!r^{2\ell}} (\Omega_{ij}^{ab})^\ell,\qquad \mu_\ell^{[r]} = \int D_rz\, \He^{[r]}_\ell(z)\sigma(z).
\end{equation}
Moreover, as $r=\Omega^{aa}_{ii}$ converges for $d$ large to the variance of the prior of $\bW^0$ by Bayes-optimality, whenever $\Omega^{aa}_{ii} \to 1$ we can specialise this formula to the simpler case $r=1$ we reported in the main text.

\section{Nishimori identities}\label{app:nishiID}
The Nishimori identities are a very general set of symmetries arising in inference in the Bayes-optimal setting as a consequence of Bayes' rule. To introduce them, consider a test function $f$ of the teacher weights, collectively denoted by $\btheta^0$, of $s-1$ replicas of the student's weights $(\btheta^a)_{2\leq a\leq s}$ drawn conditionally i.i.d. from the posterior, and possibly also of the training set $\mathcal{D}$: $f(\btheta^0,\btheta^2,\dots,\btheta^s ;\mathcal{D})$. Then
\begin{align}
    \mathbb{E}_{{\boldsymbol{\theta}}^0,\mathcal{D}}\langle f(\btheta^0,\btheta^2,\dots,\btheta^s ;\mathcal{D})\rangle=
    \mathbb{E}_{{\boldsymbol{\theta}}^0,\mathcal{D}}\langle f(\btheta^1,\btheta^2,\dots,\btheta^s ;\mathcal{D})\rangle,
\end{align}where we have replaced the teacher's weights with another replica from the student. The proof is elementary, see e.g. \cite{barbier2019glm}.

The Nishimori identities have some consequences also on our replica symmetric ansatz for the free entropy. In particular, they constrain the values of the asymptotic mean of some order parameters. For instance
\begin{align}
    m_2= \lim
    \frac{1}{d^2}\EE_{\mathcal{D},\btheta^0}\langle\Tr[\bS_2^a\bS_2^0]\rangle=
    \lim\frac{1}{d^2}\EE_{\mathcal{D}}\langle\Tr[\bS_2^a\bS_2^b]\rangle= q_2,\quad \text{for }a\neq b.
\end{align}
Combined with the concentration of such order parameters, which can be proven in great generality in Bayes-optimal learning \cite{barbier2021overlap,barbier2022strong}, it fixes the values for some of them. For instance, we have that with high probability
\begin{align}
    \frac{1}{d^2}\Tr[(\bS_2^a)^2]\to r_2=\lim\frac{1}{d^2}\EE_{\mathcal{D}}\langle\Tr[(\bS_2^a)^2]\rangle=\lim\frac{1}{d^2}\EE_{\btheta^0}\Tr[(\bS_2^0)^2]=\rho_2=1+\gamma\bar v^2.
\end{align}When the values of some order parameters are determined by the Nishimori identities (and their concentration), as for those fixed to $r_2=\rho_2$, then the respective Fourier conjugates $\hat r_2,\hat \rho_2$ vanish (meaning that the desired constraints were already asymptotically enforced without the need of additional delta functions). This is because the configurations that make the order parameters take those values exponentially (in $n$) dominate the posterior measure, so these constraints are automatically imposed by the measure.

\section{Alternative representation for the optimal mean-square generalisation error\label{app:gen_err}}

We recall that $\btheta^0=(\bv^0,\bW^0)$ and similarly for $\btheta^1=\btheta,\btheta^2,\ldots$ which are replicas, i.e., conditionally i.i.d. samples from $dP(\bW,\bv\mid \mathcal{D})$ (the reasoning below applies whether $\bv$ is learnable or quenched, so in general we can consider a joint posterior over both). In this section we report the details on how to obtain Result~\ref{res:gen_error} and how to write the generalisation error defined in \eqref{eq:gen_error_def} in a form more convenient for numerical estimation. 

From its definition, 
the Bayes-optimal mean-square generalisation error can be recast as
\begin{align}
    \varepsilon^{\rm opt}= \EE_{\btheta^0,\bx_{\rm test} } \EE[y^2_{\rm test}\mid \lambda^0]
    -2\EE_{\btheta^0,\mathcal{D},\bx_{\rm test}} \EE[y_{\rm test}\mid \lambda^0]\langle\EE[y\mid \lambda]\rangle
    +  \EE_{\btheta^0,\mathcal{D},\bx_{\rm test}}\langle \EE[y\mid \lambda]\rangle^2,\label{eyStartingPoint}
\end{align}
where $\EE[y\mid \lambda]=\int dy\, y\, P_{\rm out}(y\mid \lambda)$, and  $\lambda^0$, $\lambda$ are the random variables (random due to the test input $\bx_{\rm test}$, drawn independently of the training data $\mathcal{D}$, and their respective weights $\btheta^0,\btheta$)
\begin{align}
    \lambda^0=\lambda(\btheta^0,\bx_{\rm test}) =\frac{\bv^{0\intercal}}{\sqrt{k}}\sigma\Big(\frac{\bW^0\bx_{\rm test}}{\sqrt{d}}\Big),\qquad
    \lambda=\lambda^1=\lambda(\btheta,\bx_{\rm test}) =\frac{\bv^{\intercal}}{\sqrt{k}}\sigma\Big(\frac{\bW\bx_{\rm test}}{\sqrt{d}}\Big).
\end{align}
Recall that the bracket $\langle\,\cdot\,\rangle$ is the average w.r.t. to the posterior and acts on $\btheta^1=\btheta,\btheta^2,\ldots$. Notice that the last term on the r.h.s.\ of \eqref{eyStartingPoint} can be rewritten as
\begin{align*}
    \EE_{\btheta^0,\mathcal{D},\bx_{\rm test}}\langle \EE[y\mid \lambda]\rangle^2=
    \EE_{\btheta^0,\mathcal{D},\bx_{\rm test}}\langle   \EE[y\mid \lambda^1]\EE[y\mid \lambda^{2}]\rangle,
\end{align*}with superscripts being replica indices, i.e., $\lambda^a:=\lambda(\btheta^a,\bx_{\rm test})$.

In order to show Result~\ref{res:gen_error} for a generic $P_{\rm out}$ we assume the joint Gaussianity of the variables $(\lambda^0,\lambda^1,\lambda^2,\ldots)$, with covariance given by $K^{ab}$ with $a,b\in\{0,1,2,\ldots\}$. Indeed, in the limit ``$\lim$'', our theory considers $(\lambda^a)_{a\ge 0}$ as jointly Gaussian under the randomness of a common input, here $\bx_{\rm test}$, conditionally on the weights $(\btheta^a)$. Their covariance depends on the weights $(\btheta^a)$ through various overlap order parameters introduced in the main. But in the large limit ``$\lim$'' these overlaps are assumed to concentrate under the quenched posterior average $\EE_{\btheta^0,\mathcal{D}}\langle\,\cdot\,\rangle$ towards non-random asymptotic values corresponding to the extremiser globally maximising the RS potential in Result~\ref{res:free_entropy}, with the overlaps entering $K^{ab}$ through \eqref{eq:K_RS}. This hypothesis is then confirmed by the excellent agreement between our theoretical predictions based on this assumption and the experimental results. This implies directly the equation for $\lim\,\varepsilon^{\mathcal{C},\mathsf{f}}$ in Result~\ref{res:gen_error} from definition \eqref{eq:Bayes_error_def}. For the special case of optimal mean-square generalisation error it yields
\begin{align}\label{eq:general_gen_error_Gaussian_eq}
    \lim \,\varepsilon^{\rm opt}=\EE_{\lambda^0}\EE[y^2_{\rm test}\mid \lambda^0]-2
    \EE_{\lambda^0,\lambda^1}\EE[y_{\rm test}\mid \lambda^0]\EE[y\mid \lambda^1]
    +\EE_{\lambda^1,\lambda^2}\EE[y\mid \lambda^1]\EE[y\mid \lambda^2]
\end{align}
where, in the replica symmetric ansatz, 
\begin{align}
    \EE[(\lambda^0)^2]=K^{00},\quad \EE[\lambda^0\lambda^1]= \EE[\lambda^0\lambda^2]=K^{01},\quad \EE[\lambda^1\lambda^2]=K^{12},\quad 
    \EE[(\lambda^1)^2]=\EE[(\lambda^2)^2]=K^{11}.
\end{align} For the dependence of the elements of $\bK$ on the overlaps under this ansatz we defer the reader to \eqref{eq:m_K-q_K-def}, \eqref{eq:rho_K-r_K-def}. In the Bayes-optimal setting, using the Nishimori identities (see App.~\ref{app:nishiID}), one can show that $K^{01}=K^{12}$ and $K^{00}=K^{11}$. Because of these identifications, we would additionally have
\begin{align}
    \EE_{\lambda^0,\lambda^1}\EE[y_{\rm test}\mid \lambda^0]\EE[y\mid \lambda^1]
    =\EE_{\lambda^1,\lambda^2}\EE[y\mid \lambda^1]\EE[y\mid \lambda^2].
\end{align}Plugging the above in \eqref{eq:general_gen_error_Gaussian_eq} yields \eqref{eq:gen_err_result}.

Let us now prove a formula for the optimal mean-square generalisation error written in terms of the overlaps that will be simpler to evaluate numerically, which holds for the special case of linear readout with Gaussian label noise $P_{\rm out}(y\mid \lambda)=\exp(-\frac1{2\Delta}(y-\lambda)^2)/\sqrt{2\pi \Delta}$. The following derivation is exact and does not require any Gaussianity assumption on the random variables $(\lambda^a)$. For the linear Gaussian channel the means verify $\EE[y\mid\lambda]=\lambda$ and $\EE[y^2\mid\lambda]=\lambda^2+\Delta$. Plugged in \eqref{eyStartingPoint} this yields
\begin{align}
    \varepsilon^{\rm opt}-\Delta=\EE_{\btheta^0,\bx_{\rm test} } \lambda^2_{\rm test}
    -2\EE_{\btheta^0,\mathcal{D},\bx_{\rm test}}  \lambda^0 \langle\lambda\rangle
    +  \EE_{\btheta^0,\mathcal{D},\bx_{\rm test}}\langle  \lambda^1\lambda^2\rangle,
\end{align}whence we clearly see that the generalisation error depends only on the covariance of $\lambda_{\rm test}(\btheta^0)=\lambda^0(\btheta^0),\lambda^1(\btheta^1),\lambda^2(\btheta^2)$ under the randomness of the shared input $\bx_{\rm test}$ at fixed weights, regardless of the validity of the Gaussian equivalence principle we assume in the replica computation. This covariance was already computed in \eqref{eq:K}; we recall it here for the reader's convenience
\begin{align}
    K(\btheta^a,\btheta^b) :=\EE\lambda^a\lambda^b=\sum_{\ell=1}^\infty\frac{\mu_\ell^2}{\ell!}\frac{1}{k}\sum_{i,j=1}^k v_i^a(\Omega^{ab}_{ij})^\ell v^b_j=
    \sum_{\ell=1}^\infty\frac{\mu_\ell^2}{\ell!}Q_\ell^{ab},
\end{align}
where $\Omega^{ab}_{ij}:= \bW_{i}^{a\intercal} \bW_{j}^b/d$, and $Q_\ell^{ab}$ as introduced in \eqref{eq:K} for $a,b=0,1,2$. We stress that $K(\btheta^a,\btheta^b)$ is not the limiting covariance $K^{ab}$ whose elements are in \eqref{eq:m_K-q_K-def}, \eqref{eq:rho_K-r_K-def}, but rather the finite size one. $K(\btheta^a,\btheta^b)$ provides us with an efficient way to compute the generalisation error numerically, that is through the formula
\begin{align}
    \varepsilon^{\rm opt}-\Delta&=\EE_{\btheta^0}K(\btheta^0,\btheta^0)-2\EE_{\btheta^0,\mathcal{D}}\langle K(\btheta^0,\btheta^1)\rangle+
    \EE_{\btheta^0,\mathcal{D}}\langle K(\btheta^1,\btheta^2)\rangle=\sum_{\ell=1}^\infty\frac{\mu_\ell^2}{\ell!}\EE_{\btheta^0,\mathcal{D}} \langle Q_{\ell}^{00}- 2Q_{\ell}^{01}+ Q^{12}_{\ell} \rangle.
    \label{eq:general_finitesize_gen_error_noapprox}
\end{align}
In the above, the posterior measure $\langle\,\cdot\,\rangle$ is taken care of by Monte Carlo sampling (when it equilibrates). In addition, as in the main text, we assume that in the large system limit the (numerically confirmed) identity \eqref{eq:Qell_NSB} holds. Putting all ingredients together we get
\begin{align}
    \varepsilon^{\rm opt}-\Delta
    =\EE_{\btheta^0,\mathcal{D}} &\Big\langle\mu_1^2(Q_1^{00}-2Q^{01}_1 +Q^{12}_1)+\frac{\mu_2^2}{2}(Q_2^{00}-2Q^{01}_2+Q^{12}_2)\nonumber\\ &+
    \EE_{v\sim P_v}v^2\big[g(\mathcal{Q}_W^{00}(v))- 2g(\mathcal{Q}_W^{01}(v))+g(\mathcal{Q}_W^{12} (v))\big] \Big\rangle.    \label{eq:general_finitesize_gene_error}
\end{align}
In the Bayes-optimal setting one can use again the Nishimori identities that imply $\EE_{\btheta^0,\mathcal{D}} \langle Q^{12}_{1}\rangle=\EE_{\btheta^0,\mathcal{D}} \langle Q^{01}_{1}\rangle$, and analogously $\EE_{\btheta^0,\mathcal{D}} \langle Q^{12}_{2}\rangle=\EE_{\btheta^0,\mathcal{D}} \langle Q^{01}_{2}\rangle$ and $\EE_{\btheta^0,\mathcal{D}} \langle g(\mathcal{Q}^{12}_{W}(v))\rangle=\EE_{\btheta^0,\mathcal{D}} \langle g(\mathcal{Q}^{01}_{W}(v))\rangle$. Inserting these identities in \eqref{eq:general_finitesize_gene_error} one gets
\begin{align}
    \varepsilon^{\rm opt}-\Delta&=\EE_{\btheta^0,\mathcal{D}} \Big\langle\mu_1^2(Q_1^{00}-Q^{01}_1 )+\frac{\mu_2^2}{2}(Q_2^{00}-Q^{01}_2)+
    \EE_{v\sim P_v}v^2\big[g(\mathcal{Q}_W^{00}(v))-g(\mathcal{Q}_W^{01} (v))\big] \Big\rangle.
    \label{eq:simple_gen_error_for_numerics}
\end{align}
This formula makes no assumption (other than~\eqref{eq:Qell_NSB}), including on the law of the $\lambda$'s. That it depends only on their covariance is simply a consequence of the quadratic nature of the mean-square generalisation error.

\begin{figure}[t!!!]
    \centering
    \includegraphics[width=0.5\linewidth]{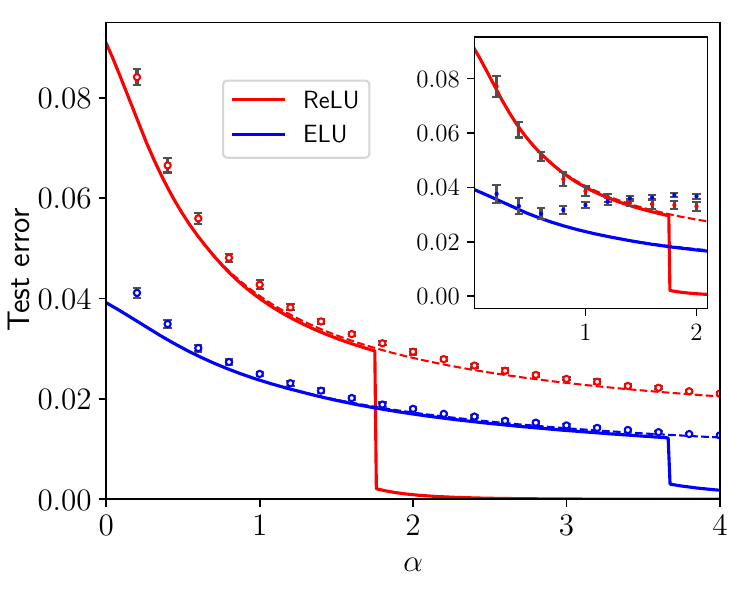}
    \caption{Theoretical prediction (solid curves) of the Bayes-optimal mean-square generalisation error for \emph{binary inner weights} and ReLU, eLU activations, with $\gamma = 0.5$, $d=150$, Gaussian label noise with $\Delta=0.1$, and fixed readouts $\bv  = \mathbf{1}$. Dashed lines are obtained from the solution of the fixed point equations \eqref{eq:NSB_equations_generic_ch} with all $\mathcal{Q}_W(\mathsf{v})=0$. Circles are the test error of GAMP-RIE \citep{maillard2024bayes} extended to generic activation.
    The MCMC points initialised uninformatively (inset) are obtained using \eqref{eq:general_finitesize_gen_error_noapprox}, to account for lack of equilibration due to glassiness, which prevents using \eqref{eq:simple_gen_error_for_numerics}. Even in the possibly glassy region, the GAMP-RIE attains the universal branch performance. 
    Data for GAMP-RIE and MCMC are averaged over 16 data instances, with error bars representing one standard deviation over instances.}
    \label{fig:GAMP-RIE_eLU-ReLU}
\end{figure}

\begin{remark}
    Note that the derivation up to \eqref{eq:general_finitesize_gen_error_noapprox} did not assume Bayes-optimality (while \eqref{eq:simple_gen_error_for_numerics} does). Therefore, one can consider it in cases where the true posterior average $\langle \,\cdot\,\rangle$ is replaced by one not verifying the Nishimori identities. This is the formula we use to compute the generalisation error of Monte Carlo-based estimators in the inset of Fig.~\ref{fig:GAMP-RIE_eLU-ReLU}. This is indeed needed to compute the generalisation in the glassy regime, where MCMC cannot equilibrate. 
\end{remark}

\begin{remark}\label{rem:Gibbs_error}
    Using the Nishimory identity of App.~\ref{app:nishiID} and again that, for the linear readout with Gaussian label noise $\EE[y\mid\lambda]=\lambda$ and $\EE[y^2\mid\lambda]=\lambda^2+\Delta$, it is easy to check that the so-called Gibbs error
    \begin{equation}
        \varepsilon^{\rm Gibbs} := \EE_{\bm{\theta}^0,\calD,\bx_{\rm test},y_{\rm test}}
        \big\langle (y_{\rm test} - \EE[y\mid \lambda_{\rm test}(\btheta)])^2 \big\rangle
        \label{eq:Gibbs_error}
    \end{equation}
    is related for this channel to the Bayes-optimal mean-square generalisation error through the identity
    \begin{equation}
        \varepsilon^{\rm Gibbs} - \Delta = 2(\varepsilon^{\rm opt} - \Delta).
        \label{eq:Gibbs_v_Bayes_error}
    \end{equation}
    We exploited this relationship together with the concentration of the Gibbs error w.r.t. the quenched posterior measure $\EE_{\btheta^0,\mathcal{D}}\langle\,\cdot\,\rangle$ when evaluating the numerical generalisation error of the Monte Carlo algorithms reported in the main text.
\end{remark}

\section{Details of the replica calculation\label{app:replicas}}

\subsection{Energetic potential}\label{app:energetic_potential}
The replicated energetic term under our Gaussian assumption on the joint law of the post-activations replicas is reported here for the reader's convenience:
\begin{equation}
    F_E = \ln \int dy\int d\blambda\frac{e^{-\frac{1}{2}\blambda^\intercal\bK^{-1}\blambda}}{\sqrt{(2\pi)^{s+1}\det \bK}} \prod_{a=0}^sP_{\rm out}(y\mid\lambda^a).
\end{equation}
After applying our ansatz \eqref{eq:Omega_ansatz} and using that $Q_1^{ab}=1$ in the quadratic-data regime, the covariance matrix $\bK$ in replica space defined in \eqref{eq:K} reads
\begin{align}
\label{eq:K_RS}
    K^{ab} &= \mu_1^2+ \frac{\mu_2^2}{2}Q^{ab}_2 + \EE_{v\sim P_v} v^2 g(\mathcal{Q}_W^{ab}(v)),
\end{align}
where the function
\begin{equation}
\label{eq:g_func}
    g(x) = \sum_{\ell = 3}^\infty \frac{\mu_{\ell}^2}{\ell !} x^\ell = 
     \E_{(y,z)|x} [\sigma(y)\sigma(z)] - \mu_0^2-\mu_1^2 x - \frac{\mu_2^2}{2}x^2, \qquad (y,z) \sim \calN\left((0,0),\begin{pmatrix}1&x\\x&1\end{pmatrix} \right).
\end{equation}
The energetic term $F_E$ is already expressed as a low-dimensional integral, but within the replica symmetric (RS) ansatz it simplifies considerably. Let us denote $\bm{\mathcal{Q}}_W(\mathsf{v})=(\mathcal{Q}_W^{ab}(\mathsf{v}))_{a,b=0}^s$. The RS ansatz amounts to assume that the saddle point solutions are dominated by order parameters of the form (below $\bm{1}_s$ and $\I_s$ are the all-ones vector and identity matrix of size $s$)
\begin{equation*}
     \bm{\mathcal{Q}}_W(\mathsf{v}) = \begin{pmatrix}
        \rho_W & m_W \bm{1}_s^\intercal\\
        m_W\bm{1}_s & (r_W - \mathcal{Q}_W) \I_s + \mathcal{Q}_W \bm{1}_s \bm{1}_s^\intercal 
    \end{pmatrix} \iff
    \hat{\bm{\mathcal{Q}}}_W(\mathsf{v}) = \begin{pmatrix}
        \hat{\rho}_W & -\hat{m}_W \bm{1}_s^\intercal\\
        -\hat{m}_W\bm{1}_s & (\hat{r}_W + \hat{\mathcal{Q}}_W) \I_s - \hat{\mathcal{Q}}_W \bm{1}_s \bm{1}_s^\intercal 
    \end{pmatrix},   
\end{equation*}
where all the above parameter $\rho_W=\rho_W(\mathsf{v}),\hat \rho_W,m_W,\ldots$ depend on $\mathsf{v}$, and similarly
\begin{equation*}
     \bQ_2 = \begin{pmatrix}
        \rho_2 & m_2 \bm{1}_s^\intercal\\
        m_2\bm{1}_s & (r_2 - q_2) \I_s + q_2 \bm{1}_s \bm{1}_s^\intercal 
    \end{pmatrix} \iff
    \hat{\bQ}_2 = \begin{pmatrix}
        \hat{\rho}_2 & -\hat{m}_2 \bm{1}_s^\intercal\\
        -\hat{m}_2\bm{1}_s & (\hat{r}_2 + \hat{q}_2) \I_s - \hat{q}_2 \bm{1}_s \bm{1}_s^\intercal 
    \end{pmatrix}, 
    \end{equation*}
where we reported the ansatz also for the Fourier conjugates\footnote{We are going to use repeatedly the Fourier representation of the delta function, namely $\delta(x)  = \frac1{2\pi} \int d\hat x \exp(i\hat x x)$.
Because the integrals we will end-up with will always be at some point evaluated by saddle point, implying a deformation of the integration contour in the complex plane, tracking the imaginary unit $i$ in the delta functions will be irrelevant. Similarly, the normalization $1/2\pi$ will always contribute to sub-leading terms in the integrals at hand. Therefore, we will allow ourselves to formally write $  \delta(x)  = \int d\hat x \exp( r\hat x x)$ for a convenient constant $r$, keeping in mind these considerations (again, as we evaluate the final integrals by saddle point, the choice of $r$ ends-up being irrelevant).} for future convenience, though not needed for the energetic potential. The RS ansatz, which is equivalent to an assumption of concentration of the order parameters in the high-dimensional limit, is known to be exact when analysing Bayes-optimal inference and learning, as in the present paper, see \cite{nishimori2001statistical,barbier2021overlap,barbier2022strong}. Under the RS ansatz $\bK$ acquires a similar form:
\begin{align}
    \bK=\begin{pmatrix}
        \rho_K & m_K \bm{1}_s^\intercal\\
        m_K\bm{1}_s & (r_K - q_K) \I_s + q_K \bm{1}_s \bm{1}_s^\intercal 
    \end{pmatrix}
\end{align}with 
\begin{align}
\label{eq:m_K-q_K-def}
    &m_K=\mu_1^2+ \frac{\mu_2^2}{2}m_2 + \EE_{v\sim P_v} v^2 g(m_W(v)),\quad &q_K=\mu_1^2+ \frac{\mu_2^2}{2}q_2 + \EE_{v\sim P_v} v^2 g(\mathcal{Q}_W(v)), \\
    \label{eq:rho_K-r_K-def}
    &\rho_K=\mu_1^2+ \frac{\mu_2^2}{2}\rho_2 + \EE_{v\sim P_v} v^2 g(\rho_W(v))
    ,\quad &r_K=\mu_1^2+ \frac{\mu_2^2}{2}r_2 + \EE_{v\sim P_v} v^2 g(r_W(v)).
\end{align}
In the RS ansatz it is thus possible to give a convenient low-dimensional representation of the multivariate Gaussian integral of $F_E$ in terms of white Gaussian random variables:
\begin{align}
&\lambda^a=\xi\sqrt{q_K}+u^a\sqrt{r_K-q_K}\quad \text{for }a=1,\dots,s,\qquad \lambda^0=\xi\sqrt{\frac{m_K^2}{q_K}}+u^0\sqrt{\rho_K-\frac{m_K^2}{q_K}}
\end{align}where $\xi,(u^a)_{a=0}^s$ are i.i.d. standard Gaussian variables. Then
\begin{align}
    F_E=\ln\int dy \,\EE_{\xi,u^0}P_{\rm out}\Big(y\mid \xi\sqrt{\frac{m_K^2}{q_K}}+u^0\sqrt{\rho_K-\frac{m_K^2}{q_K}}\Big)\prod_{a=1}^s\EE_{u^a}P_{\rm out}(y\mid \xi\sqrt{q_K}+u^a\sqrt{r_K-q_K}).
\end{align}
The last product over the replica index $a$ contains identical factors thanks to the RS ansatz. Therefore, by expanding in $s\to 0^+$ we get
\begin{align}\label{eq:energetic_pot_RS_general}
    F_E=s\int dy\, \EE_{\xi,u^0}P_{\rm out}\Big(y\mid \xi\sqrt{\frac{m_K^2}{q_K}}+u^0\sqrt{\rho_K-\frac{m_K^2}{q_K}}\Big)\ln \EE_{u}P_{\rm out}(y\mid \xi\sqrt{q_K}+u\sqrt{r_K-q_K})+O(s^2).
\end{align}
For the linear readout with Gaussian label noise $P_{\rm out}(y\mid\lambda)=\exp(-\frac1{2\Delta}(y-\lambda)^2)/\sqrt{2\pi\Delta}$ the above gives
\begin{align}\label{eq:F_E_GaussChann}
    F_E=-\frac{s}{2}\ln\big[2\pi(\Delta+r_K-q_K)\big]-\frac{s}{2}\frac{\Delta+\rho_K-2m_K+q_K}{\Delta+r_K-q_K}+O(s^2).
\end{align}
In the Bayes-optimal setting the Nishimori identities enforce 
\begin{align}\label{eq:nishi_r_2}
 &r_2=\rho_2 = \lim_{d\to\infty}\frac{1}{d^2}\EE\Tr[(\bS_2^0)^2]= 1 + \gamma \bar v^2 \quad \text{and} \quad  m_2=q_2,\\ 
 &r_W(\mathsf{v})=\rho_W(\mathsf{v})=1 \quad \text{and} 
\quad m_W(\mathsf{v})=\mathcal{Q}_W(\mathsf{v})\ \forall \ \mathsf{v}\in\mathsf{V},\label{eq:nishi_q2}
\end{align}
which implies also that
\begin{align}
\label{eq:covariance_vars}
r_K=\rho_K=\mu_1^2+\frac12r_2\mu_2^2 +g(1) ,\,\quad 
 m_K=q_K.
\end{align}
Therefore the above simplifies to
\begin{align}\label{eq:energetic_pot_RS_BayesOpt}
    F_E&=s\int dy\, \EE_{\xi,u^0}P_{\rm out}(y\mid \xi\sqrt{q_K}+u^0\sqrt{r_K-q_K})\ln \EE_{u}P_{\rm out}(y\mid \xi\sqrt{q_K}+u\sqrt{r_K-q_K})+O(s^2)\\
    &=:s \, \psi_{P_{\rm{out}}}(q_K(q_2,\mathcal{Q}_W);r_K)+O(s^2).\label{eq:defPsiPout}
\end{align}
Notice that the energetic contribution to the free entropy has the same form as in the generalised linear model \cite{barbier2019glm}. For our running example of linear readout with Gaussian noise 
the function $\psi_{P_{\rm out}}$ reduces to
\begin{align}\label{eq:F_E_BayesOpt_GaussChann}
     \psi_{P_{\rm{out}}}(q_K(q_2,\mathcal{Q}_W);r_K)=-\frac{1}{2}\ln\big[2\pi e(\Delta+r_K-q_K)\big].
\end{align}
In what follows we shall restrict ourselves only to the replica symmetric ansatz, in the Bayes-optimal setting. Therefore, identifications as the ones in \eqref{eq:nishi_r_2}, \eqref{eq:nishi_q2} are assumed.

\subsection{Second moment of \texorpdfstring{$P((\bS_2^a)\mid\bm{\mathcal{Q}}_W)$}{P(S|QW)}\label{app:second_moment_P(S2|QW)}}
For the reader's convenience we report here the measure
\begin{align}
    P((\bS_2^a)\mid\bm{\mathcal{Q}}_W ) &=
    V_W^{kd}(\bm{\mathcal{Q}}_W)^{-1}\int \prod_{a}^{0,s}dP_W(\bW^a) \delta(\bS^a_2-
    \bW^{a\intercal}(\bv )\bW^a/\sqrt{k})
    \prod_{a\leq b}^{0,s}\prod_{\mathsf{v}\in \mathsf{V}} \prod_{i\in\mathcal{I}_{\mathsf{v}}}\delta({d}\,\mathcal{Q}_W^{ab}(\mathsf{v})-\bW^{a\intercal}_i\bW_i^{b}).
    \label{eq:truePS2_app}
\end{align}Recall $\mathsf{V}$ is the support of $P_v$ (assumed discrete for the moment). Recall also that we have quenched the readout weights to the ground truth. Indeed, as discussed in the main, considering them learnable or fixed to the truth does not change the leading order of the information-theoretic quantities. 

In this measure, one can compute the asymptotic of its second moment
\begin{align}
    \int dP((\bS_2^a)\mid \bm{\mathcal{Q}}_W ) \frac{1}{d^2}\Tr\,\bS_2^a\bS_2^{b}&=
    V_W^{kd}(\bm{\mathcal{Q}}_W)^{-1}\int \prod_{a}^{0,s}dP_W(\bW^a)\frac{1}{kd^2}\Tr[\bW^{a\intercal}(\bv )\bW^{a}\bW^{b\intercal}(\bv )\bW^{b}]\nonumber\\
    &\qquad\qquad \times\prod_{a\leq b}^{0,s}\prod_{\mathsf{v}\in \mathsf{V}} \prod_{i\in\mathcal{I}_v}\delta({d}\,\mathcal{Q}_W^{ab}(\mathsf{v})-\bW^{a\intercal}_i\bW_i^{b}).
\end{align}
The measure is coupled only through the latter $\delta$'s. We can decouple the measure at the cost of introducing Fourier conjugates whose values will then be fixed by a saddle point computation. The second moment computed will not affect the saddle point, hence it is sufficient to determine the value of the Fourier conjugates through the computation of $V_W^{kd}(\bm{\mathcal{Q}}_W)$, which rewrites as
\begin{align}
    V_W^{kd}(\bm{\mathcal{Q}}_W)&=\int \prod_{a}^{0,s}dP_W(\bW^a)\prod_{a\leq b}^{0,s}\prod_{\mathsf{v}\in \mathsf{V}} \prod_{i\in\mathcal{I}_{\mathsf{v}}}d\hat B^{ab}_{i}(\mathsf{v})\exp\big[-\hat B^{ab}_{i}(\mathsf{v})
    ({d}\,\mathcal{Q}_W^{ab}(\mathsf{v})-\bW^{a\intercal}_i\bW_i^{b})
    \big]\nonumber\\
    &\approx\prod_{\mathsf{v}\in \mathsf{V}}\prod_{i\in\mathcal{I}_{\mathsf{v}}}\exp\Big(d\,\extr_{(\hat B^{ab}_{i}(\mathsf{v}))}\Big[-\sum_{a\le b,0}^s\hat B^{ab}_{i}(\mathsf{v}) \mathcal{Q}_W^{ab}(\mathsf{v})+\ln\int \prod_{a=0}^sdP_W(w_a)e^{\sum_{a\leq b,0}^s \hat B_i^{ab}(\mathsf{v})w_a w_b}\Big]\Big).
\end{align}In the last line we have used saddle point integration over $\hat B^{ab}_{i}(\mathsf{v})$ and the approximate equality is up to a multiplicative $\exp(o(n))$ constant. 
From the above, it is clear that the stationary $\hat B^{ab}_i(\mathsf{v})$ are such that
\begin{align}
    \mathcal{Q}_W^{ab}(\mathsf{v})=\frac{\int \prod_{r=0}^sdP_W(w_r) w_aw_b \prod_{r\leq t,0}^s e^{ \hat B_i^{rt}(\mathsf{v}) w_{r}w_{t}}}{
    \int \prod_{r=0}^sdP_W(w_r) \prod_{r\leq t,0}^s e^{ \hat B_i^{rt}(\mathsf{v}) w_{r}w_{t}}
    }=:\langle w_aw_b\rangle_{\hat \bB(\mathsf{v})}.
\end{align}Hence $\hat B_i^{ab}(\mathsf{v})=\hat B^{ab}(\mathsf{v})$ are homogeneous. Using these notations, the asymptotic trace moment of the $\bS_2$'s at leading order becomes
\begin{align}
    \int dP((\bS_2^a)\mid \bm{\mathcal{Q}}_W ) &\frac{1}{d^2}\Tr\,\bS_2^a\bS_2^{b}=\frac{1}{kd^2}\sum_{i,l=1}^k\sum_{j,p=1}^d \langle W_{ij}^{a}v_iW_{ip}^{a}W_{lj}^{b}v_l
    W_{lp}^{b}\rangle_{\{\hat \bB(\mathsf{v})\}_{\mathsf{v}\in\mathsf{V}}}\nonumber\\
    &=
    \frac{1}{k}\sum_{\mathsf{v}\in\mathsf{V}} \mathsf{v}^2 \sum_{i\in\mathcal{I}_{\mathsf{v}}}\Big\langle \Big(\frac{1}{d}\sum_{j=1}^d W_{ij}^{a}W_{ij}^{b} \Big)^2\Big\rangle_{\hat \bB(\mathsf{v})}+
    \frac{1}{k} \sum_{j=1}^d \Big\langle 
    \sum_{i=1}^k \frac{v_i(W_{ij}^{a})^2}{d} \sum_{l\neq i,1}^k \frac{v_l(W_{lj}^{b})^2}{d}
    \Big\rangle_{\hat \bB(\mathsf{v})}.
\end{align}We have used the fact that $\smash{\langle\,\cdot\,\rangle_{\hat \bB(\mathsf{v})}}$ is symmetric if the prior $P_W$ is, thus forcing us to match $j$ with $p$ if $i\neq l$. Considering that by the Nishimori identities $\mathcal{Q}_W^{aa}(\mathsf{v})=1$, it implies $\hat B^{aa}(\mathsf{v})=0$ for any $a=0,1,\dots,s$ and $\mathsf{v}\in\mathsf{V}$. Furthermore, the measure $\langle\,\cdot\,\rangle_{\hat \bB(\mathsf{v})}$ is completely factorised over neuron and input indices. Hence every normalised sum can be assumed to concentrate to its expectation by the law of large numbers. Specifically, we can write that with high probability as $d,k\to \infty$,
\begin{align}
  \frac{1}{d}\sum_{j=1}^d W_{ij}^{a}W_{ij}^{b} \xrightarrow{}\mathcal{Q}_W^{ab}(\mathsf{v}) \ \forall \ i\in \mathcal{I}_\mathsf{v},\qquad \frac{1}{k} \sum_{\mathsf{v},\mathsf{v}'\in \mathsf{V}} \mathsf{v}\mathsf{v}' \sum_{j=1}^d  
    \sum_{i\in \mathcal{I}_{\mathsf{v}}} \frac{(W_{ij}^{a})^2}{d} \sum_{l\in \mathcal{I}_{\mathsf{v}'},l\neq i} \frac{(W_{lj}^{b})^2}{d}
    \approx \gamma  \sum_{\mathsf{v},\mathsf{v}'\in \mathsf{V}} \frac{|\mathcal{I}_{\mathsf{v}}||\mathcal{I}_{\mathsf{v}'}|}{k^2} \mathsf{v}\mathsf{v}'\to \gamma \bar v^2,
\end{align}
where we used $|\mathcal{I}_\mathsf{v}|/k\to P_v(\mathsf{v})$ as $k$ diverges. Consequently, the second moment at leading order appears as claimed:
\begin{align}
\label{eq:truePS2_second_moment}
    \int dP((\bS_2^a)\mid \bm{\mathcal{Q}}_W) &\frac{1}{d^2}\Tr\,\bS_2^a\bS_2^{b}=\sum_{\mathsf{v}\in\mathsf{V}}P_v(\mathsf{v}) \mathsf{v}^2 \mathcal{Q}_W^{ab}(\mathsf{v})^2+\gamma\bar v^2=\EE_{v\sim P_v} v^2 \mathcal{Q}_W^{ab}(v)^2+\gamma\bar v^2.
\end{align} 

Notice that the effective law $\tilde P((\bS_2^a)\mid \bm{\mathcal{Q}}_W) $ in \eqref{eq:effectivePS2} is the least restrictive choice among the Wishart-type distributions with a trace moment fixed precisely to the one above. In more specific terms, it is the solution of the following maximum entropy problem:
\begin{align}
    \inf_{P,\tau}\Big\{D_{\rm KL}(P\,\|\,P_S^{\otimes s+1})+\sum_{a\leq b,0}^s\tau^{ab} \Big(\EE_P\frac{1}{d^2}\Tr\,\bS_2^a\bS_2^b-\gamma\bar v^2-\EE_{v\sim P_v} v^2 \mathcal{Q}_W^{ab}(v)^2\Big)\Big\},
\end{align}where $P_S$ is a generalised Wishart distribution (as defined above \eqref{eq:effectivePS2}), and $P$ is in the space of joint probability distributions over $s+1$ symmetric matrices of dimension $d\times d$. The rationale behind the choice of $P_S$ as a base measure is that, in absence of any other information, a statistician can always use a generalised Wishart measure for the $\bS_2$'s if they assume universality in the law of the inner weights. This ansatz would yield the theory of \cite{maillard2024bayes}, which still describes a non-trivial performance, achieved by the adaptation of GAMP-RIE of Appendix~\ref{app:GAMP}.

Note that if $a=b$ then, by \eqref{eq:nishi_r_2}, the second moment above matches precisely $r_2=1+\gamma\bar v^2$. This entails directly $\tau^{aa}=0$, as the generalised Wishart prior $P_S$ already imposes this constraint.

\subsection{Entropic potential}\label{app:entropic_contribution}
We now use the results from the previous section to compute the entropic contribution $F_S$ to the free entropy:
\begin{align}\label{eq:eFS}
    e^{F_S}:= 
    V_W^{kd}(\bm{\mathcal{Q}}_W) \int dP((\bS_2^a)\mid\bm{\mathcal{Q}}_W) \prod_{a\leq b}^{0,s} \delta(d^2Q_2^{ab}-{\Tr \,\bS_{2}^{a} \bS_{2}^b}).
\end{align}The factor $V_W^{kd}(\bm{\mathcal{Q}}_W)$ was already treated in the previous section. However, here it will contribute as a tilt of the overall entropic contribution, and the Fourier conjugates $\hat{\mathcal{Q}}_W^{ab}(\mathsf{v})$ will appear in the final variational principle. 

Let us now proceed with the relaxation of the measure $P((\bS_2^a)\mid \bm{\mathcal{Q}}_W)$ by replacing it with $\tilde P((\bS_2^a)\mid \bm{\mathcal{Q}}_W)$ given by \eqref{eq:effectivePS2}:
\begin{align}\label{eq:F_S_step1_appendix}
    e^{F_S}= 
    V_W^{kd}(\bm{\mathcal{Q}}_W) \int d\hat{\bQ}_2 \exp\Big(-\frac{d^2}{2}\sum_{a\leq b,0}^s\hat Q^{ab}_2Q^{ab}_2\Big)\frac{1}{\tilde V^{kd}_W(\bm{\mathcal{Q}}_W)}\int \prod_{a=0}^s dP_S(\bS_2^a) \exp\Big(\sum_{a\leq b,0}^s\frac{\tau_{ab}+\hat Q_2^{ab}}{2}\Tr\,\bS_2^a\bS_2^b\Big)
\end{align}where we have introduced another set of Fourier conjugates $\hat\bQ_2$ for $\bQ_2$. As usual, the Nishimori identities impose $Q_2^{aa}=r_2=1+\gamma\bar v^2$ without the need of any Fourier conjugate. Hence, similarly to $\tau^{aa}$, $\hat Q_2^{aa}=0$ too. Furthermore, in the hypothesis of replica symmetry, we set  $\tau^{ab}=\tau$ and $\hat Q_2^{ab}=\hat q_2$ for all $0\leq a<b\leq s$.

Then, when the number of replicas $s$ tends to $0^+$, we can recognise the free entropy of a matrix denoising problem. More specifically, using the Hubbard–Stratonovich transformation (i.e.,  $\E_{\bZ} \exp(\frac d 2 \Tr\,\bM \bZ) = \exp(\frac d 4 \Tr\,\bM^2)$
for a $d\times d$ symmetric matrix $\bM$ with $\bZ$ a standard GOE matrix) we get
\begin{align}
\label{eq:J_replicas}
    J_n(\tau,\hat q_2)&:=\lim_{s\to 0^+}\frac{1}{ns}\ln\int \prod_{a=0}^s dP_S(\bS_2^a) \exp\Big(\frac{\tau+\hat q_2}{2}\sum_{a< b,0}^s\Tr\,\bS_2^a\bS_2^b\Big)\nonumber\\
    &=\frac{1}{n}\EE\ln\int dP_S(\bS_2)\exp\frac{1}{2}\Tr\Big(\sqrt{\tau+\hat q_2}\bY\bS_2-(\tau+\hat q_2)\frac{\bS_2^2}{2}\Big),
\end{align}where $\bY=\bY(\tau+\hat q_2)=\sqrt{\tau+\hat q_2}\bS_2^0+\bxi$ with $\bxi/\sqrt{d}$ a standard GOE matrix, and the outer expectation is w.r.t.\ $\bY$ (or $\bS^0,\bxi$). Thanks to the fact that the base measure $P_S$ is rotationally invariant, the above can be solved exactly in the limit $n\to\infty,\,n/d^2\to\alpha$ (see e.g. \cite{matrix_inference_Barbier}):
\begin{align}
    J(\tau,\hat q_2)=\lim  J_n(\tau,\hat q_2)=\frac{1}{\alpha}\Big(\frac{(\tau+\hat q_2)r_2}{4}-\iota(\tau+\hat q_2)\Big), \quad \text{with} \quad \iota(\eta):=\frac{1}{8}+\frac{1}{2}\Sigma(\mu_{\bY(\eta)}).
\end{align}Here $\iota(\eta)=\lim I(\bY(\eta);\bS^0_2)/d^2$ is the limiting mutual information between data $\bY(\eta)$ and signal $\bS^0_2$ for the channel $\bY(\eta)=\sqrt{\eta}\bS^0_2+\bxi$, the measure $\mu_{\bY(\eta)}$ is the asymptotic spectral law of the rescaled observation matrix $\bY(\eta)/\sqrt{d}$, and $\Sigma(\mu):=\int \ln|x-y|d\mu(x)d\mu(y)$. Using free probability, the law $\mu_{\bY(\eta)}$ can be obtained as the free convolution of a generalised Marchenko-Pastur distribution (the asymptotic spectral law of $\bS^0_2$, which is a generalised Wishart random matrix) and the semicircular distribution (the asymptotic spectral law of $\bxi$), see \cite{potters2020first}. We provide the code to obtain this distribution numerically in the attached repository. The function ${\rm mmse}_S(\eta)$ is obtained through a derivative of $\iota$, using the so-called I-MMSE relation \cite{guo2005mutual,matrix_inference_Barbier}:
\begin{align}
\label{eq:def_mmse_function}
4\frac{d}{d\eta}\iota(\eta)={\rm mmse}_S(\eta)=\frac{1}{\eta}\Big(1-\frac{4\pi^2}{3}\int \mu^3_{\bY(\eta)}(y)dy\Big).
\end{align}
The normalisation $\frac1{ns}\ln\tilde V_W^{kd}(\bm{\mathcal{Q}}_W)$ in the limit $n\to \infty,s\to0^+$ can be simply computed as $J(\tau,0)$.

For the other normalisation, following the same steps as in the previous section, we can simplify $V^{kd}_W(\bm{\mathcal{Q}}_{W})$ as follows:
\begin{align}
    \frac{1}{ns}\ln V_W^{kd}(\bm{\mathcal{Q}}_W)\approx \frac{\gamma}{\alpha s} \sum_{\mathsf{v}\in\mathsf{V}}\frac{1}{k}\sum_{i\in\mathcal{I}_{\mathsf{v}}}
    {\rm extr}\Big[-\sum_{a\leq b,0}^s\hat {\mathcal{Q}}^{ab}_{W,i}(\mathsf{v}) \mathcal{Q}^{ab}_W(\mathsf{v})+
    \ln\int \prod_{a=0}^sdP_W(w_a)e^{\sum_{a\leq b,0}^s\hat {\mathcal{Q}}^{ab}_{W,i}(\mathsf{v})w_aw_b}\Big],
\end{align} as $n$ grows,
where extremisation is w.r.t.\ the hatted variables only. As in the previous section, $\hat {\mathcal{Q}}^{ab}_{W,i}(\mathsf{v})$ is homogeneous over $i\in\mathcal{I}_\mathsf{v}$ for a given $\mathsf{v}$. Furthermore, thanks to the Nishimori identities we have that at the saddle point $\hat{\mathcal{Q}}_W^{aa}(\mathsf{v})=0$ and ${\mathcal{Q}}_W^{aa}(\mathsf{v})=1+\gamma\bar v^2$. This, together with standard steps and the RS ansatz, allows to write the $d\to\infty,s\to0^+$ limit of the above as
\begin{align}
    \lim_{s\to0^+}\lim\frac{1}{ns}
    \ln V_W^{kd}(\bm{\mathcal{Q}}_W)= \frac{\gamma}{\alpha} \EE_{v\sim P_v}
    {\rm extr}\Big[-\frac{\hat{\mathcal{Q}}_W(v) \mathcal{Q}_W(v)}{2} +
    \psi_{P_W}(\hat{\mathcal{Q}}_W(v))\Big]
\end{align}with $\psi_{P_W}(\,\cdot\,)$ as in the main. Gathering all these results yields directly
\begin{align}
    \lim_{s\to0^+}\lim\frac{F_S}{ns}= {\rm extr}\Big\{&\frac{\hat q_2(r_2-q_2)}{4\alpha}-\frac{1}{\alpha}\big[\iota(\tau+\hat q_2)-\iota(\tau)\big] +\frac{\gamma}{\alpha}\EE_{v\sim P_v}\Big[\psi_{P_W}(\hat{\mathcal{Q}}_W(v))-\frac{\hat{\mathcal{Q}}_W(v) \mathcal{Q}_W(v)}{2}\Big]
    \Big\}.
\end{align} Extremisation is w.r.t.\ $ \hat q_2,\hat{\mathcal{Q}}_W$. $\tau$ has to be intended as a function of $\mathcal{Q}_W=\{ \calQ_W(\mathsf{v})\mid \mathsf{v}\in\mathsf{V}\}$ through the moment matching condition:
\begin{align}
    4\alpha\, \partial_\tau J(\tau,0)=r_2- 4\iota'(\tau)=\EE_{v\sim P_v} v^2 \mathcal{Q}_W(v)^2 +\gamma\bar v^2,
\end{align}which is the $s\to 0^+$ limit of the moment matching condition between $P((\bS_2^a)\mid \bm{\mathcal{Q}}_W)$ and $\tilde P((\bS_2^a)\mid \bm{\mathcal{Q}}_W)$. Simplifying using the value of $r_2=1+\gamma\bar v^2$ according to the Nishimori identities, and using the I-MMSE relation between $\iota(\tau)$ and ${\rm mmse}_S(\tau)$, we get
\begin{align}
\label{eq:moment_matching}
    {\rm mmse}_S(\tau)=1-\EE_{v\sim P_v} v^2 \mathcal{Q}_W(v)^2\quad\iff\quad \tau={\rm mmse}_S^{-1}\big(
    1-\EE_{v\sim P_v} v^2 \mathcal{Q}_W(v)^2
    \big).
\end{align} Since ${\rm mmse}_S$ is a monotonic decreasing function of its argument (and thus invertible), the above always has a solution, and it is unique for a given collection $\mathcal{Q}_W$.

\subsection{RS free entropy and saddle point equations}
Putting the energetic and entropic contributions together we obtain the variational replica symmetric free entropy potential:
\begin{align}
     f^{\alpha,\gamma}_{\rm RS}&:= \psi_{P_{\text{out}}}(q_K(q_2,\mathcal{Q}_W);r_K) + \frac{1}{4\alpha}(1 + \gamma \bar{v}^2-q_2) \Hat{q}_2 + \frac{\gamma}{\alpha}\EE_{ v\sim P_v}\big[
     \psi_{P_W}(\Hat{\mathcal{Q}}_W(v))-\frac{1}{2}\mathcal{Q}_W(v) \Hat{\mathcal{Q}}_W(v) \big]\nonumber \\
     &\qquad +\frac{1}\alpha\big[\iota(\tau(\mathcal{Q}_W)) - \iota(\hat q_2 + \tau(\mathcal{Q}_W))\big],
     \label{eq:free_ent_appendix}
\end{align}
which is then extremised w.r.t. $\{\hat{\mathcal{Q}}_W(\mathsf{v}),\mathcal{Q}_W(\mathsf{v})\mid \mathsf{v}\in\mathsf{V}\},\hat q_2,q_2$ while $\tau$ is a function of ${\mathcal{Q}}_W$ through the moment matching condition \eqref{eq:moment_matching}.  
The saddle point equations are then 
\begin{equation}
\label{eq:NSB_equations_generic_ch}
\begin{sqcases}
& {\mathcal{Q}}_{W}(\mathsf{v}) = \EE_{w^0,\xi} [ w^0 \thav{w}_{\Hat{\mathcal{Q}}_{W}(\mathsf{v})} ], \\
& \hat {\mathcal{Q}}_{W}(\mathsf{v}) = \frac{1}{2\gamma}(q_2 - \gamma\bar v^2-\EE_{v\sim P_v}v^{2} \mathcal{Q}_W(v)^2)\partial_{{\mathcal{Q}}_{W}(\mathsf{v})} \tau(\mathcal{Q}_W) + 2\frac{\alpha}{\gamma} \partial_{{\mathcal{Q}}_{W}(\mathsf{v})} \psi_{P_{\text{out}}}(q_K(q_2,\mathcal{Q}_W);r_K), \\
& q_2= r_2-\frac{1}{\hat q_2 + \tau(\mathcal{Q}_W)}(1-\frac{4\pi^2}{3}\int \mu^3_{\bY(\hat q_2 + \tau(\mathcal{Q}_W))}(y)dy) , \\
& \hat q_2 = 4\alpha \,\partial_{q_2} \psi_{P_{\text{out}}}(q_K(q_2,\mathcal{Q}_W);r_K) ,
\end{sqcases}   
\end{equation}
where, letting i.i.d. $w^0,\xi\sim \mathcal{N}(0,1)$, we define the measure
\begin{align}\label{eq:bracket_FP_equations}
    \langle\,\cdot\,\rangle_{x}=\langle\,\cdot\,\rangle_{x}(w^0,\xi):=
    \frac{\int dP_W(w)(\,\cdot\,)e^{
    (\sqrt{x}\xi+x w^0)w-\frac{1}{2}xw^2
    }}{\int dP_W(w)e^{
    (\sqrt{x}\xi+xw^0)w-\frac{1}{2}xw^2
    }}   . 
\end{align}

All the above formulae are easily specialised for the linear readout with Gaussian label noise using \eqref{eq:F_E_BayesOpt_GaussChann}. We report here the saddle point equations in this case (recalling that $g$ is defined in \eqref{eq:g_func}):
\begin{equation}
\begin{sqcases}
\label{NSB_equations_gaussian_ch}
& {\mathcal{Q}}_{W}(\mathsf{v}) = \EE_{w^0,\xi} [ w^0 \thav{w}_{\Hat{\mathcal{Q}}_{W}(v)} ], \\
& \hat {\mathcal{Q}}_{W}(\mathsf{v}) = \frac{1}{2\gamma}(q_2 -\gamma\bar v^2- \EE_{v\sim P_v}v^2 \mathcal{Q}_W(v)^2)\partial_{{\mathcal{Q}}_{W}(\mathsf{v})} \tau(\mathcal{Q}_W) + \frac{\alpha}{\gamma} \frac{ \mathsf{v}^2 \, g'(\mathcal{Q}_W(\mathsf{v}))}{\Delta + \frac{1}{2}\mu_2^2(r_2-q_2) + g(1) - \EE_{v\sim P_v} {v}^2 g(\mathcal{Q}_W(v))}, \\
& q_2= r_2-\frac{1}{\hat q_2 + \tau}(1-\frac{4\pi^2}{3}\int \mu^3_{\bY(\hat q_2 + \tau(\mathcal{Q}_W))}(y)dy) ,\\
& \hat q_2 =  \frac{\alpha\mu_2^2}{\Delta +\frac{1}{2}\mu_2^2(r_2-q_2) + g(1) - \EE_{v\sim P_v}  v^2g(\mathcal{Q}_W(v))}.
\end{sqcases}    
\end{equation}

If one assumes that the overlaps appearing in \eqref{eq:simple_gen_error_for_numerics} are self-averaging around the values that solve the saddle point equations (and maximise the RS potential), that is $Q^{00}_1,Q_1^{01}\to1$ (as assumed in this scaling), $Q_2^{00}\to r_2, Q_2^{01}\to q_2^*$, and ${\mathcal{Q}}_{W}^{00}(\mathsf{v})\to1,{\mathcal{Q}}_{W}^{01}(\mathsf{v})\to {\mathcal{Q}}_{W}^*(\mathsf{v})$, then the limiting Bayes-optimal mean-square generalisation error for the linear readout with Gaussian noise case appears as
\begin{equation}
\begin{aligned}
    \varepsilon^{\rm opt}-\Delta = r_K -q_K^*=\frac{\mu_2^2}{2}(r_2-q_2^*) +  g(1) - \EE_{v\sim P_v} v^2g(\mathcal{Q}^*_W(v)) .
\end{aligned}    \label{eq:Gaussian_output_channel_generror}
\end{equation}
This is the formula used to evaluate the theoretical Bayes-optimal mean-square generalisation error used along the paper.

\subsection{Non-centred activations}\label{app:non-centered}
Consider a non-centred activation function, i.e., $\mu_0\neq 0$ in \eqref{eq:sigma_hermite}. This reflects on the law of the post-activations, which will still be Gaussian, centred at
\begin{align}
    \EE_\bx\lambda^a=\frac{\mu_0}{\sqrt
    k}\sum_{i=1}^kv_i=:\mu_0
    \Lambda,
\end{align}and with the covariance given by \eqref{eq:K} (we are assuming $ Q_{W}^{aa}=1$; if not, $ Q_{W}^{aa}=r$, the formula can be generalised as explained in App.~\ref{app:hermite}, and that the readout weights are quenched). In the above, we have introduced the new mean parameter $\Lambda$. Notice that, if the $\bv $'s have a $\bar v=O(1)$ mean, then $\Lambda$ scales as $\sqrt{k}$ due to our choice of normalisation.

One can carry out the replica computation for a fixed $\Lambda$. This new parameter, being quenched, does not affect the entropic term. It will only appear in the energetic term as a shift to the means, yielding
    \begin{equation}
    F_E=F_E(\bK,\Lambda) = \ln \int dy\int d\blambda\frac{e^{-\frac{1}{2}\blambda^\intercal\bK^{-1}\blambda}}{\sqrt{(2\pi)^{s+1}\det \bK}} \prod_{a=0}^s P_{\rm{out}}(y\mid \lambda^a+\mu_0 \Lambda).
\end{equation}
Within the replica symmetric ansatz, the above turns into
\begin{align*}
    e^{F_E}=\int dy \,\EE_{\xi,u^0}P_{\rm out}\Big(y\mid \mu_0 \Lambda + \xi\sqrt{\frac{m_K^2}{q_K}}+u^0\sqrt{\rho_K-\frac{m_K^2}{q_K}}\Big)\prod_{a=1}^s\EE_{u^a}P_{\rm out}(y\mid \mu_0 \Lambda + \xi\sqrt{q_K}+u^a\sqrt{r_K-q_K}).
\end{align*}
Therefore, the simplification of the potential $F_E$ proceeds as in the centred activation case, yielding at leading order in the number $s$ of replicas
\begin{align*}
    \frac{F_E(r_K,q_K,\Lambda)}{s}\!=\!\int dy\, \EE_{\xi,u^0}P_{\rm out}\Big(y\mid \mu_0 \Lambda + \xi\sqrt{q_K}+u^0\sqrt{r_K-q_K}\Big)\ln \EE_{u}P_{\rm out}(y\mid\mu_0 \Lambda + \xi\sqrt{q_K}+u\sqrt{r_K-q_K}) + O(s)
\end{align*}
in the Bayes-optimal setting. In the case when $P_{\rm out}(y\mid \lambda)=f(y-\lambda)$ then one can verify that the contributions due to the means, containing $\mu_0$, cancel each other. This is verified in our running example where $P_{\rm out}$ is the Gaussian channel:
\begin{equation}
    \frac{F_E(r_K,q_K,\Lambda)}{s} = 
    -\frac{1}{2}\ln\big[2\pi(\Delta+r_K-q_K)\big]-\frac{1}{2} - \frac{\mu_0^2}{2}\frac{(\Lambda - \Lambda)^2}{\Delta+r_K-q_K} + O(s)=-\frac{1}{2}\ln\big[2\pi(\Delta+r_K-q_K)\big]-\frac{1}{2} + O(s).
\end{equation}

\section{Alternative simplifications of \texorpdfstring{$P((\bS^a_2)\mid \bm{\calQ}_W)$}{TEXT} through moment matching}\label{app:alternatives}

A crucial step that allowed us to obtain a closed-form expression for the model's free entropy is the relaxation $\tilde{P}((\bS^a_2)\mid \bm{\calQ}_W)$~\eqref{eq:effectivePS2} of the true measure $P((\bS^a_2)\mid \bm{\calQ}_W)$~\eqref{eq:truePS2} entering the replicated partition function, as explained in Sec.~\ref{sec:theory}. The specific form we chose (tilted Wishart distribution with a matching second moment) has the advantage of capturing crucial features of the true measure, such as the fact that the matrices $\bS^a_2$ are generalised Wishart matrices with coupled replicas, while keeping the problem solvable with techniques derived from random matrix theory of rotationally invariant ensembles. In this appendix, we report some alternative routes one can take to simplify, or potentially improve the theory.

\subsection{A factorised simplified distribution}\label{app:simpler_ansatz}

In the specialisation phase, one can assume that the only crucial feature to keep track in relaxing $P((\bS^a_2)\mid\bm{\calQ}_W)$~\eqref{eq:truePS2} is the coupling between different replicas, becoming more and more relevant as $\alpha$ increases. In this case, inspired by \cite{sakata2013,kabashima2016phase}, in order to relax~\eqref{eq:truePS2} we can propose the Gaussian ansatz
\begin{align}
    &d\bar{P}((\bS_2^a)\mid\bm{\calQ}_W) = \prod_{a=0}^s d \bS^a_{2}\prod_{\alpha=1}^d  \delta(S^a_{2;\alpha \alpha}  - \sqrt{k} \bar{v})\times  \prod_{\alpha_1<\alpha_2}^d  
    \frac{e^{-\frac{1}{2}\sum_{a,b=0}^s S^{a}_{2;\alpha_1\alpha_2} \bar{\tau}^{ab}(\bm{\calQ}_W) S^{b}_{2;\alpha_1\alpha_2}}}{\sqrt{(2\pi)^{s+1}\det(\bar{\bm{\tau}}(\bm{\calQ}_W)^{-1})}},\label{eq:gaussPS2}
\end{align}
where $\bar{v}$ is the mean of the readout prior $P_v$, and $\bar{\bm{\tau}}(\bm{\calQ}_W):= (\bar{\tau}^{ab}(\bm{\calQ}_W))_{a,b}$ is fixed by 
\begin{equation*}
    [\bar{\bm{\tau}}(\bm{\calQ}_W)^{-1}]_{ab} = \E_{v\sim P_v} v^2 \calQ_W^{ab}(v)^2.
\end{equation*}
In words, first, the diagonal elements of $\bS_2^a$ are $d$ random variables whose $O(1)$ fluctuations cannot affect the free entropy in the asymptotic regime we are considering, being too few compared to $n=\Theta(d^2)$. Hence, we assume they concentrate to their mean. Concerning the $d(d-1)/2$ off-diagonal elements of the matrices $(\bS_2^a)_a$, they are zero-mean variables whose distribution at given $\bm{\calQ}_W$ is assumed to be factorised over the input indices. The definition of $\bar{\bm{\tau}}(\bm{\calQ}_W)$ ensures matching with the true second moment~\eqref{eq:truePS2_second_moment}.

\eqref{eq:gaussPS2} is considerably simpler than~\eqref{eq:effectivePS2}:
following this ansatz, the entropic contribution to the free entropy gives

\begin{align}
    &e^{\bar{F}_S}:= \int \prod_{a\le b,0}^s d \hat{Q}_2^{ab}\,
     e^{kd \ln V_W(\bm{\mathcal{Q}}_W)+\frac{d^2}{4}\Tr \hat \bQ^{\intercal}_2 \bQ_2}
    \Big[\int \prod_{a=0}^s dS^a_2\,\frac{e^{-\frac{1}{2}\sum_{a,b=0}^s S^{a}_2  [\bar{\tau}^{ab}(\bm{\calQ}_W) +\hat Q_2^{ab} ]S^{b}_2}}{\sqrt{(2\pi )^{s+1}\det( \bar{\bm{\tau}}(\bm{\calQ_W})^{-1})}}
    \Big]^{d(d-1)/2}\nonumber\\
    &\qquad\qquad\qquad\times \int\prod_{a=0}^s\prod_{\alpha=1}^d dS^a_{2;\alpha\alpha} \delta(S^a_{2;\alpha\alpha}-\sqrt{k}\bar{v})\,e^{-\frac{1}{4}\sum_{a,b=0}^s\hat Q_2^{ab}\sum_{\alpha=1}^dS_{2;\alpha\alpha}^aS_{2;\alpha\alpha}^b},
\end{align}
instead of~\eqref{eq:F_S_step1_appendix}. Integration over the diagonal elements $(S_{2;\alpha\alpha}^a)_{\alpha}$ can be done straightforwardly, yielding
\begin{align}
    e^{\bar{F}_S} &=\int \prod_{a\leq b,0}^s d\hat{Q}_2^{ab} \,e^{kd\ln V_W(\bm{\mathcal{Q}}_W) +\frac{d^2}{4}\Tr\hat \bQ_2^\intercal(\bQ_2-\gamma\mathbf{1}\mathbf{1}^\intercal \bar{v}^2)}\Big[\int \prod_{a=0}^s dS^a_2\,\frac{e^{-\frac{1}{2}\sum_{a,b=0}^s S^{a}_2  [\bar{\tau}^{ab}(\bm{\calQ}_W) +\hat Q_2^{ab} ]S^{b}_2}}{\sqrt{(2\pi )^{s+1}\det( \bar{\bm{\tau}}(\bm{\calQ}_W)^{-1})}}
    \Big]^{d(d-1)/2}.
\end{align}
The remaining Gaussian integral over the off-diagonal elements of $\bS_2$ can be performed exactly, leading to
\begin{align}
\label{eq:exp_Z^n}
    e^{\bar{F}_S} &=\int \prod_{a\leq b,0}^s d\hat{Q}_2^{ab} \,e^{kd\ln V_W(\bm{\mathcal{Q}}_W) +\frac{d^2}{4}\Tr\hat \bQ_2^\intercal(\bQ_2-\gamma\mathbf{1}\mathbf{1}^\intercal\bar{v}^2)-\frac{d(d-1)}{4}\ln\det[\I_{s+1}+\hat\bQ_2 \bar{\bm{\tau}}(\bm{\calQ}_W)^{-1}]}.
\end{align}
In order to proceed and perform the $s\to 0^+$ limit, we use the RS ansatz for the overlap matrices, combined with the Nishimori identities, as explained above. The only difference w.r.t. the approach detailed in Appendix~\ref{app:replicas} is the determinant in the exponent of the integrand of~\eqref{eq:exp_Z^n}, which reads
\begin{align}
\ln\det[\I_{s+1}+\hat\bQ_2\bar{\bm{\tau}}(\bm{\calQ}_W)^{-1}] = 
s\ln [1+\hat{q}_2 (1-\E_{v\sim P_v}v^2 \mathcal{Q}_W(v)^2)]-s\hat{q}_2
+O(s^2).
\end{align}
After taking the replica and high-dimensional limits, the resulting free entropy is
\begin{equation}
\begin{aligned}
\label{freeEntSpecialisation-Bayes-Opt}
    f_{\rm sp}^{\alpha,\gamma} = {}&
    \psi_{P_{\text{out}}}(q_K(q_2,\calQ_W);r_K)
    + \frac{(1+\gamma \bar{v}^2 -q_2) \Hat{q}_2}{4\alpha}
    + \frac{\gamma}{\alpha} \E_{v\sim P_v} \big[ \psi_{P_W}(\hat \calQ_W (v))  
    -\frac{1}{2} \mathcal{Q}_W(v) \Hat{\calQ}_W(v) \big] \\
    &\qquad- \frac{1}{4\alpha}\ln \big[ 1+\hat q_2   (1 - \E_{v\sim P_v} v^2 \mathcal{Q}_W(v)^2) \big],
\end{aligned}    
\end{equation}
to be extremised w.r.t. $q_2, \hat{q}_2, \{\calQ_W (\mathsf{v}), \hat{\calQ}_W (\mathsf{v})\}$.
The main advantage of this expression over~\eqref{eq:free_ent_appendix} is its simplicity: the moment-matching condition fixing $\bar{\bm{\tau}}(\bm{\calQ}_W)$ is straightforward (and has been solved explicitly in the final formula) and the result does not depend on the non-trivial (and difficult to numerically evaluate) function $\iota(\eta)$, which is the mutual information of the associated matrix denoising problem (which has been effectively replaced by the much simpler denoising problem of independent Gaussian variables under Gaussian noise). Moreover, one can show, in the same fashion as done in Appendix~\ref{app:large_alpha}, that the generalisation error predicted from this expression has the same large-$\alpha$ behaviour than the one obtained from~\eqref{eq:free_ent_appendix}. However, not surprisingly, being derived from an ansatz ignoring the Wishart-like nature of the matrices $\bS_2^a$, this expression does not reproduce the expected behaviour of the model in the universal phase, i.e. for $\alpha < \alpha_{\rm sp}(\gamma)$. 

\begin{figure}[t!!!]
    \centering
    \includegraphics[width=0.485\linewidth,clip]{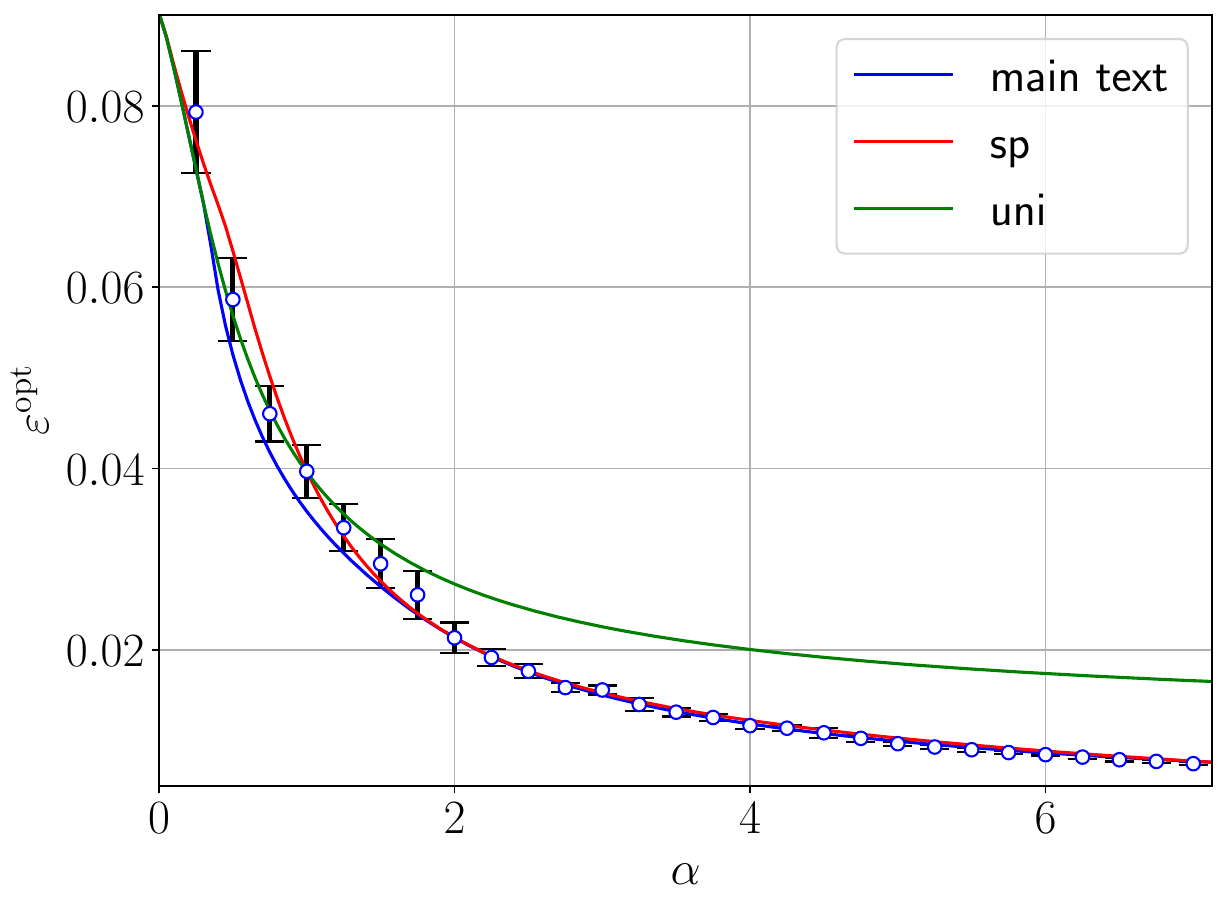}
    \includegraphics[width=0.485\linewidth,clip]{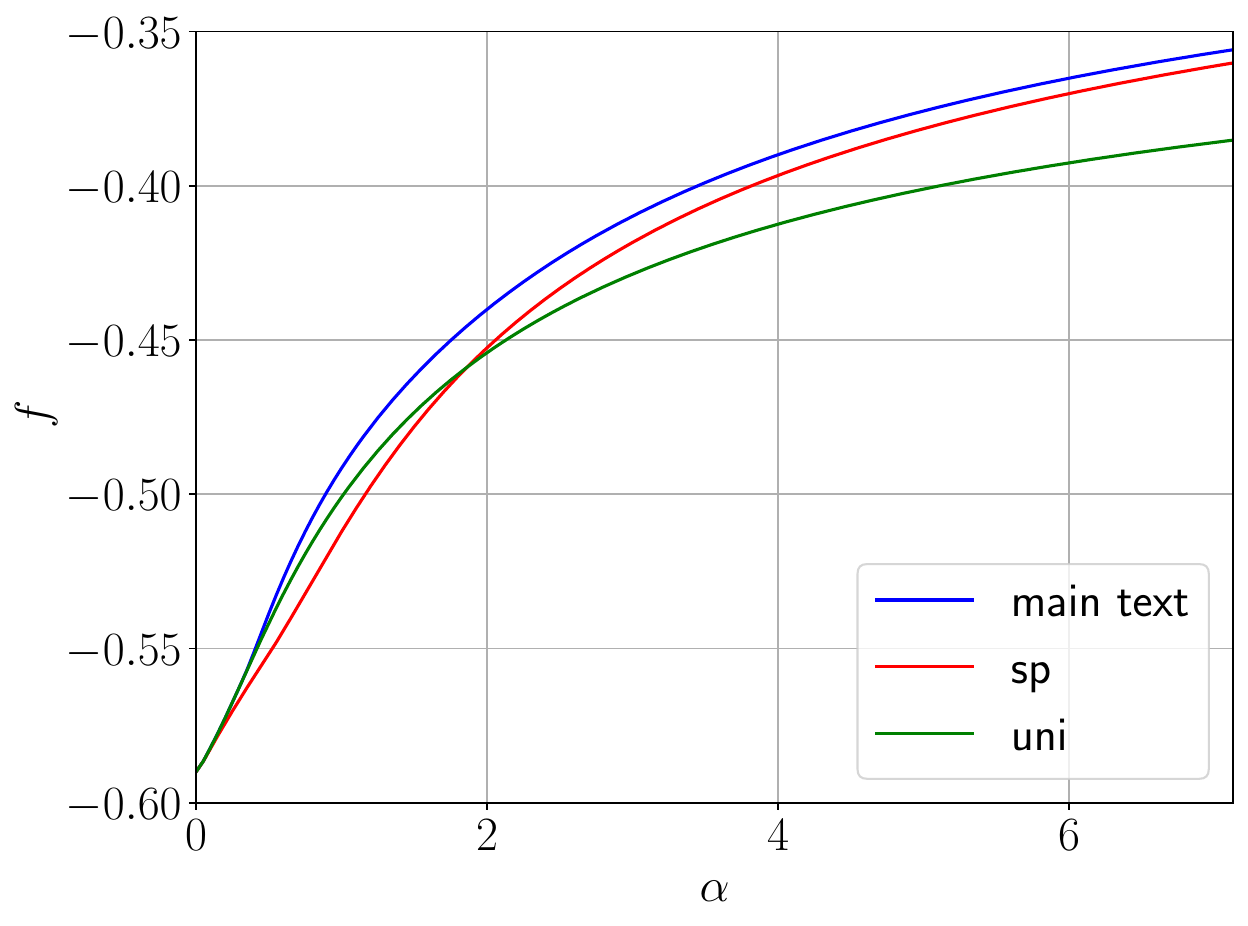}
    \includegraphics[width=0.485\linewidth,clip]{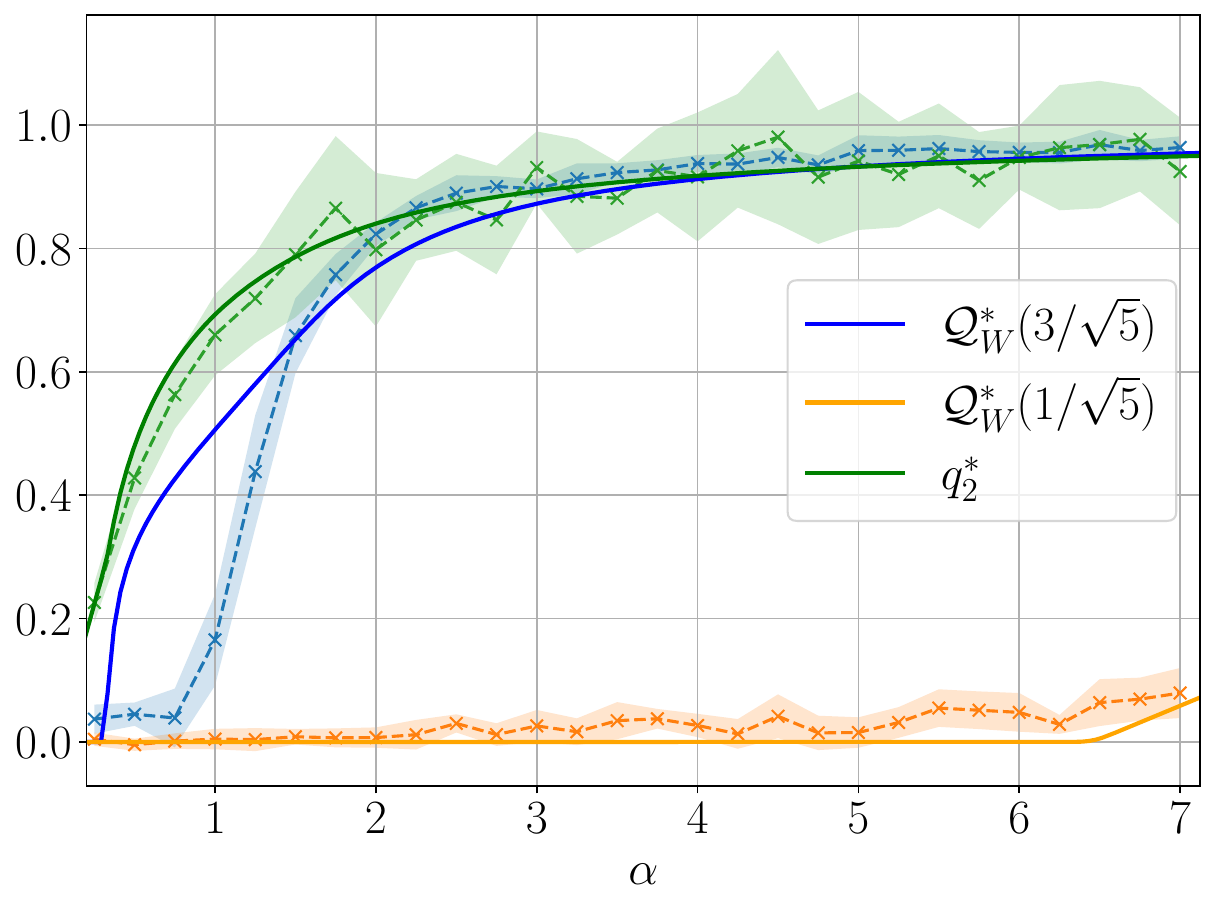}
    \includegraphics[width=0.485\linewidth,clip]{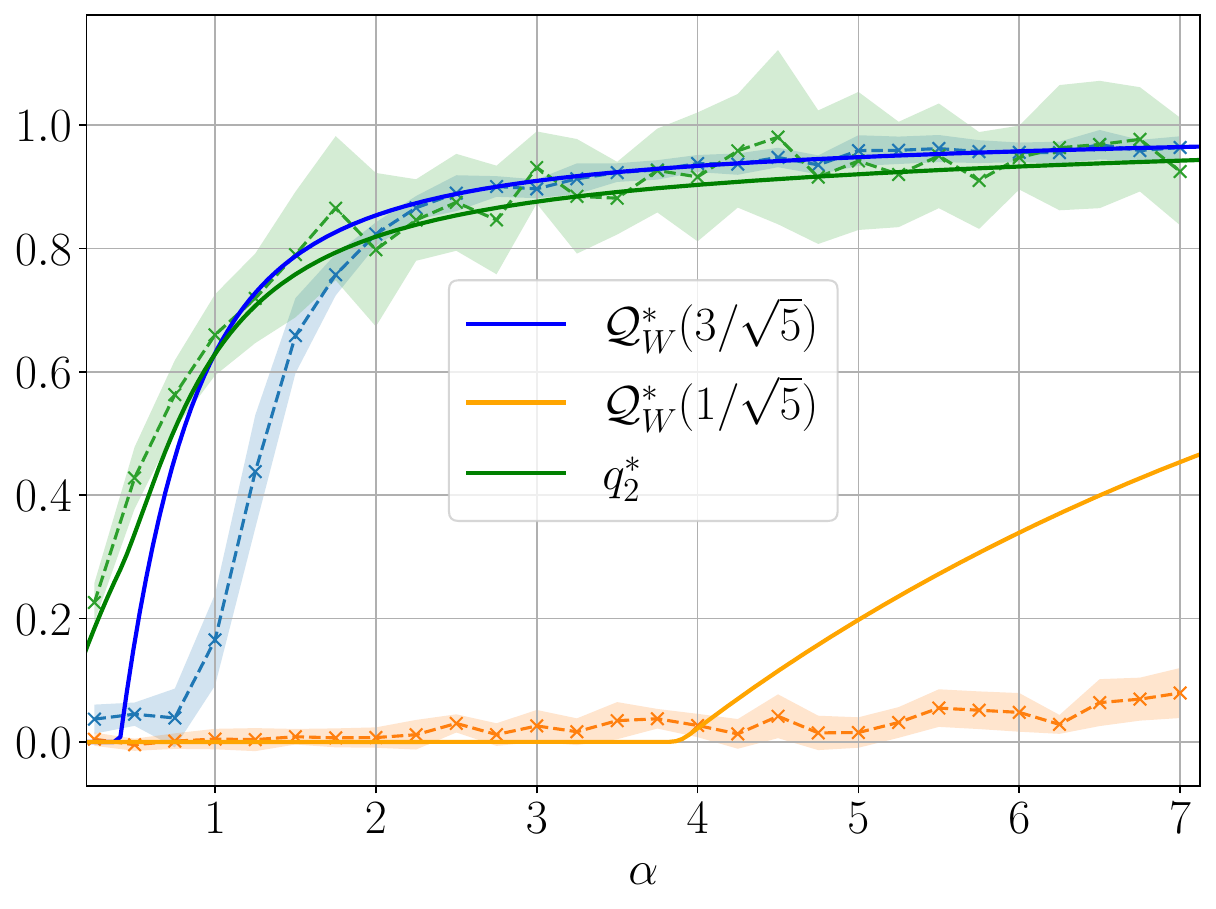}
    \caption{Different theoretical curves and numerical results for ReLU(x) activation, $P_v = \frac{1}{4}(\delta_{-3/\sqrt{5}} + \delta_{-1/\sqrt{5}} + \delta_{1/\sqrt{5}} + \delta_{3/\sqrt{5}})$ , $d=150$, $\gamma=0.5$, with linear readout with Gaussian noise of variance $\Delta=0.1$
    \textbf{Top left}: Optimal mean-square generalisation error predicted by the theory reported in the main text (solid blue) versus the branch obtained from the simplified ansatz~\eqref{eq:gaussPS2} (solid red); the green solid line shows the universal branch corresponding to $\mathcal{Q}_W \equiv 0$, and empty circles are HMC results with informative initialisation. 
    \textbf{Top right}: Theoretical free entropy curves (colors and linestyles as top left). 
    \textbf{Bottom}: Predictions for the overlaps $\mathcal{Q}_W(\mathsf{v})$ and $q_2$ from the theory devised in the main text (\textbf{left}) and in Appendix~\ref{app:simpler_ansatz} (\textbf{right}).
    }
    \label{fig:Simplified_vs_Unified}
\end{figure}

To fix this issue, one can compare the predictions of the theory derived from this ansatz, with the ones obtained by plugging $\calQ_W(\mathsf{v}) = 0 \ \forall \ \mathsf{v}$ (denoted $\calQ_W\equiv 0$) in the theory devised in the main text~\eqref{eq:fRS},
\begin{equation}
    f_{\rm uni}^{\alpha,\gamma} :=  \psi_{P_{\text{out}}}(q_K(q_2,\mathcal{Q}_W\equiv 0);r_K) + \frac{1}{4\alpha}(1 + \gamma \bar{v}^2-q_2) \Hat{q}_2
     -\frac{1}{\alpha} \iota(\hat q_2 )  ,
\end{equation}
to be extremised now only w.r.t. the scalar parameters $q_2$, $\hat{q}_2$ (one can easily verify that, for $\calQ_W\equiv 0$, $\tau(\calQ_W) = 0$ and the extremisation w.r.t. $\hat{\calQ}_W$ in~\eqref{eq:fRS} gives $\hat{\calQ}_W \equiv 0$). Notice that $f_{\rm uni}^{\alpha,\gamma}$ is not depending on the prior over the inner weights, which is the reason why we are calling it ``universal''. For consistency, the two free entropies $f_{\rm sp}^{\alpha,\gamma}$, $ f_{\rm uni}^{\alpha,\gamma}$ should be compared through a discrete variational principle, that is the free entropy of the model is predicted to be
\begin{equation}
    \bar{f}^{\alpha,\gamma}_{\rm  RS} := \max \{\extr f_{\rm uni}^{\alpha,\gamma}, \extr f_{\rm sp}^{\alpha,\gamma}\},
    \label{eq:f_meta}
\end{equation}
instead of the unified variational form~\eqref{eq:fRS}. Quite generally, $\extr f_{\rm uni}^{\alpha,\gamma} > \extr f_{\rm sp}^{\alpha,\gamma}$ for low values of $\alpha$, so that the behaviour of the model in the universal phase is correctly predicted. The curves cross at a critical value
\begin{equation}
    \bar{\alpha}_{\rm sp} (\gamma) = \sup \{ \alpha \mid \extr f_{\rm uni}^{\alpha,\gamma} > \extr f_{\rm sp}^{\alpha,\gamma} \},
    \label{eq:alpha_c_meta}
\end{equation}
instead of the value $\alpha_{\rm sp} (\gamma)$ reported in the main. This approach has been profitably adopted in~\cite{barbier2024phase} in the context of matrix denoising\footnote{This is also the approach we used in a earlier version of this paper (superseded by the present one), accessible on ArXiv at \href{https://arxiv.org/pdf/2501.18530?}{this link}.}, a problem sharing some of the challenges presented in this paper. In this respect, it provides a heuristic solution that quantitatively predicts the behaviour of the model in most of its phase diagram. Moreover, for any activation $\sigma$ with a second Hermite coefficient $\mu_2=0$ (e.g., all odd activations) the ansatz \eqref{eq:gaussPS2} yields the same theory as the one devised in the main text, as in this case $q_K(q_2, \calQ_W)$ entering the energetic part of the free entropy does not depend on $q_2$, so that the extremisation selects $q_2=\hat{q}_2=0$ and the remaining parts of~\eqref{freeEntSpecialisation-Bayes-Opt} match the ones of~\eqref{eq:fRS}. Finally, \eqref{eq:gaussPS2} is consistent with the observation that specialisation never arises in the case of quadratic activation and Gaussian prior over the inner weights: in this case, one can check that the universal branch $\extr f_{\rm uni}^{\alpha,\gamma}$ is always higher than $\extr f_{\rm sp}^{\alpha,\gamma}$, and thus never selected by~\eqref{eq:f_meta}. For a convincing check on the validity of this approach, and a comparison with the theory devised in the main text and numerical results, see Fig.~\ref{fig:Simplified_vs_Unified}, top left panel.

\begin{figure}[t!!!]
    \centering
    \includegraphics[width=0.5\linewidth,clip]{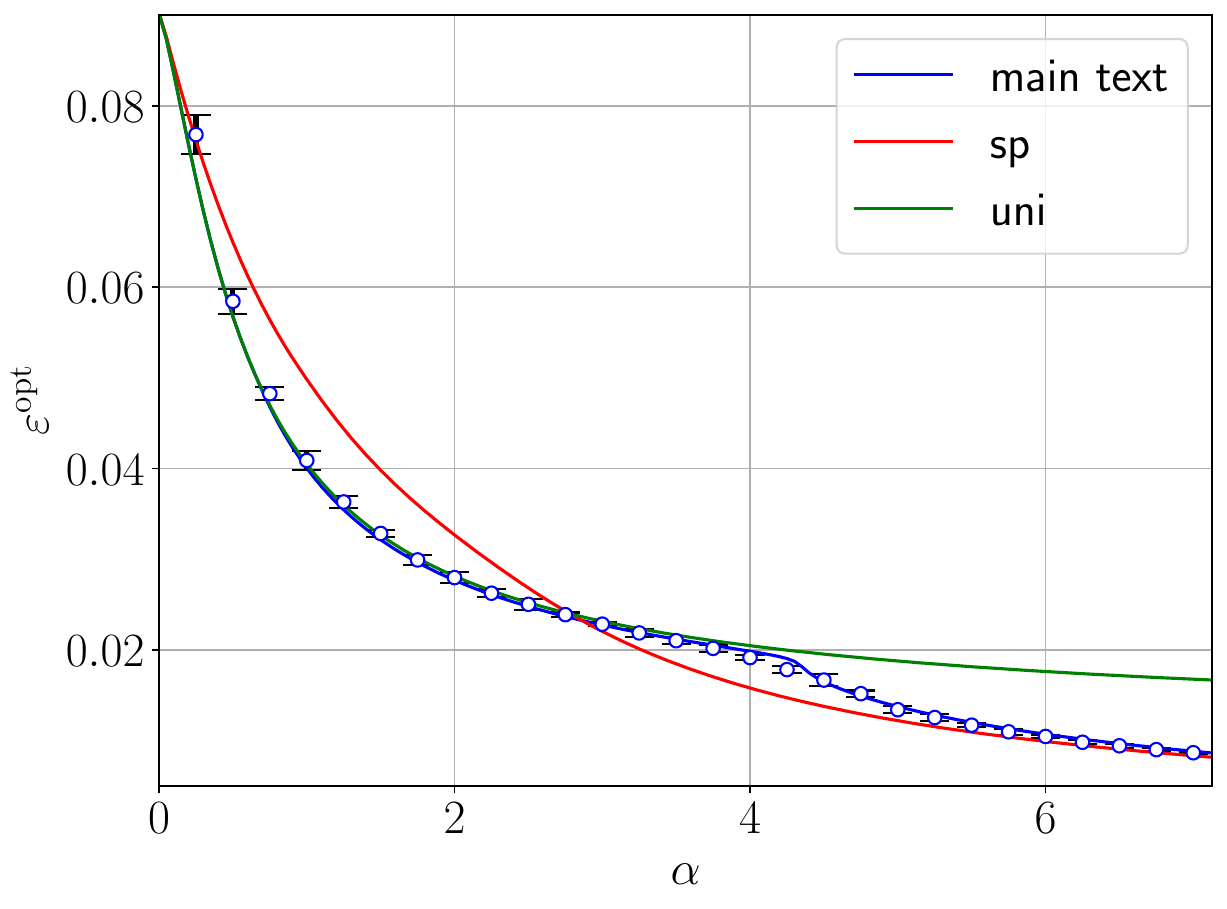}
    
    \caption{Generalisation error for ReLU activation and Rademacher readout prior $P_v$ of the theory reported in the main text (solid blue) versus the branch obtained from the simplified ansatz~\eqref{eq:gaussPS2} (solid red); the green solid line shows $\mathcal{Q}_W \equiv 0$ (universal branch), and empty circles are HMC results with informative initialisation.}
    \label{fig:Simplified_vs_Unified_2}
\end{figure}

However, despite its merits listed above, this Appendix's approach presents some issues, both from the theoretical and practical points of view:
\begin{enumerate}
    \item[(i)] the final free entropy of the model is obtained by comparing curves derived from completely different ans\"atze for the distribution $P((\bS^a_2)\mid\bm{\calQ}_W)$ (Gaussian with coupled replicas, leading to $f_{\rm sp}$, vs. pure generalised Wishart with independent replicas, leading to $f_{\rm uni}$), rather than within a unified theory as in the main text;
    \item[(ii)] the predicted critical value $\bar{\alpha}_{\rm sp} (\gamma)$ seems to be systematically larger than the one observed in experiments (see Fig.~\ref{fig:Simplified_vs_Unified}, top right panel, and compare the crossing point of the ``sp'' and ``uni'' free entropies with the actual transition where the numerical points depart from the universal branch in the top left panel); 
    \item[(iii)] predictions for the functional overlap $\calQ_W^*$ from this approach are in much worse agreement with experimental data w.r.t. the ones from the theory presented in the main text (see Fig.~\ref{fig:Simplified_vs_Unified}, bottom panel, and compare with Fig.~\ref{fig:v_non-constant} in the main text);
    \item[(iv)] in the cases we tested, the prediction for the generalisation error from the theory devised in the main text are in much better agreement with numerical simulations than the one from this Appendix (see Fig.~\ref{fig:Simplified_vs_Unified_2} for a comparison).
\end{enumerate}
Therefore, the more elaborate theory presented in the main is not only more meaningful from the theoretical viewpoint, but also in overall better agreement with simulations.

\subsection{Possible refined analyses with structured \texorpdfstring{$\bS_2$}{S2} matrices}\label{app:structured_S2}

In the main text, we kept track of the inhomogeneous profile of the readouts induced by the non-trivial distribution $P_v$, which is ultimately responsible for the sequence of specialisation phase transitions occurring at increasing $\alpha$, thanks to a functional order parameter $\calQ_W(\mathsf{v})$ measuring how much the student's hidden weights corresponding to all the readout elements equal to $\mathsf{v}$ have aligned with the teacher's. However, when writing $\tilde{P}((\bS_2^a)\mid \bm{\calQ}_W)$ we treated the tensor $\bS_2^a$ as a whole, without considering the possibility that its ``components'' 
\begin{align}
S_{2;\alpha_1\alpha_2}^a(\mathsf{v}) :=  \frac{\mathsf{v}}{\sqrt{| \mathcal{I}_{\mathsf{v}}|}}\sum_{i\in \mathcal{I}_{\mathsf{v}}} W^a_{i\alpha_1} W^a_{i\alpha_2}\label{eq:S2_components}    
\end{align}
could follow different laws for different $\mathsf{v}\in \mathsf{V}$.
To do so, let us define
\begin{align}\label{eq:Q2_general_relation}
   Q_{2}^{ab}=\frac{1}{k} \sum_{\mathsf{v},\mathsf{v}'} \mathsf{v}\,\mathsf{v}' \sum_{i\in \mathcal{I}_{\mathsf{v}}, j \in \mathcal{I}_{\mathsf{v'}}} (\Omega_{ij}^{ab})^2  = \sum_{\mathsf{v},\mathsf{v}'} \frac{\sqrt{|\mathcal{I}_{\mathsf{v}} | |\mathcal{I}_{\mathsf{v'}}|}}{k} \calQ_2^{ab}(\mathsf{v},\mathsf{v}'), \quad \text{where} \quad \calQ_2^{ab}(\mathsf{v},\mathsf{v}') := \frac1{d^2}\Tr \,\bS_2^a(\mathsf{v})\bS_2^{b}(\mathsf{v}')^\intercal.
\end{align}
The generalisation of \eqref{eq:truePS2_second_moment} then reads
\begin{align}
    \int dP((\bS_2^a)\mid \bm{\mathcal{Q}}_W) &\frac{1}{d^2}\Tr\,\bS_2^a(\mathsf{v})\bS_2^{b}(\mathsf{v}')^\intercal=\delta_{\mathsf{v}\mathsf{v}'} \mathsf{v}^2 \mathcal{Q}_W^{ab}(\mathsf{v})^2+\gamma \,\mathsf{v}\mathsf{v}' \sqrt{P_v(\mathsf{v}) P_v(\mathsf{v}')} 
    \label{eq:moment_matching_v}
\end{align}
w.r.t. the true distribution $P((\bS_2^a)\mid \bm{\mathcal{Q}}_W)$ reported in~\eqref{eq:truePS2}. Despite the already good match of the theory in the main with the numerics, taking into account this additional level of structure thanks to a refined simplified measure could potentially lead to further improvements. The simplified measure able to match these moment-matching conditions while taking into account the Wishart form \eqref{eq:S2_components}     of the matrices $(\bS_2^a(\mathsf{v}))$ is
\begin{align}\label{eq:unsolvable_relaxation}
    d\bar{P}((\bS_2^a)\mid\bm{\calQ}_W) \propto \prod_{\mathsf{v} \in \mathsf{V}} \prod_{a} dP_S^{\mathsf{v}}(\bS_2^a(\mathsf{v}))  \times\prod_{\mathsf{v} \in \mathsf{V}} \prod_{a < b}e^{\frac{1}{2}  \bar{\tau}^{ab}_\mathsf{v}(\bm{\calQ}_W)\Tr \bS_2^a(\mathsf{v}) \bS_2^b(\mathsf{v})},
\end{align}
where $P_S^\mathsf{v}$ is the law of a random matrix $\mathsf{v} \bar{\bW} \bar{\bW}^\intercal |\mathcal{I}_{\mathsf{v}}|^{-1/2}$ 
with $\bar\bW\in\mathbb{R}^{d\times|\mathcal{I}_{\mathsf{v}}|}$ 
having i.i.d. standard Gaussian entries. For properly 
chosen $(\bar{\tau}_\mathsf{v}^{ab})$, \eqref{eq:moment_matching_v} is verified for this simplified measure. 

However, the order parameters $(\calQ_2^{ab}(\mathsf{v},\mathsf{v}'))$ are difficult to deal with if keeping a general form, as they not only imply coupled replicas $(\bS_2^a(\mathsf{v}))_a$ for a given $\mathsf{v}$ (a kind of coupling that is easily linearised with a single Hubbard-Stratonovich transformation, within the replica symmetric treatment justified in Bayes-optimal learning), but also a coupling for different values of the variable $\mathsf{v}$. Linearising it would yield a more complicated matrix model than the integral reported in~\eqref{eq:J_replicas}, because the resulting coupling field would break rotational invariance and therefore the model does not have a form which is known to be solvable, see \cite{kazakov2000solvable}.

A first idea to simplify $P((\bS^a_2)\mid\bm{\calQ}_W)$~\eqref{eq:truePS2} while taking into account the additional structure induced by \eqref{eq:Q2_general_relation}, \eqref{eq:moment_matching_v} and maintaining a solvable model, is to consider a generalisation of the relaxation \eqref{eq:gaussPS2}. This entails dropping entirely the dependencies among matrix entries, induced by their Wishart-like form \eqref{eq:S2_components}, for each $\bS_2^a(\mathsf{v})$. In this case, the moment constraints \eqref{eq:moment_matching_v} can be exactly enforced by choosing the simplified measure
\begin{align}
    &d\bar{P}((\bS_2^a)\mid\bm{\calQ}_W) = \prod_{\mathsf{v}\in \mathsf{V}}\prod_{a=0}^s d \bS^a_{2}(\mathsf{v})\prod_{\alpha=1}^d  \delta(S^a_{2;\alpha \alpha}(\mathsf{v}) - \mathsf{v} \sqrt{|\mathcal{I}_\mathsf{v}|})\times  \prod_{\mathsf{v}\in \mathsf{V}}\prod_{\alpha_1<\alpha_2}^d  
    \frac{e^{-\frac{1}{2}\sum_{a,b=0}^s S^{a}_{2;\alpha_1\alpha_2}(\mathsf{v}) \bar{\tau}_\mathsf{v}^{ab}(\bm{\calQ}_W) S^{b}_{2;\alpha_1\alpha_2}(\mathsf{v})}}{\sqrt{(2\pi)^{s+1}\det(\bar{\bm{\tau}}_\mathsf{v}(\bm{\calQ}_W)^{-1})}}.
\end{align}
The parameters $(\bar\tau^{ab}_{\mathsf{v}}(\bm{\calQ}_W))$ are then properly chosen to enforce \eqref{eq:moment_matching_v} for all $0\le a\le b\le s$ and $\mathsf{v}, \mathsf{v}'\in \mathsf{V}$. Using this measure, the resulting entropic term, taking into account the degeneracy of the order parameters $(\calQ_2^{ab}(\mathsf{v},\mathsf{v}'))$ and $(\calQ_W^{ab}(\mathsf{v}))$, remains tractable through Gaussian integrals (the energetic term is obviously unchanged once we express $(Q_2^{ab})$ entering it using these new order parameters through the identity \eqref{eq:Q2_general_relation}, and keeping in mind that nothing changes for higher order overlaps compared to the theory in the main). We leave for future work the analysis of this Gaussian relaxation and other possible simplifications of \eqref{eq:unsolvable_relaxation} leading to solvable models.

\section{Linking free entropy and mutual information\label{app:mutual_info}}

It is possible to relate the mutual information (MI) of the inference task to the free entropy $f_n=\E\ln \mathcal{Z}$ introduced in the main. Indeed, we can write the MI as
\begin{equation}
    \frac{I(\bW^0;\mathcal{D})}{kd} = \frac{\mathcal{H}(\mathcal{D})}{kd} - \frac{\mathcal{H}(\mathcal{D}\mid\bW^0)}{kd},
\end{equation}
where $\mathcal{H}(Y\mid X)$ is the conditional Shannon entropy of $Y$ given $X$. It is straightforward to show that the free entropy is
\begin{equation}
  -\frac{\alpha}{\gamma}f_n = \frac{\mathcal{H}(\{ y_\mu\}_{\mu \leq n}\mid \{ \bx_\mu\}_{\mu \leq n})}{kd} = \frac{\mathcal{H}(\mathcal{D})}{kd} - \frac{\mathcal{H}(\{ \bx_\mu\}_{\mu \leq n})}{kd},
\end{equation}
by the chain rule for the entropy. On the other hand $\mathcal{H}(\mathcal{D}\mid\bW^0)=\mathcal{H}(\{y_\mu\}\mid\bW^0,\{\bx_\mu\})+\mathcal{H}(\{\bx_\mu\})$, i.e.,
\begin{equation}
    \frac{\mathcal{H}(\mathcal{D}\mid \bW^0)}{kd} \approx -\frac{\alpha}{\gamma} \EE_{\lambda}  \int dy P_{\text{out}}(y\mid \lambda) \ln P_{\text{out}}(y\mid\lambda) + \frac{\mathcal{H}(\{ \bx_\mu\}_{\mu \leq n})}{kd} ,
\end{equation}
where $\lambda\sim \calN(0,r_K)$, with $r_K$ given by \eqref{eq:covariance_vars} (assuming here that $\mu_0=0$, see App.~\ref{app:non-centered} if the activation $\sigma$ is non-centred), and the equality holds asymptotically in the limit $\lim$.
This allows us to express the MI as
\begin{equation}
\label{eq:MI_generic_channel}
    \frac{I(\bW^0;\mathcal{D})}{kd} = -\frac{\alpha}{\gamma} f_n + \frac{\alpha}{\gamma} \EE_{\lambda}  \int dy P_{\text{out}}(y|\lambda) \ln P_{\text{out}}(y|\lambda).
\end{equation} 
Specialising the equation to the Gaussian channel, one obtains
\begin{equation}
\label{eq:MI_gaussian_channel}
    \frac{I(\bW^0;\mathcal{D})}{kd} = -\frac{\alpha}{\gamma} f_n - \frac{\alpha}{2\gamma} \ln(2\pi e \Delta).
\end{equation} 
Note that the choice of normalising by $kd$ is not accidental. Indeed, the number of parameters is $kd+k\approx kd$. Hence with this choice one can interpret the parameter $\alpha$ as an effective signal-to-noise ratio.

\begin{remark}
 The arguments of \cite{barbier2024phase} to show the existence of an upper bound on the mutual information per variable in the case of discrete variables and the associated inevitable breaking of prior universality beyond a certain threshold in matrix denoising apply to the present model too. It implies, as in the aforementioned paper, that the mutual information per variable cannot go beyond $\ln 2$ for Rademacher inner weights. Our theory is consistent with this fact; this is a direct consequence of the analysis in App.~\ref{app:large_alpha} (see in particular~\eqref{eq:large_alpha_limit}) specialised to binary prior over $\bW$.
\end{remark}

\section{Large sample rate limit of \texorpdfstring{$f_{\rm RS}^{\alpha,\gamma}$}{fRS}}\label{app:large_alpha}
In this section we show that when the prior over the weights $\bW$ is discrete the MI can never exceed the entropy of the prior itself. 

To do this, we first need to control the function $\rm mmse$ when its argument is large. By a saddle point argument, it is not difficult to show that the leading term for ${\rm mmse}_S(\tau)$ when $\tau\to\infty$ if of the type $C(\gamma)/\tau$ for a proper constant $C$ depending at most on the rectangularity ratio $\gamma$. 

We now notice that the equation for $\hat{\mathcal{Q}}_W(v)$ in \eqref{eq:NSB_equations_generic_ch} can be rewritten as
\begin{align}
    \hat {\mathcal{Q}}_{W}(v) = \frac{1}{2\gamma}[{\rm mmse}_S(\tau)-{\rm mmse}_S(\tau+\hat q_2)]\partial_{{\mathcal{Q}}_{W}(v)} \tau  + 2\frac{\alpha}{\gamma} \partial_{{\mathcal{Q}}_{W}(v)} \psi_{P_{\text{out}}}(q_K(q_2,\mathcal{Q}_W);r_K).
\end{align}
For $\alpha\to\infty$ we make the self-consistent ansatz $\mathcal{Q}_W(v)=1-o_\alpha(1)$. As a consequence $1/\tau$ has to vanish by the moment matching condition \eqref{eq:moment_matching} as $o_\alpha(1)$ too. Using the very same equation, we are also able to evaluate $\partial_{\mathcal{Q}_W(v)}\tau$ as follows:
\begin{align}
    \partial_{\mathcal{Q}_W(v)}\tau=\frac{-2v^2 \mathcal{Q}_W(v)}{{\rm mmse'}(\tau)}\sim \tau^2
\end{align}
as $\alpha\to\infty$, where we have used ${\rm mmse}_S(\tau)\sim C(\gamma)/\tau$ to estimate the derivative. We use the same approximation for the two $\rm mmse$'s appearing in the fixed point equation for $\hat{\mathcal{Q}}_W(v)$:
\begin{align}
    \hat {\mathcal{Q}}_{W}(v) \sim \frac{\hat q_2}{2\gamma(\tau(\tau+\hat q_2))}\tau^2 + 2\frac{\alpha}{\gamma} \partial_{{\mathcal{Q}}_{W}(v)} \psi_{P_{\text{out}}}(q_K(q_2,\mathcal{Q}_W);r_K).
\end{align}
From the last equation in \eqref{eq:NSB_equations_generic_ch} we see that $\hat q_2$ cannot diverge more than $O(\alpha)$. Thanks to the above approximation and the first equation of \eqref{eq:NSB_equations_generic_ch} this entails that $\mathcal{Q}_W(v)$ is approaching $1$ exponentially fast in $\alpha$, which in turn implies $\tau$ is diverging exponentially in $\alpha$. As a consequence
\begin{align}
    \frac{\tau^2}{\tau(\tau+\hat q_2)}\sim 1.
\end{align}
Furthermore, one also has
\begin{align}
    \frac{1}{\alpha}[\iota(\tau)-\iota(\tau+\hat q_2)]=-\frac{1}{4\alpha}\int_\tau^{\tau+\hat q_2}  {\rm mmse}_S(t)\,dt\approx-\frac{C(\gamma)}{4\alpha}\log (1+\frac{\hat q_2}{\tau})\xrightarrow[]{\alpha\to\infty}0,
\end{align}as $\frac{\hat q_2}{\tau}$ vanishes with exponential speed in $\alpha$.

Concerning the function $\psi_{P_W}$, given that it is realted to a Bayes-optimal scalar Gaussian channel, and its SNRs $\hat{\mathcal{Q}}_W(v)$ are all diverging one can compute the integral by saddle point, which is inevitably attained at the ground truth:
\begin{align}
    \psi_{P_W}(\hat{\mathcal{Q}}_W(v))&-\frac{\hat{\mathcal{Q}}_W(v)\mathcal{Q}_W(v)}{2}\approx \EE_{w^0}\ln\int dP_W(w)\mathbbm{1}(w=w^0)\nonumber\\
    &+\EE\Big[
    (\sqrt{\hat{\mathcal{Q}}_W(v)}\xi+\hat{\mathcal{Q}}_W(v)w^0)w^0-\frac{\hat{\mathcal{Q}}_W(v)}{2}(w^0)^2
    \Big]-\frac{\hat{\mathcal{Q}}_W(v)(1-o_\alpha(1))}{2}=-
    \mathcal{H}(W)+o_\alpha(1).
\end{align}
Considering that $\psi_{P_{\text{out}}}(q_K(q_2,\mathcal{Q}_W);r_K)\xrightarrow[]{\alpha\to\infty}\psi_{P_{\text{out}}}(r_K;r_K)$, and using \eqref{eq:MI_generic_channel}, it is then straightforward to check that our RS version of the MI saturates to the entropy of the prior $P_W$ when $\alpha\to\infty$:
\begin{align}
\label{eq:large_alpha_limit}
    -\frac{\alpha}{\gamma}\text{extr} f_{\rm RS}^{\alpha,\gamma} +\frac{\alpha}{\gamma} \EE_{\lambda}  \int dy P_{\text{out}}(y|\lambda) \ln P_{\text{out}}(y|\lambda)\xrightarrow[]{\alpha\to\infty}
    \mathcal{H}(W).
\end{align}

\section{Extension of GAMP-RIE to arbitrary activation} \label{app:GAMP}

\begin{algorithm}[t]
   \caption{GAMP-RIE for training shallow neural networks with arbitrary activation}
   \label{alg:gamp}
\begin{algorithmic}\label{algo}
    \STATE {\bfseries Input:} Fresh data point $\bx_{\text{test}}$ with unknown associated response $y_{\text{test}}$, dataset $\mathcal{D}=\{(\bx_\mu, y_\mu)\}_{\mu=1}^n$. 
    \STATE {\bfseries Output:} Estimator $\hat y_{\text{test}}$ of $y_{\text{test}}$.
    \STATE Estimate $y^{(0)} := \mu_0 \bv^{\intercal} \bm{1}/\sqrt{k}$ as 
    \begin{equation*}
        \hat y^{(0)} = \frac{1}{n}\sum_{\mu} y_\mu ;
    \end{equation*}
    \STATE  Estimate $\langle \bW^{\intercal} \bv \rangle/\sqrt k$ using (\ref{eq:mmse-S1}).
    \STATE Estimate the $\mu_1$ term in the Hermite expansion (\ref{eq:hexpan}) as
    \begin{align*}
        \hat y_\mu^{(1)} &= \mu_1 \frac{ \langle \bv^\intercal \bW \rangle \bx_\mu }{\sqrt{kd}} ;
    \end{align*}
    \STATE  Compute 
    \begin{align*}
        \tilde y_\mu &=  \frac{y_\mu - \hat y_\mu^{(0)} - \hat y_\mu^{(1)}}{\mu_2/2} ; \qquad \tilde \Delta = \frac{\Delta + g(1)}{\mu_2^2/4} ;
    \end{align*}
    \STATE  Input $\{(\bx_\mu, \tilde y_\mu)\}_{\mu=1}^n$ and $\tilde \Delta$ into Algorithm 1 in \cite{maillard2024bayes} to estimate $\langle \bW^\intercal (\bv) \bW \rangle$;
    \STATE  Output 
    \begin{align} \label{eq:output_GAMP_RIE}
        \hat y_{\text{test}} = \hat y^{(0)} + \mu_1 \frac{ \langle \bv^\intercal \bW \rangle \bx_{\text{test}}}{\sqrt{kd}} + \frac{\mu_2}{2} \frac{1}{d\sqrt k} \Tr[ (\bx_{\text{test}} \bx_{\text{test}}^\intercal - \I ) \langle \bW^\intercal (\bv) \bW \rangle ].
    \end{align}
\end{algorithmic}
\end{algorithm}

For simplicity, let us consider $P_{\rm out}(y\mid\lambda)=\exp(-\frac1{2\Delta}(y-\lambda)^2)/\sqrt{2\pi\Delta}$, which entails:
\begin{align}
y_{\mu} \mid (\btheta^0,\bx_\mu) \overset{\rm{d}}{=} \frac{\bv^{\intercal}}{\sqrt k} \sigma \Big( \frac{\bW^0 \bx_\mu}{\sqrt d} \Big) + \sqrt{\Delta} \,z_\mu, \quad \mu =1\dots,n,
\end{align}
where $z_\mu$ are i.i.d. standard Gaussian random variables and $\overset{\rm d}{{}={}}$ means equality in law. Expanding $\sigma$ in the Hermite polynomial basis we have
\begin{align}\label{eq:hexpan}
y_\mu \mid (\btheta^0,\bx_\mu) \overset{\rm{d}}{=} \mu_0\frac{\bv^{\intercal} \bm{1}_k}{\sqrt{k}}+ \mu_1 \frac{\bv^{\intercal} \bW^0 \bx_\mu}{\sqrt{kd}} + \frac{\mu_2}{2} \frac{\bv^{\intercal}}{\sqrt k} \He_2 \Big( \frac{\bW^0 \bx_\mu}{\sqrt d} \Big) + \dots + \sqrt{\Delta} z_\mu 
\end{align}
where $\dots$ represents the terms beyond second order. 
Without loss of generality, for this choice of output channel we can set $\mu_0 = 0$ as discussed in App.~\ref{app:non-centered}. For low enough $\alpha$ it is reasonable to assume that higher order terms in $\dots$ cannot be learnt given quadratically many samples and, as a result, play the role of effective noise, which we assume independent of the first three terms. We shall see that this reasoning actually applies to the extension of the GAMP-RIE we derive, which plays the role of a ``smart'' spectral algorithm, regardless of the value of $\alpha$.
Therefore, these terms accumulate in an asymptotically Gaussian noise thanks to the central limit theorem (it is a projection of a centred function applied entry-wise to a vector with i.i.d.\ entries), with variance $g(1)$ (see \eqref{eq:g_func}). We thus obtain the effective model 
\begin{align}
y_\mu \mid (\btheta^0,\bx_\mu) \overset{\rm{d}}{=}  \mu_1 \frac{\bv^{\intercal} \bW^0 \bx_\mu}{\sqrt{kd}} + \frac{\mu_2}{2} \frac{\bv^{\intercal}}{\sqrt k} \He_2 \Big( \frac{\bW^0 \bx_\mu}{\sqrt d} \Big) + \sqrt{\Delta + g(1) } \, z_\mu .
\end{align}
The first term in this expression can be learnt with vanishing error given quadratically many samples (Remark \ref{rem:linear}), thus can be ignored. This further simplifies the model to
\begin{align}
\bar y_\mu :=  y_\mu - \mu_1 \frac{\bv^{\intercal} \bW^0 \bx_\mu}{\sqrt{kd}}\overset{\rm d}{{}={}} \frac{\mu_2}{2} \frac{\bv^{\intercal}}{\sqrt k} \He_2 \Big( \frac{\bW^0 \bx_\mu}{\sqrt d} \Big) + \sqrt{\Delta + g(1) } \, z_\mu,
\end{align}
where $\bar y_\mu$ is $ y_\mu$ with the (asymptotically) perfectly learnt linear term removed, and the last equality in distribution is again conditional on $(\btheta^0,\bx_\mu)$. From the formula
\begin{align}
\frac{\bv^{\intercal}}{\sqrt{k}} \He_2 \Big( \frac{\bW^0 \bx_\mu}{\sqrt d} \Big) = \Tr \frac{\bW^{0\intercal} (\bv ) \bW^0}{d\sqrt{k}} \bx_\mu \bx_\mu^\intercal - \frac{\bv^{\intercal} \bm{1}_k}{\sqrt{k}}\approx \frac{1}{\sqrt{k}d} \Tr[( \bx_\mu \bx_\mu^\intercal - \I_d)\bW^{0\intercal} (\bv ) \bW^0 ],
\end{align}
where $\approx$ is exploiting the concentration $\Tr \bW^{0\intercal} (\bv ) \bW^0 /(d\sqrt{k}) \to \bv^{\intercal} \bm{1}_k/\sqrt{k}$,
and the Gaussian equivalence property that $\bM_\mu:=(\bx_\mu \bx_\mu^\intercal - \I_d)/\sqrt{d}$ behaves like a GOE sensing matrix, i.e., a symmetric matrix whose upper triangular part has i.i.d. entries from $\mathcal{N}(0,(1+\delta_{ij})/d)$ \cite{maillard2024bayes}, the model can be seen as a GLM with signal $\bar\bS^0_2 := \bW^{0\intercal} (\bv ) \bW^0/\sqrt{kd}$:
\begin{align}\label{eq:Maillard_starting_point}
    y^{\rm GLM}_\mu=\frac{\mu_2}{2}\Tr [\bM_\mu \bar\bS^0_2] +\sqrt{\Delta+g(1)}\,z_\mu.
\end{align}
Starting from this equation, the arguments of App.~\ref{app:replicas} and \cite{maillard2024bayes}, based on known results on the GLM \cite{barbier2019glm} and matrix denoising \cite{barbier2022statistical,maillard2022perturbative,matrix_inference_Barbier}, allow us to obtain the free entropy of such matrix sensing problem. The result is consistent with the $\mathcal{Q}_W \equiv 0$ solution of the saddle point equations obtained from the replica method in App.~\ref{app:replicas}, which, as anticipated, corresponds to the case where the Hermite polynomial combinations of the signal following the second one are not learnt.

Note that, as supported by the numerics, the model actually admits specialisation when $\alpha$ is big enough, hence the above equivalence cannot hold on the whole phase diagram at the information theoretic level. In fact, if specialisation occurs one cannot consider the $\dots$ terms in \eqref{eq:hexpan} as noise uncorrelated with the first ones, as the model is aligning with the actual teacher's weights, such that it learns all the successive terms at once.

We now assume that this mapping holds at the algorithmic level, namely, that we can process the data algorithmically as if they were coming from the identified GLM, and thus try to infer the signal $\bar\bS_2^0 = \bW^{0\intercal} (\bv ) \bW^0/\sqrt {kd}$ and construct a predictor from it. Based on this idea, we propose Algorithm~\ref{alg:gamp} that can indeed reach the performance predicted by the $\mathcal{Q}_W \equiv 0$ solution of our replica theory. 

\begin{remark}\label{rem:linear}
In the linear data regime, where $n/d$ converges to a fixed constant $\alpha_1$, only the first term in (\ref{eq:hexpan}) can be learnt while the rest behaves like noise. By the same argument as above, the model is equivalent to 
\begin{align}\label{eq:linear-model}
     y_\mu = \mu_1 \frac{\bv^{\intercal} \bW^0 \bx_\mu}{\sqrt{kd}} + \sqrt{\Delta + \nu - \mu_0^2 - \mu_1^2} \, z_\mu,
\end{align}
where $\nu = \EE_{z\sim\calN(0,1)}\sigma^2(z)$.
This is again a GLM with signal $\bS_1^0 = \bW^{0\intercal} \bv /\sqrt k$ and Gaussian sensing vectors $\bx_{\mu}$. Define $q_1$ as the limit of $\bS_1^{a\intercal}\bS_1^b /d$ where $\bS_1^a, \bS_1^b$ are drawn independently from the posterior. With $k \rightarrow \infty$, the signal converges in law to a standard Gaussian vector. Using known results on GLMs with Gaussian signal \cite{barbier2019glm}, we obtain the following  equations characterising $q_1$:
\begin{align*}
q_1 & = \frac{\hat q_1}{\hat q_1 + 1}, \qquad
\hat q_1 = \frac{\alpha_1}{1 + \Delta_1 - q_1},\quad \text{where} \quad \Delta_1 = \frac{\Delta + \nu - \mu_0^2 - \mu_1^2}{\mu_1^2}.
\end{align*}
In the quadratic data regime, as $\alpha_1=n/d$ goes to infinity, the overlap $q_1$ converges to $1$ and the first term in (\ref{eq:hexpan}) is learnt with vanishing error.

Moreover, since $\bS_1^0$ is asymptotically Gaussian, the linear problem (\ref{eq:linear-model}) is equivalent to denoising the Gaussian vector $(\bv^{\intercal} \bW^0 \bx_\mu/\sqrt{kd})_{\mu=0}^n$ whose covariance is known as a function of $\bX = (\bx_1 , \dots, \bx_n) \in \R^{d\times n}$. This leads to the following simple MMSE estimator for $\bS_1^0$:
\begin{align}\label{eq:mmse-S1}
    \langle \bS_1^0 \rangle = \frac{1}{\sqrt{d \Delta_1}}\left(\mathbf I + \frac{1}{d \Delta_1} \bX \bX^\intercal \right)^{-1} \bX \by
\end{align}
where $\by = (y_1, \dots, y_n)$. Note that the derivation of this estimator does not assume the Gaussianity of $\bx_\mu$.
\end{remark}

\begin{remark}
The same argument can be easily generalised for general $P_{\text{out}}$, leading to the following equivalent GLM in the universal $\calQ_W^*\equiv 0$ phase of the quadratic data regime:
\begin{align}
    y_\mu^{\rm GLM} \sim \tilde P_{\text{out}}(\cdot\mid \Tr [\bM_\mu \bar\bS^0_2] ), \quad \text{where} \quad  \tilde P_{\text{out}}(y|x) := \mathbb E_{z \sim \mathcal N(0,1)} P_{\text{out}}\Big(y \mid  \frac{\mu_2}{2} x + z\sqrt{g(1)} \Big),
\end{align}
and $\bM_\mu$ are independent GOE sensing matrices.
\end{remark}

\begin{remark}
One can show that the system of equations $({\rm S})$ in \eqref{NSB_equations_gaussian_ch} with $\mathcal{Q}_W(\mathsf{v})$ all set to $0$ (and consequently $\tau=0$) can be mapped onto the fixed point of the state evolution equations (92), (94) of the GAMP-RIE in \cite{maillard2024bayes} up to changes of variables. This confirms that when such a system has a unique solution, which is the case in all our tests, the GAMP-RIE asymptotically matches our universal solution. Assuming the validity of the aforementioned effective GLM, a potential improvement for discrete weights could come from a generalisation of GAMP which, in the denoising step, would correctly exploit the discrete prior over inner weights rather than using the RIE (which is prior independent). However, the results of \cite{barbier2024phase} suggest that optimally denoising matrices with discrete entries is hard, and the RIE is the best efficient procedure to do so. Consequently, we tend to believe that improving GAMP-RIE in the case of discrete weights is out of reach without strong side information about the teacher, or exploiting non-polynomial-time algorithms (see Appendix~\ref{app:hardness}).
\end{remark}

\section{Algorithmic complexity of finding the specialisation solution\label{app:hardness}}

\begin{figure}[p]
\begin{center}
\centerline{
\includegraphics[width=.49\linewidth,trim={0 0 0 0},clip]{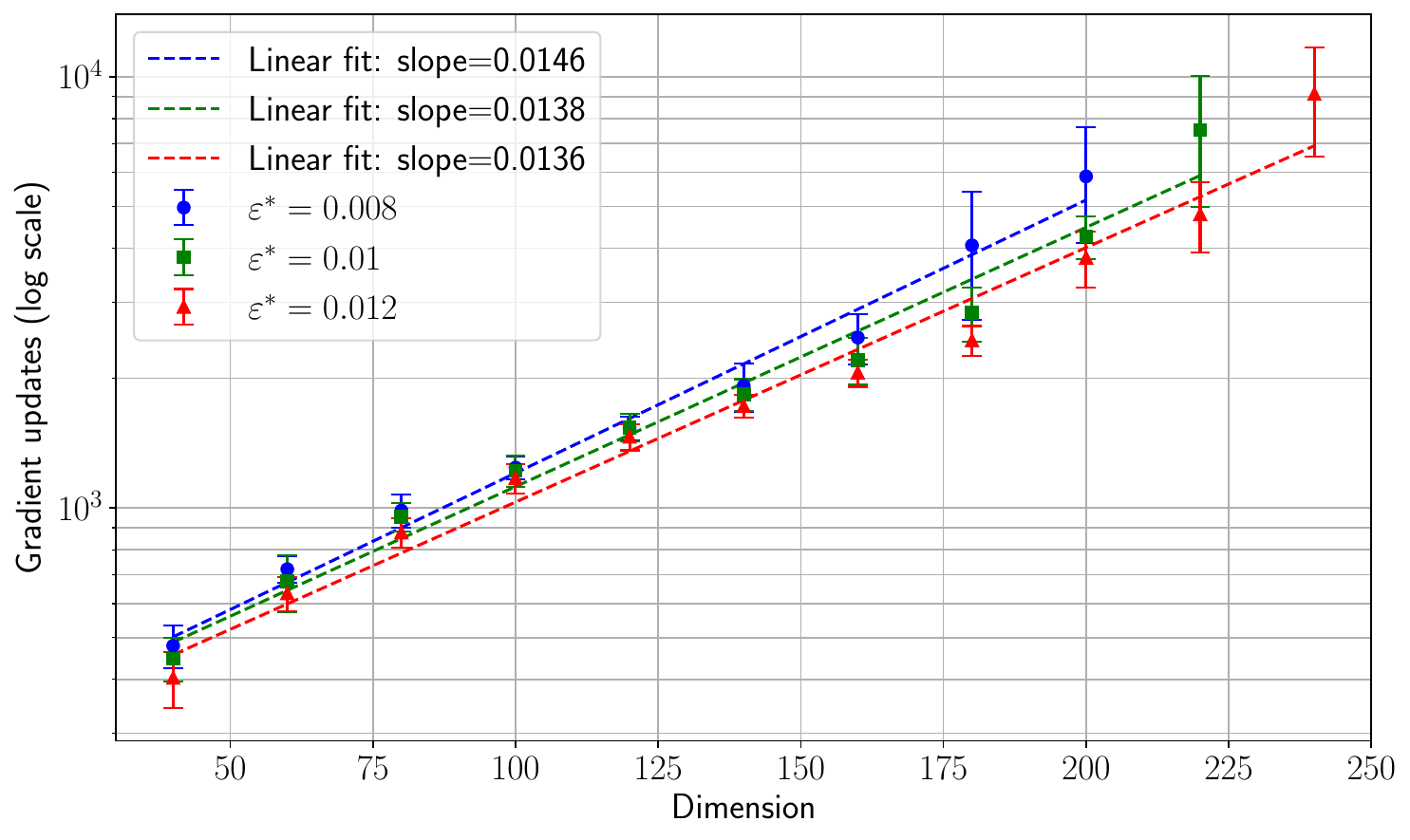}
\includegraphics[width=.49\linewidth,trim={0 0 0 0},clip]{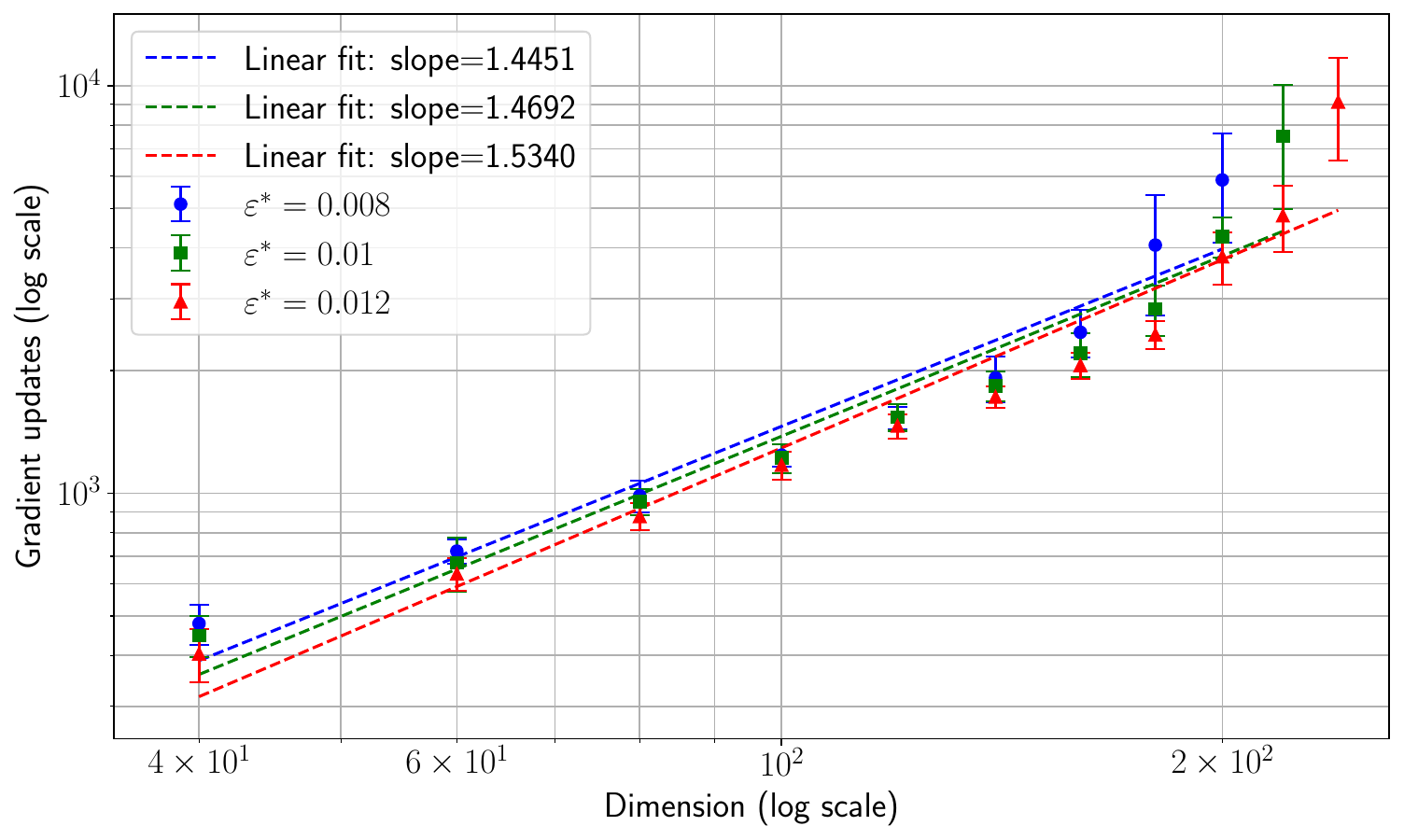}
}
\vspace{1pt} 
\centerline{
\includegraphics[width=.49\linewidth,trim={0 0 0 0},clip]{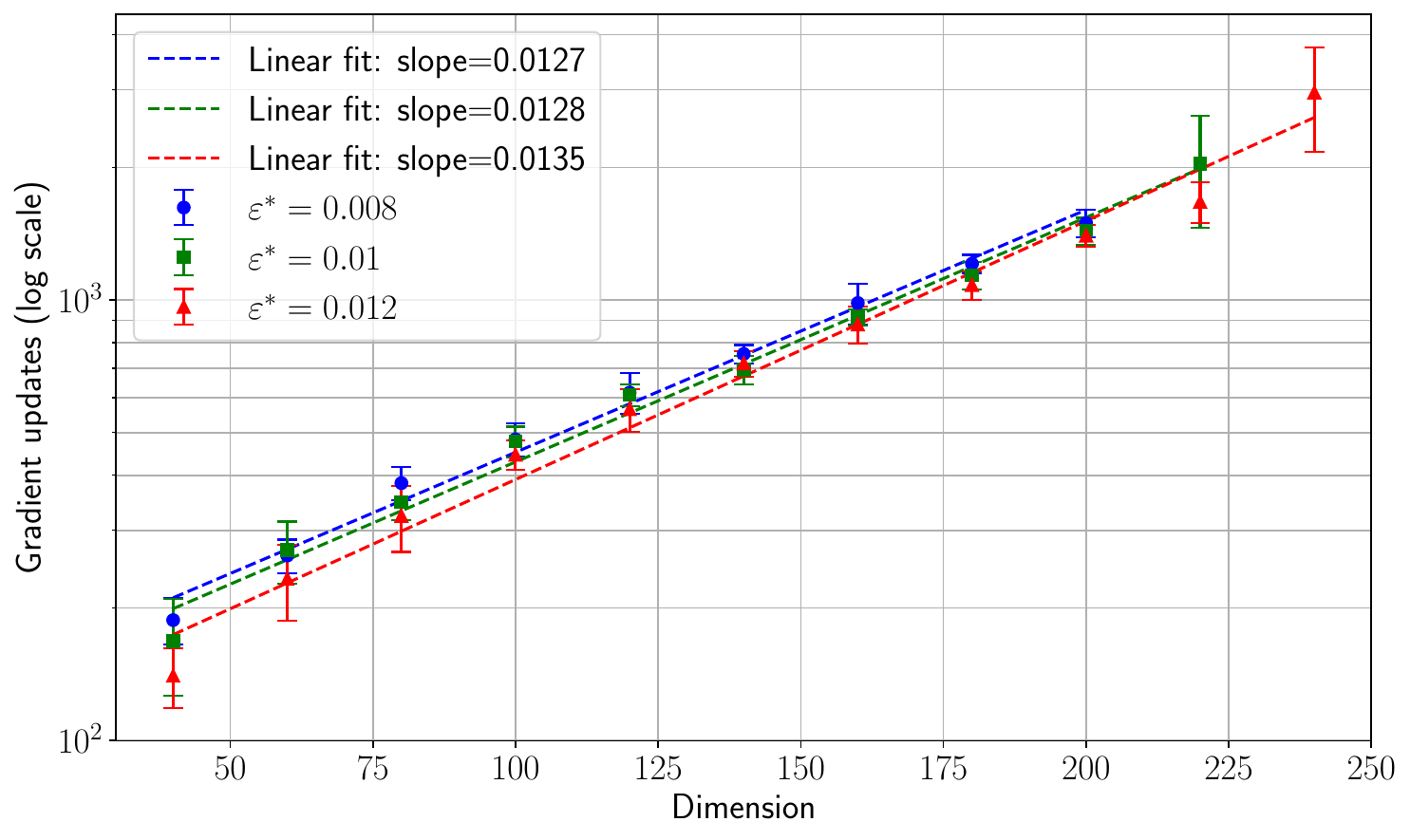}
\includegraphics[width=.49\linewidth,trim={0 0 0 0},clip]{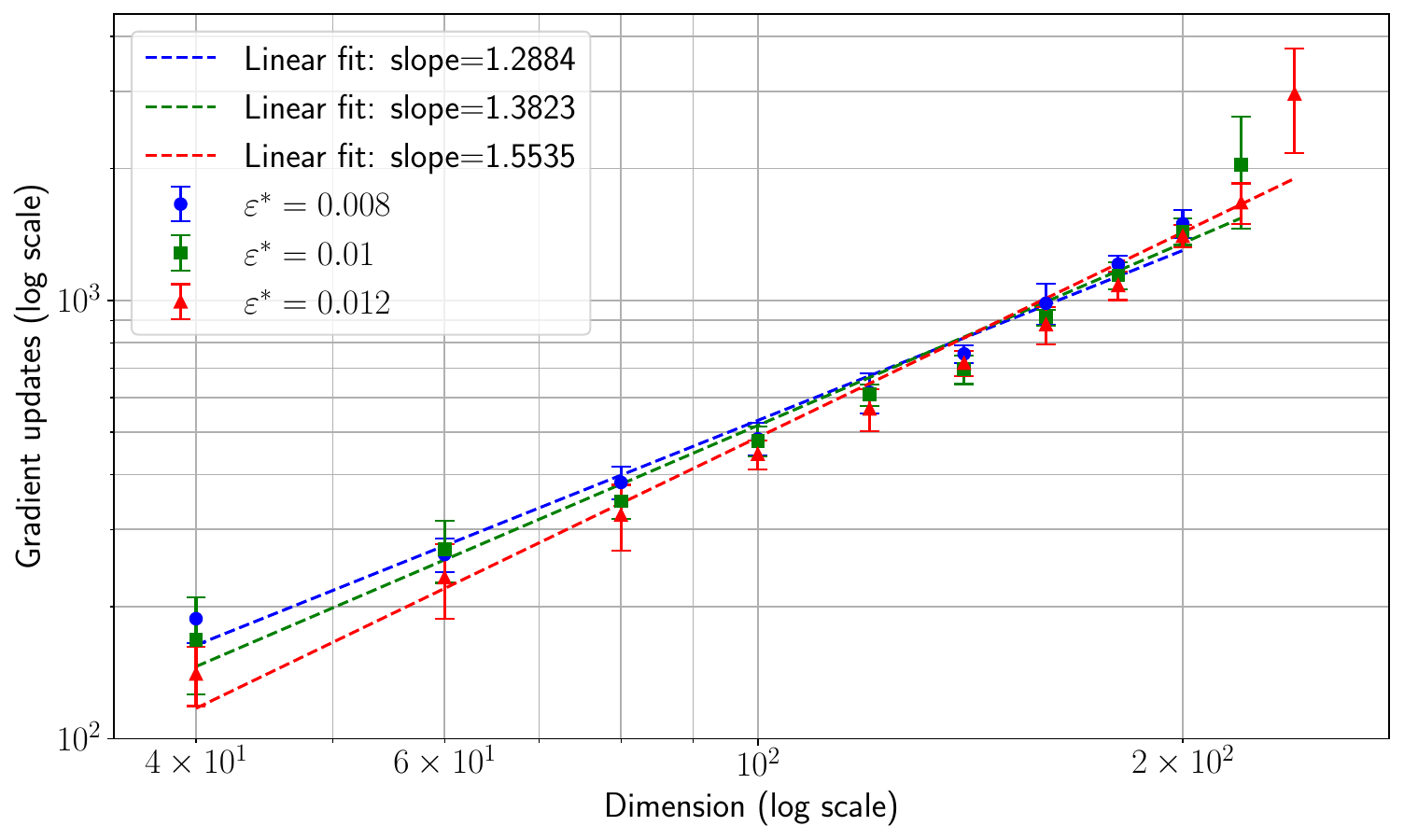}
}
\vspace{1pt} 
\centerline{
\includegraphics[width=.49\linewidth,trim={0 0 0 0},clip]{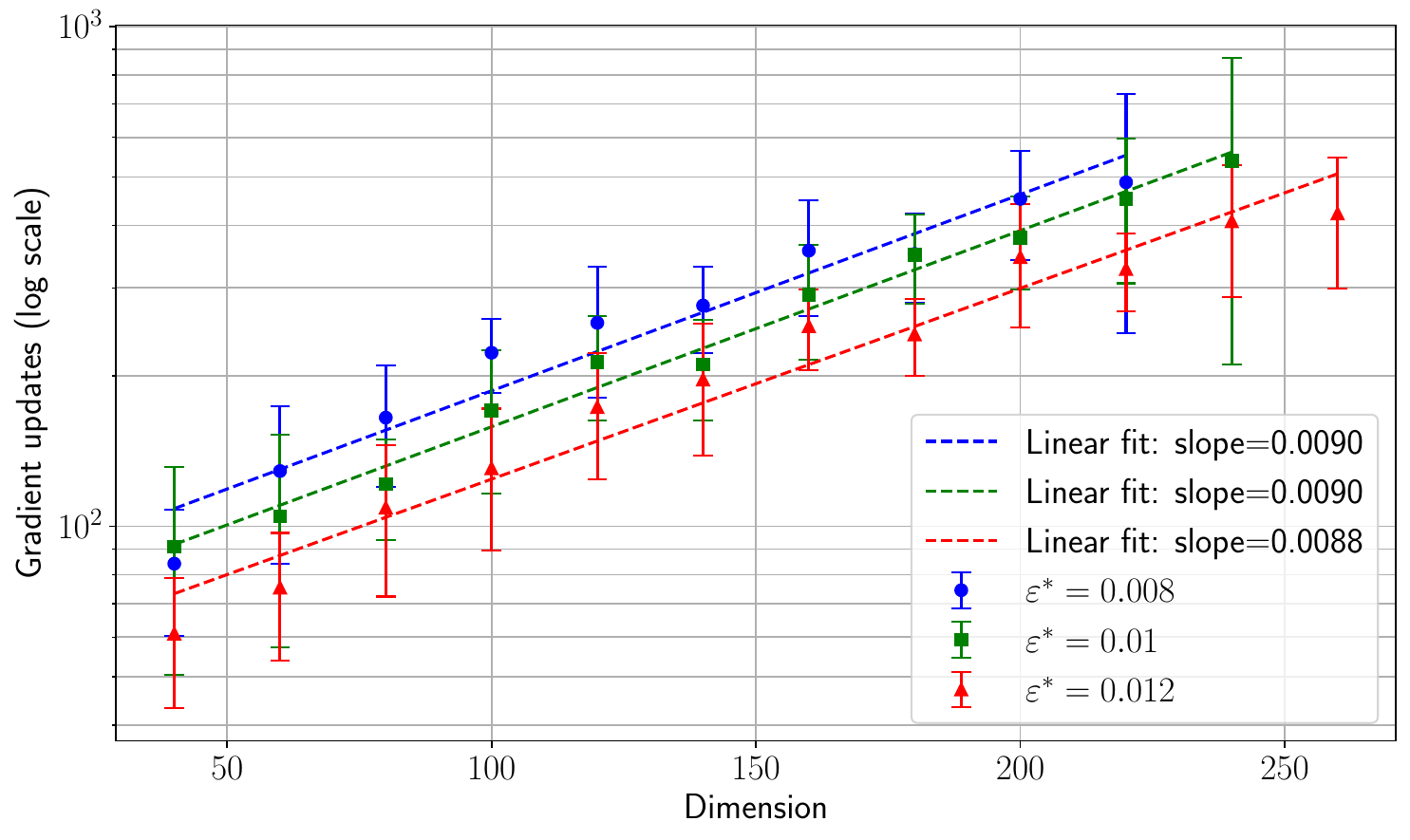}
\includegraphics[width=.49\linewidth,trim={0 0 0 0},clip]{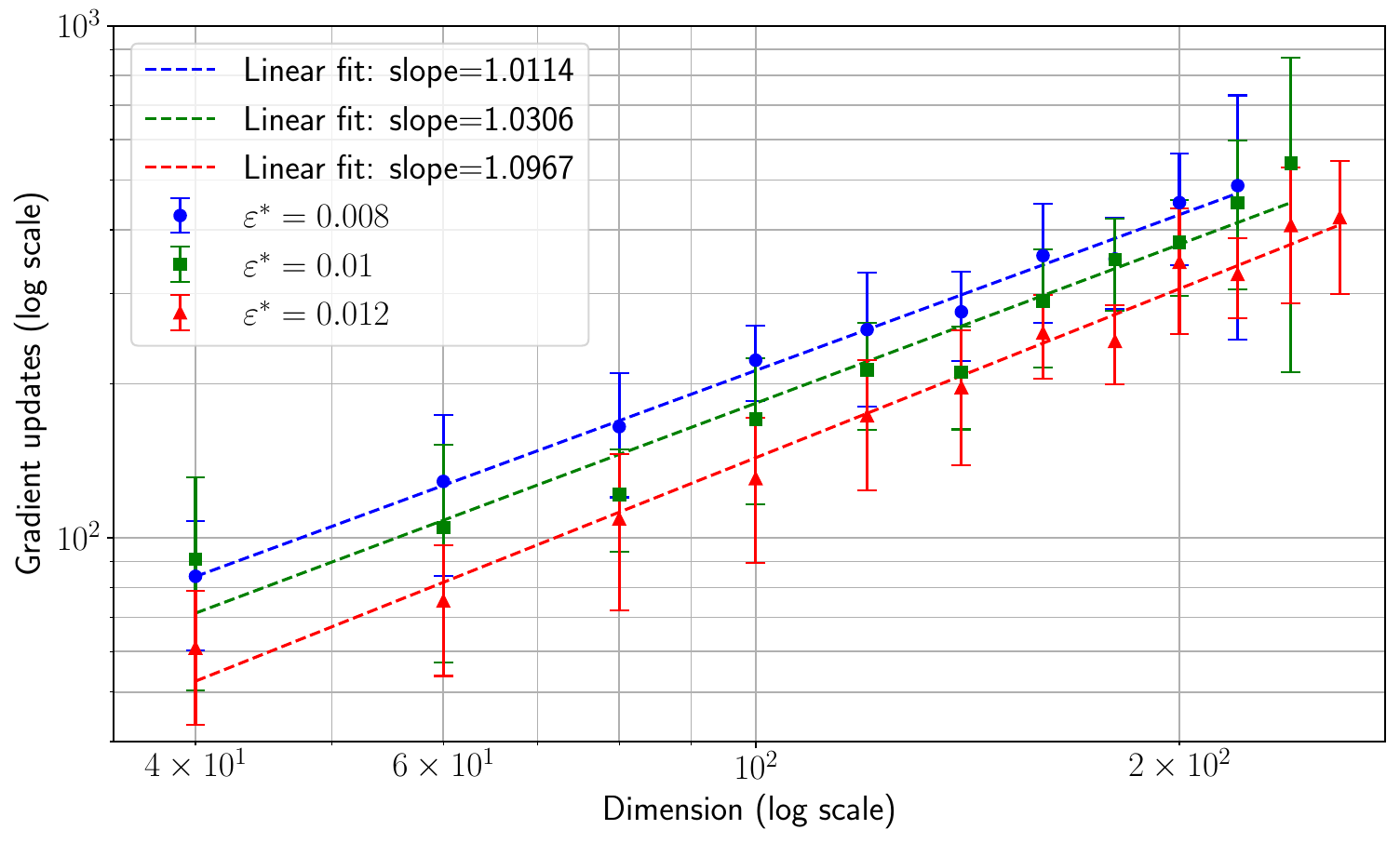}
}
\vspace{-10pt}
    \caption{
    Semilog (\textbf{Left}) and log-log (\textbf{Right}) plots of the number of gradient updates needed to achieve a test loss below the threshold $\varepsilon^*< \varepsilon^{\rm uni}$. Student network trained with ADAM with optimised batch size for each point. The dataset was generated from a teacher network with ReLU activation and parameters $\Delta = 10^{-4}$ for the Gaussian noise variance of the linear readout, $\gamma=0.5$ and $\alpha=5.0$ for which $\varepsilon^{\rm opt}-\Delta=1.115 \times 10^{-5}$.
     Points are obtained averaging over 10 teacher/data instances with error bars representing the standard deviation.
    Each row corresponds to a different distribution of the readouts, kept fixed during training. \textbf{Top}: homogeneous readouts, for which the error of the universal branch is $\varepsilon^{\rm uni}-\Delta=1.217\times 10^{-2}$. \textbf{Centre}: Rademacher readouts, for which $\varepsilon^{\rm uni}-\Delta=1.218\times 10^{-2}$.
    \textbf{Bottom}: Gaussian readouts, for which $\varepsilon^{\rm uni}-\Delta=1.210\times 10^{-2}$. The quality of the fits can be read from Table~\ref{tab:adam}.
}
    \label{fig:hardness_adam}
\end{center}
\vskip -0.3in
\end{figure}

\begin{table}[p]
    \centering
    \begin{tabular}{lc|c|c|c|c|c|c|}
            &        & \multicolumn{3}{c|}{$\chi^2$ exponential fit} & \multicolumn{3}{c|}{$\chi^2$ power law fit}\\
        \hline
        Readouts    &$\varepsilon^*=$& $0.008$ & $0.010$ & $0.012$ & $0.008$ & $0.010$ & $0.012$ \\
        \hline
         Homogeneous&&$\bm{5.57}$ & $\bm{9.00}$ & $\bm{21.1}$ &$32.3$&$26.5$&$61.1$ \\
         Rademacher&& $\bm{4.51}$ & $\bm{6.84}$ & $\bm{12.7}$ &$12.0$&$17.4$ &$16.0$\\
         Uniform $[-\sqrt{3},\sqrt{3}]$&& $\bm{5.08}$ & $\bm{1.44}$ & ${4.21}$ &$8.26$ &$8.57$ &$\bm{3.82}$\\
         Gaussian&&$2.66$&$\bm{0.76}$&$3.02$ & $\bm{0.55}$&$2.31$ &$\bm{1.36}$\\
    \end{tabular}
    \caption{$\chi^2$ test for exponential and power-law fits for the time needed by ADAM to reach the thresholds $\varepsilon^*$, for various priors on the readouts. Fits are displayed in \figurename~\ref{fig:hardness_adam}. Smaller values of $\chi^2$ (in bold, for given threshold and readouts) indicate a better compatibility with the hypothesis.}
    \label{tab:adam}
\end{table}

\begin{figure}[t]
    \centering
    \includegraphics[width=0.4\linewidth]{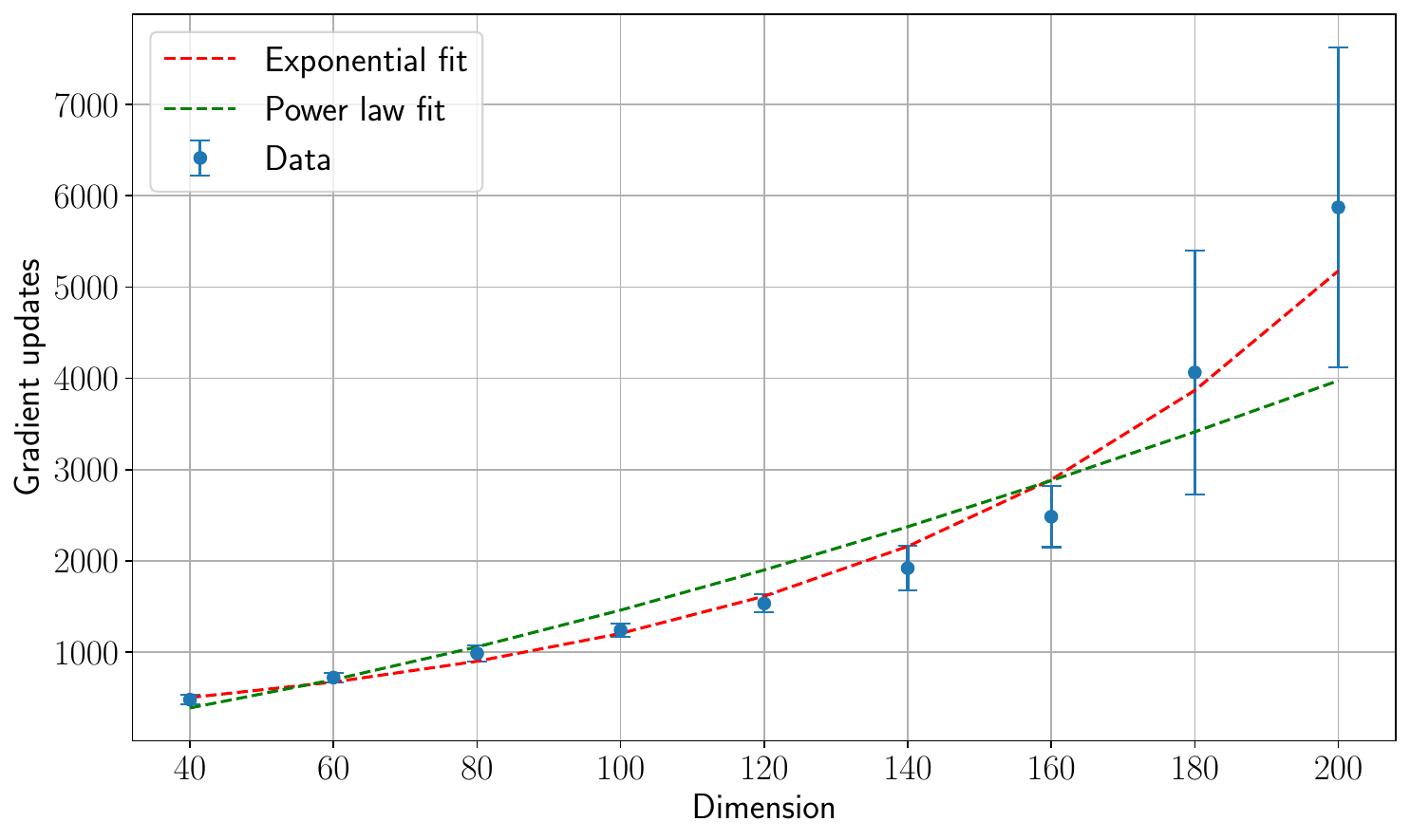}
    \includegraphics[width=0.4\linewidth]{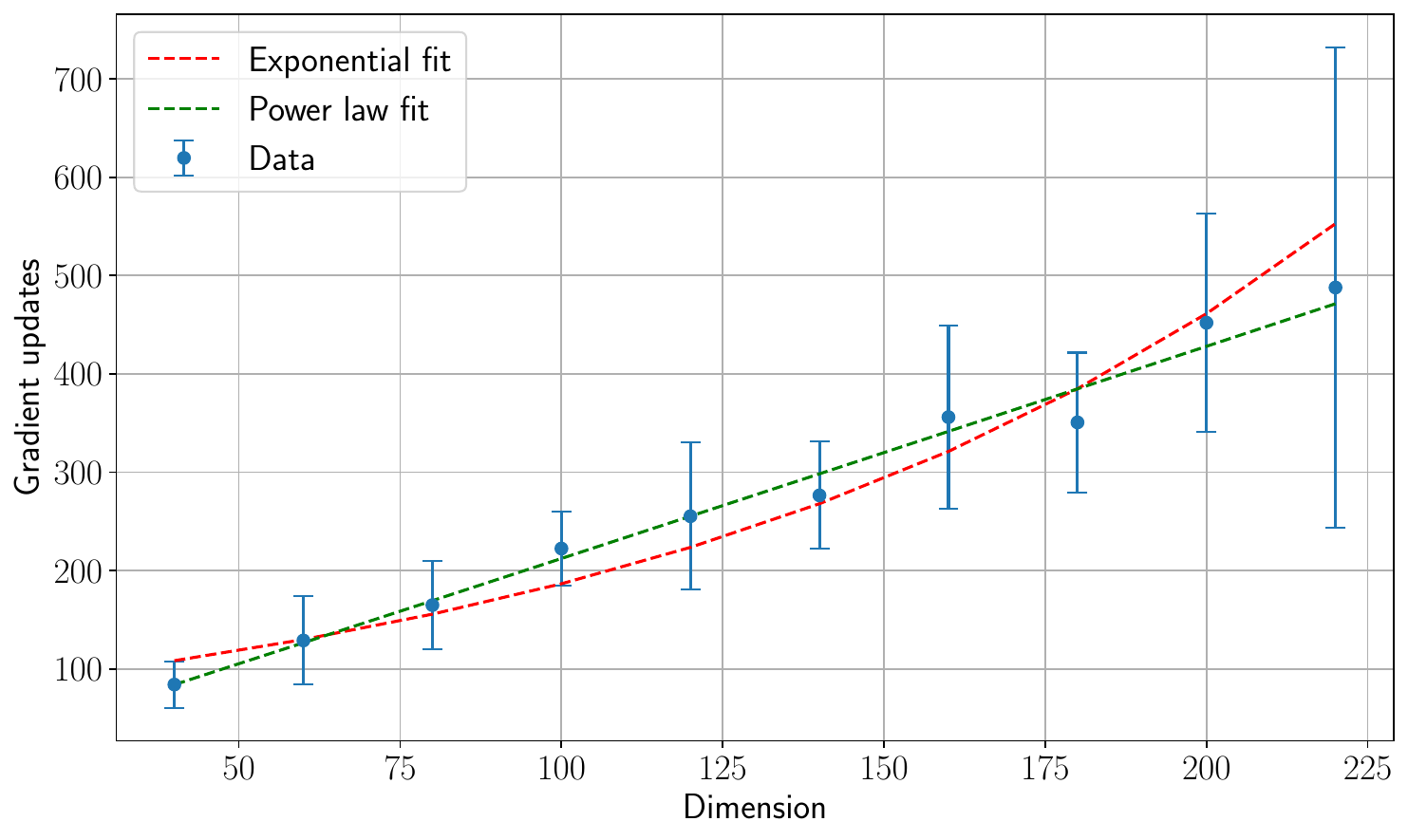}
    \caption{Same as in Fig.~\ref{fig:hardness_adam}, but in linear scale for better visualisation, for homogeneous readouts (\textbf{Left}) and Gaussian readouts (\textbf{Right}), with threshold $\varepsilon^*=0.008$.}
    \label{fig:hardness_adam_linear}
\end{figure}

\begin{figure}[t]
    \centering
    \includegraphics[width=0.34\linewidth,trim={0 0 0 0},clip]{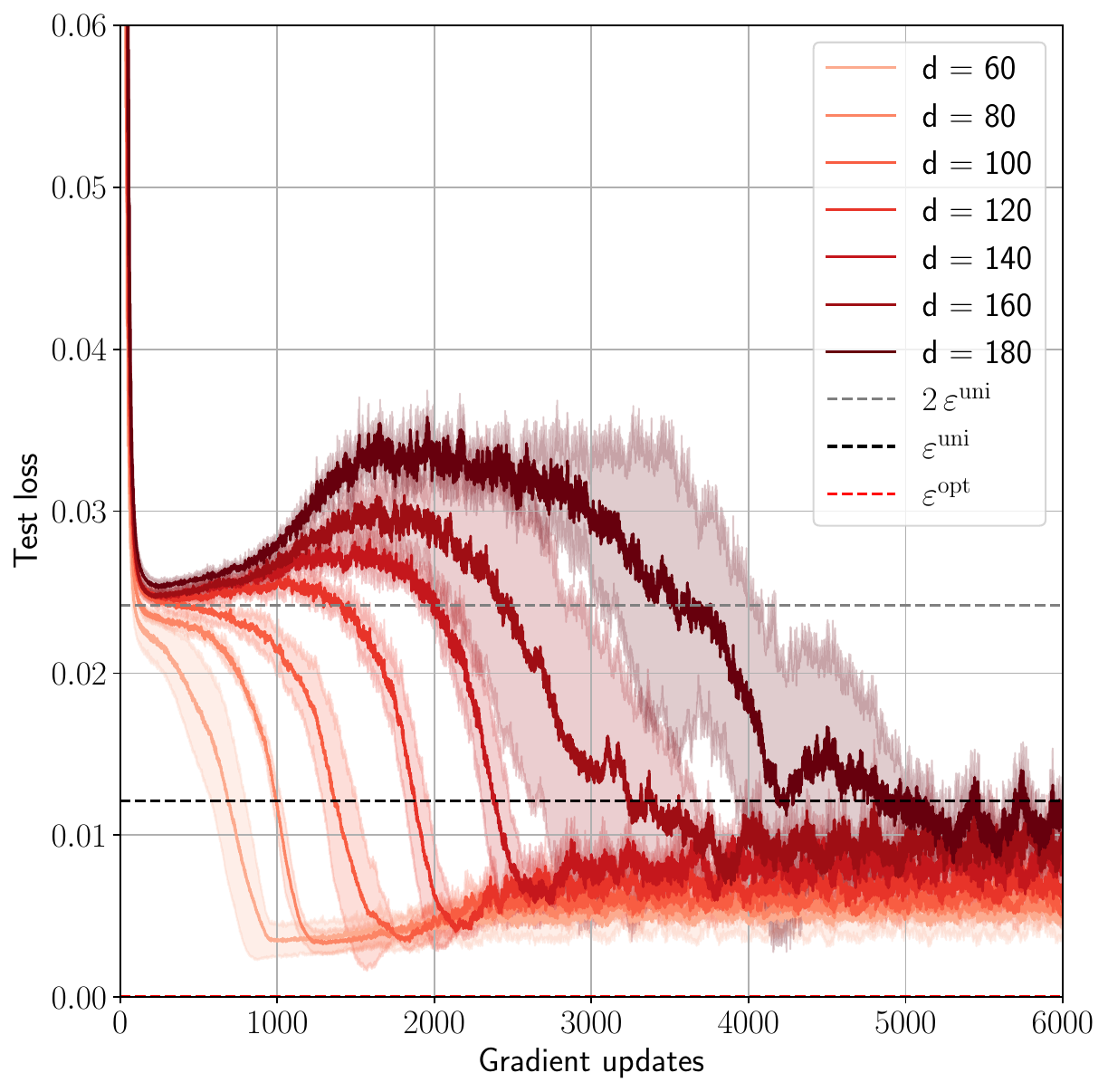}
    \includegraphics[width=0.325\linewidth,trim={0.9cm 0 0 0},clip]{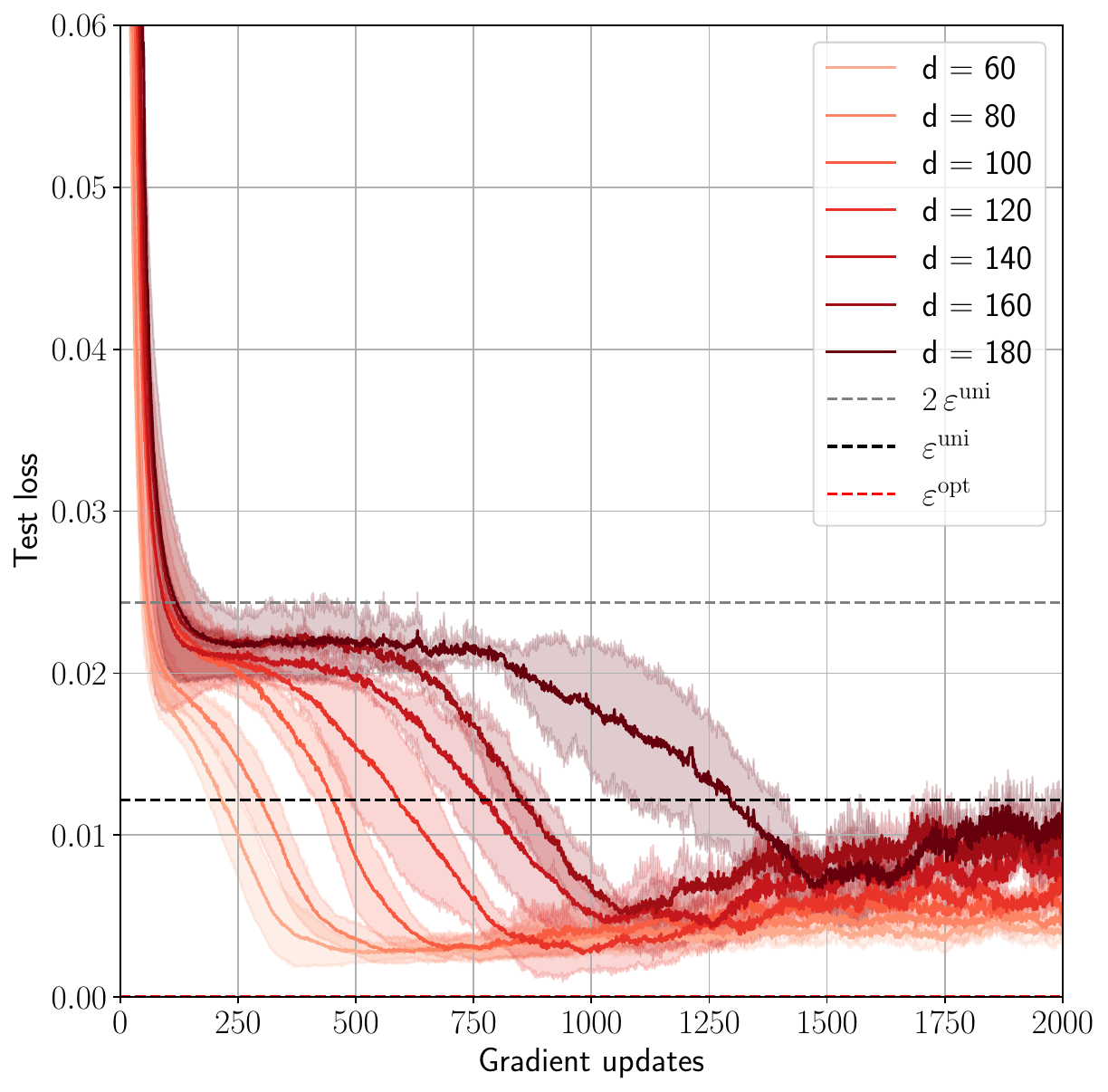}
    \includegraphics[width=0.32\linewidth,trim={0.9cm 0 0 0},clip]{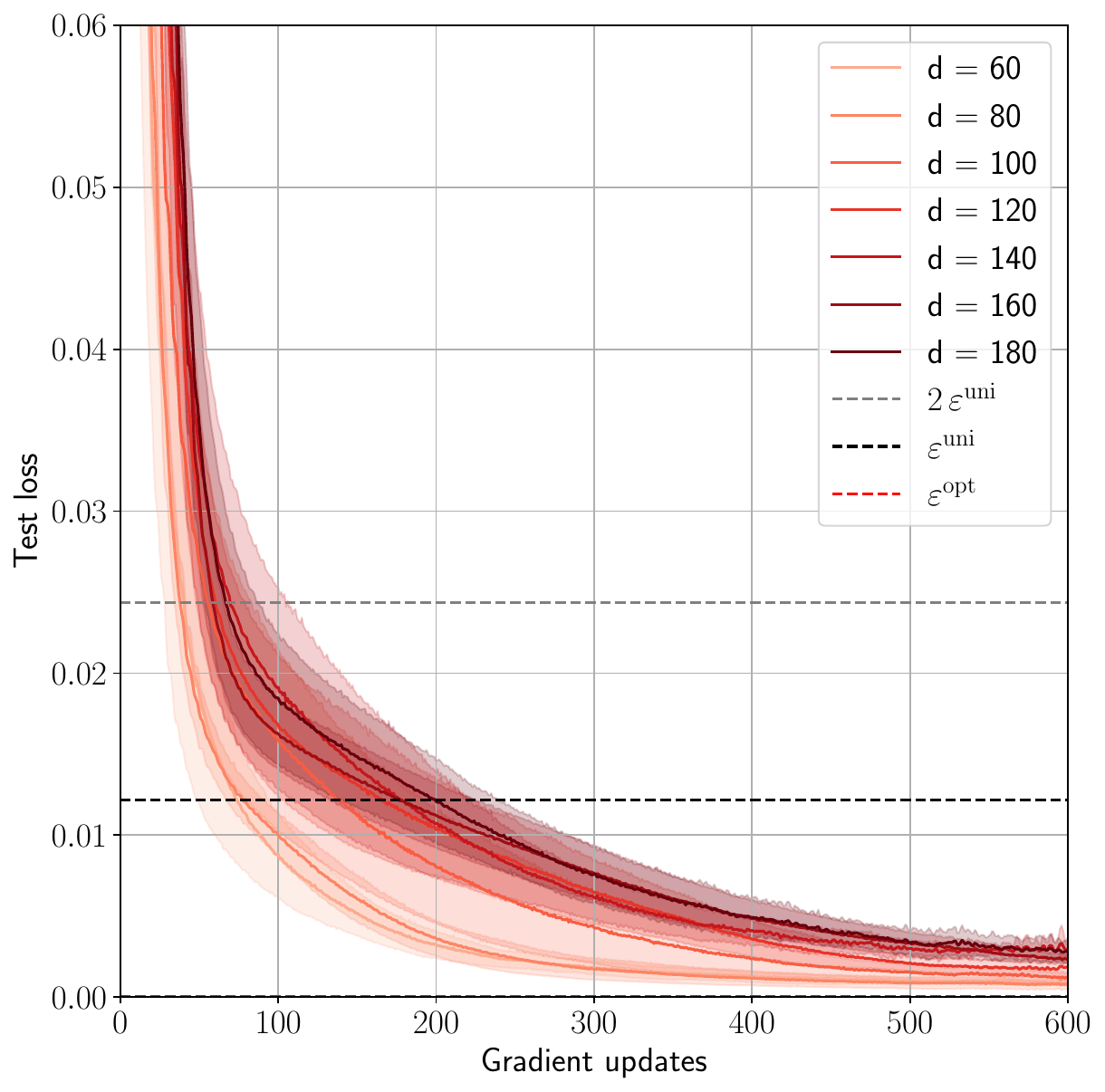}
    \caption{Trajectories of the generalisation error of neural networks trained with ADAM at fixed batch size $B=\lfloor n/4\rfloor$, learning rate 0.05, for ReLU activation with parameters $\Delta = 10^{-4}$ for the linear readout, $\gamma=0.5$ and $\alpha=5.0 > \alpha_{\rm sp}$ ($= 0.22,0.12,0.02$ for homogeneous, Rademacher and Gaussian readouts respectively). The error $\varepsilon^{\rm uni}$ is the mean-square generalisation error associated to the universal solution with overlap $\mathcal{Q}_W \equiv 0$.
    \textbf{Left}: Homogeneous readouts.
    \textbf{Centre}: Rademacher readouts.
    \textbf{Right}: Gaussian readouts. Readouts are kept fixed (and equal to the teacher's) in all cases during training. Points on the solid lines are obtained by averaging over 5 teacher/data instances, and shaded regions around them correspond to one standard deviation.
    }
    \label{fig:hardness_adam_runs}
\end{figure}

We now provide empirical evidence concerning the computational complexity to attain specialisation, namely to have one of the $\mathcal{Q}_W(\mathsf{v})>0$, or equivalently to beat the ``universal'' performance ($\mathcal{Q}_W(\mathsf{v})=0$ for all $\mathsf{v}\in\mathsf{V}$) in terms of generalisation error. We tested two algorithms that can find it in affordable computational time: ADAM with optimised batch size for every dimension tested (the learning rate is automatically tuned), and Hamiltonian Monte Carlo (HMC), both trying to infer a two-layer teacher network with Gaussian inner weights. 

\paragraph{ADAM}
We focus on ReLU activation, with $\gamma=0.5$, Gaussian output channel with low label noise ($\Delta=10^{-4}$) and $\alpha=5.0>\alpha_{\rm sp}$ ($= 0.22,0.12,0.02$ for homogeneous, Rademacher and Gaussian readouts respectively, thus we are deep in the specialisation phase in all the cases we report), so that the specialisation solution exhibits a very low generalisation error. We test the learnt model at each gradient update measuring the generalisation error with a moving average of 10 steps to smoothen the curves. Let us define $\varepsilon^{\rm uni}$ as the generalisation error associated to the overlap $\mathcal{Q}_W \equiv 0$, then fixing a threshold $ \varepsilon^{\rm opt} < \varepsilon^* < \varepsilon^{\rm uni}$, we define $t^*(d)$ the time (in gradient updates) needed for the algorithm to cross the threshold for the first time. We optimise over different batch sizes $B_p$ as follows: we define them as $B_p = \left\lfloor \frac{n}{2^p} \right\rfloor,\quad p = 2,3,\dots,\lfloor \log_2(n) \rfloor-1$. Then for each batch size, the student network is trained until the moving average of the test loss drops below $\varepsilon^*$ and thus outperforms the universal solution; we have checked that in such a scenario, the student ultimately gets close to the performance of the specialisation solution. The batch size that requires the least gradient updates is selected. We used the ADAM routine implemented in PyTorch. 

We test different distributions for the readout weights (kept fixed to $\bv $ during training of the inner weights). We report all the values of $t^*(d)$ in Fig.~\ref{fig:hardness_adam} for various dimensions $d$ at fixed $(\alpha,\gamma)$, providing an exponential fit $t^*(d) = \exp(a d + b)$ (left panel) and a power-law fit $t^*(d) = a d^b $ (right panel). We report the $\chi^2$ test for the fits in Table~\ref{tab:adam}.
We observe that for homogeneous and Rademacher readouts, the exponential fit is more compatible with the experiments, while for Gaussian readouts the comparison is inconclusive.

In Fig.~\ref{fig:hardness_adam_runs}, we report the test loss of ADAM as a function of the gradient updates used for training, for various dimensions and choice of the readout distribution (as before, the readouts are not learnt but fixed to the teacher's). Here, we fix a batch size for simplicity. For both the cases of homogeneous ($\bv=\bm{1}$) and Rademacher readouts (left and centre panels), the model experiences plateaux in performance increasing with the system size, in accordance with the observation of exponential complexity we reported above. The plateaux happen at values of the test loss comparable with twice the value for the Bayes error predicted by the universal branch of our theory (remember the relationship between Gibbs and Bayes errors reported in App.~\ref{app:gen_err}). The curves are smoother for the case of Gaussian readouts.

\paragraph{Hamiltonian Monte Carlo}

\begin{figure}[pb]
    \centering
    \includegraphics[width=0.31\linewidth,trim={1.1cm 0 0.3cm 0},clip]{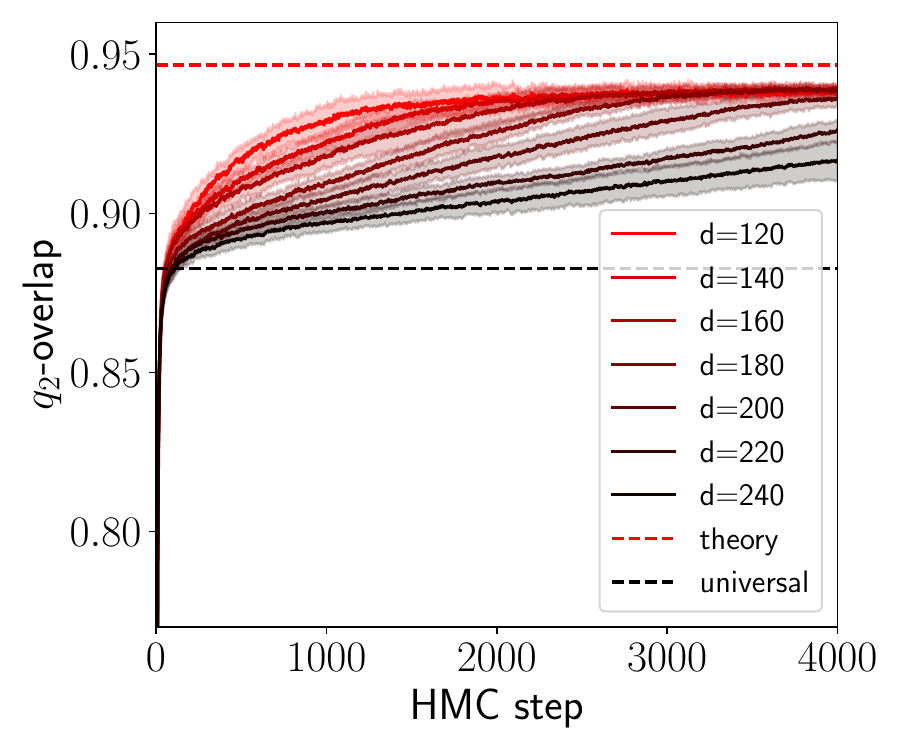}
    \includegraphics[width=0.31\linewidth,trim={1.1cm 0 0.3cm 0},clip]{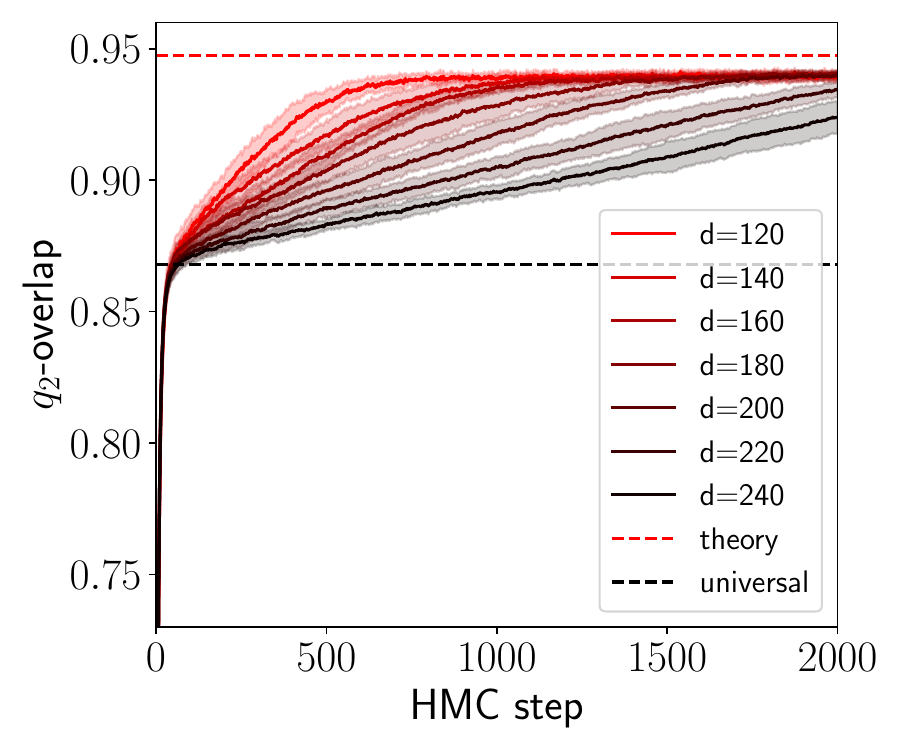}
    \includegraphics[width=0.31\linewidth,trim={1.1cm 0 0.3cm 0},clip]{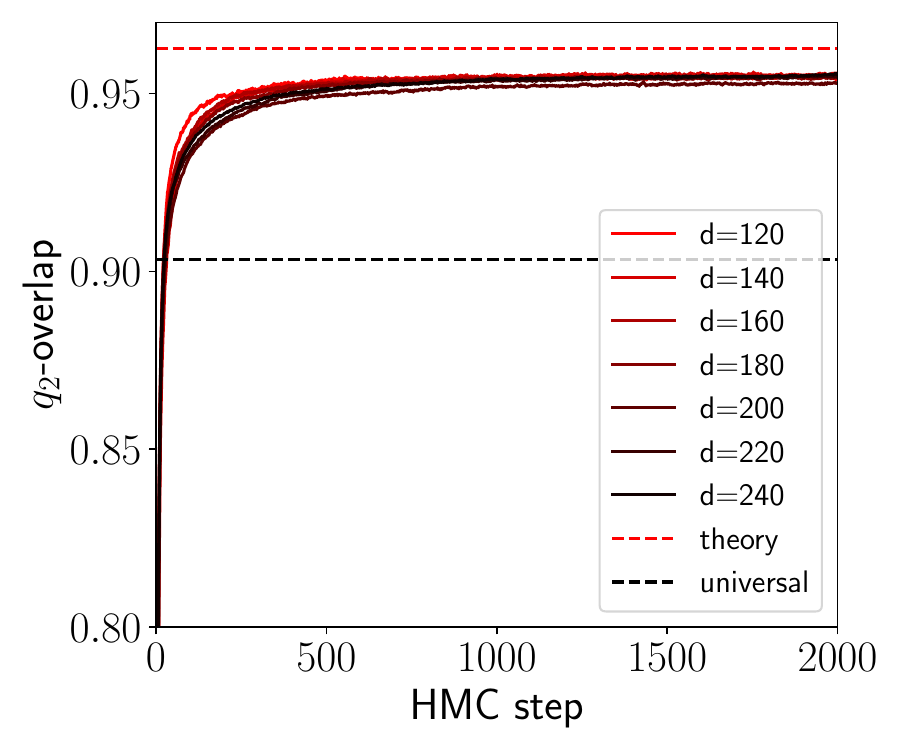}
    \caption{Trajectories of the overlap $q_2$ in HMC runs initialised uninformatively for the polynomial activation $\sigma_3= \He_2/\sqrt 2 + \He_3/6$ with parameters $\Delta = 0.1$ for the linear readout, $\gamma=0.5$ and $\alpha=1.0$. 
    \textbf{Left}: Homogeneous readouts.
    \textbf{Centre}: Rademacher readouts.
    \textbf{Right}: Gaussian readouts.
    Points on the solid lines are obtained by averaging over 10 teacher/data instances, and shaded regions around them correspond to one standard deviation.
    Notice that the $y$-axes are limited for better visualisation. For the left and centre plot, any threshold (horizontal line in the plot) between the prediction of the $ \mathcal{Q}_W \equiv 0$ branch of our theory (black dashed line) and its prediction for the Bayes-optimal $q_2$ (red dashed line) crosses the curves in points $t^*(d)$ more compatible with an exponential fit (see Fig.~\ref{fig:hardness_HMC} and Table~\ref{tab:HMC}, where these fits are reported and $\chi^2$-tested). For the cases of homogeneous and Rademacher readouts, the value of the overlap at which the dynamics slows down (predicted by the $\mathcal{Q}_W\equiv 0$ branch) is in quantitative agreement with the theoretical predictions (lower dashed line). The theory is instead off by $\approx 1\%$ for the values $q_2$ at which the runs ultimately converge.
    }
    \label{fig:hardness_HMC_runs}
\end{figure}

The experiment is performed for the polynomial activation $\sigma_3 = \He_2/\sqrt 2 + \He_3/6$ with parameters $\Delta = 0.1$ for the Gaussian noise in the linear readout, $\gamma=0.5$ and $\alpha=1.0 >\alpha_{\rm sp}$ ($=0.26,0.30,0.02$ for homogeneous, Rademacher and Gaussian readouts respectively). Our HMC consists of $4000$ iterations for homogeneous readouts, or $2000$ iterations for Rademacher and Gaussian readouts. Each iteration is adaptive (with initial step size of $0.01$) and uses $10$ leapfrog steps. Instead of measuring the Gibbs error, whose relationship with $\varepsilon^{\rm opt}$ holds only at equilibrium (see the last remark in App.~\ref{app:gen_err}), we measured the teacher-student $q_2$-overlap which is meaningful at any HMC step and is informative about the learning. For a fixed threshold $q_2^*$ and dimension $d$, we measure $t^*(d)$ as the number of HMC iterations needed for the $q_2$-overlap between the HMC sample (obtained from uninformative initialisation) and the teacher weights $\bW^0$ to cross the threshold. This criterion is again enough to assess that the student outperforms the universal solution.

\begin{figure}[pt]
\begin{center}
\centerline{
\includegraphics[width=.49\linewidth,trim={0 0 0 0},clip]{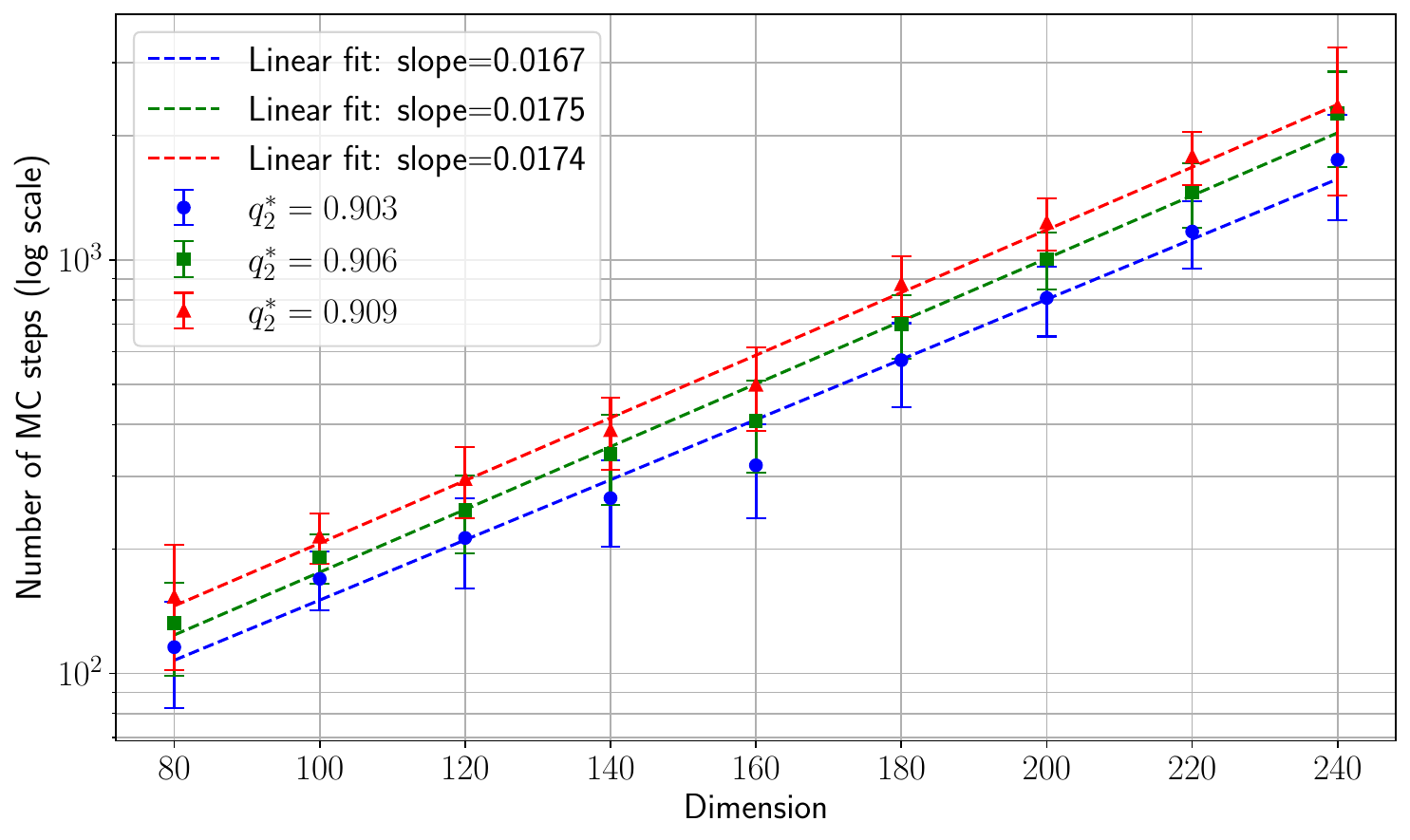}
\includegraphics[width=.49\linewidth,trim={0 0 0 0},clip]{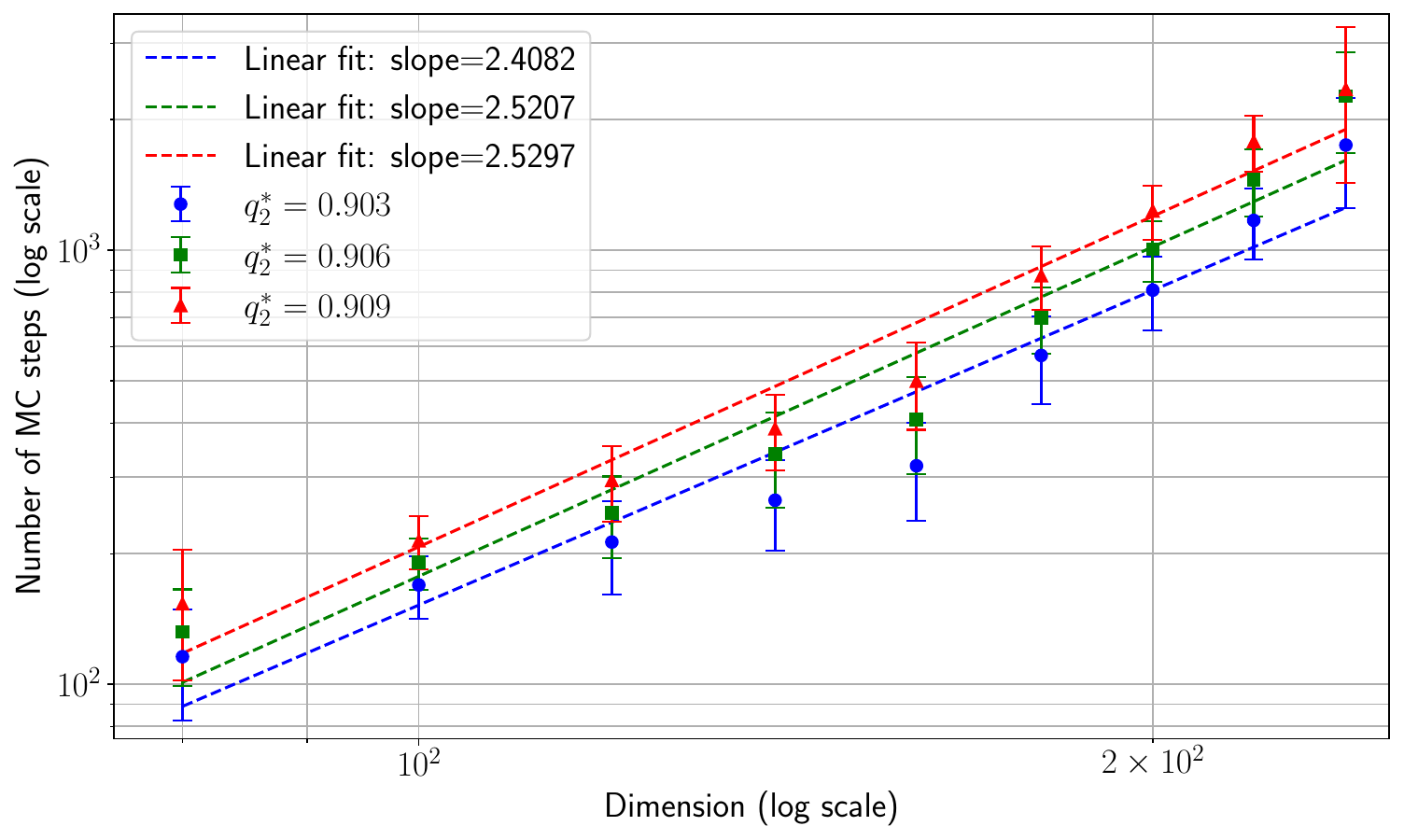}
}
\vspace{1pt} 
\centerline{
\includegraphics[width=.49\linewidth,trim={0 0 0 0},clip]{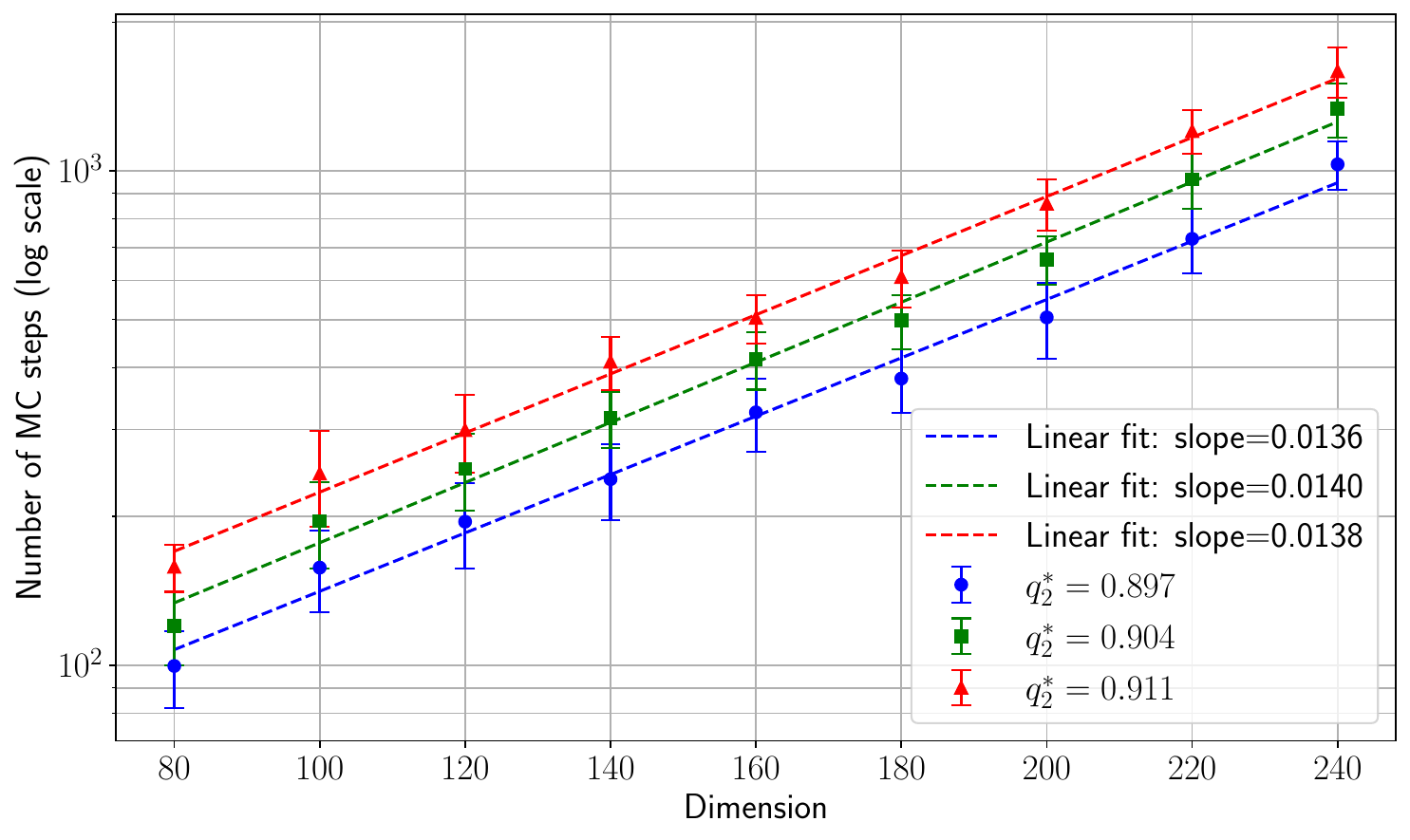}
\includegraphics[width=.49\linewidth,trim={0 0 0 0},clip]{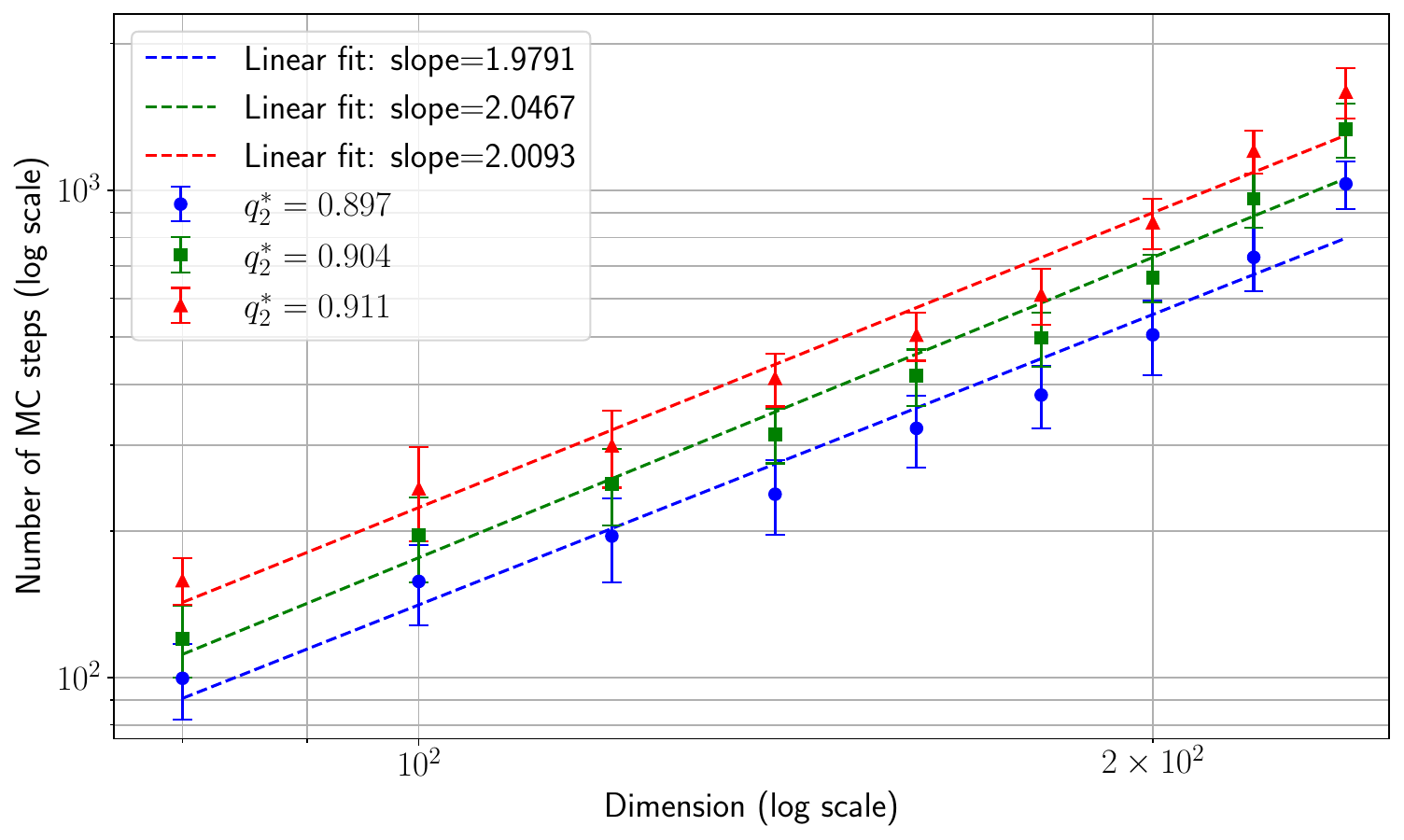}
}
\vspace{1pt} 
\centerline{
\includegraphics[width=.49\linewidth,trim={0 0 0 0},clip]{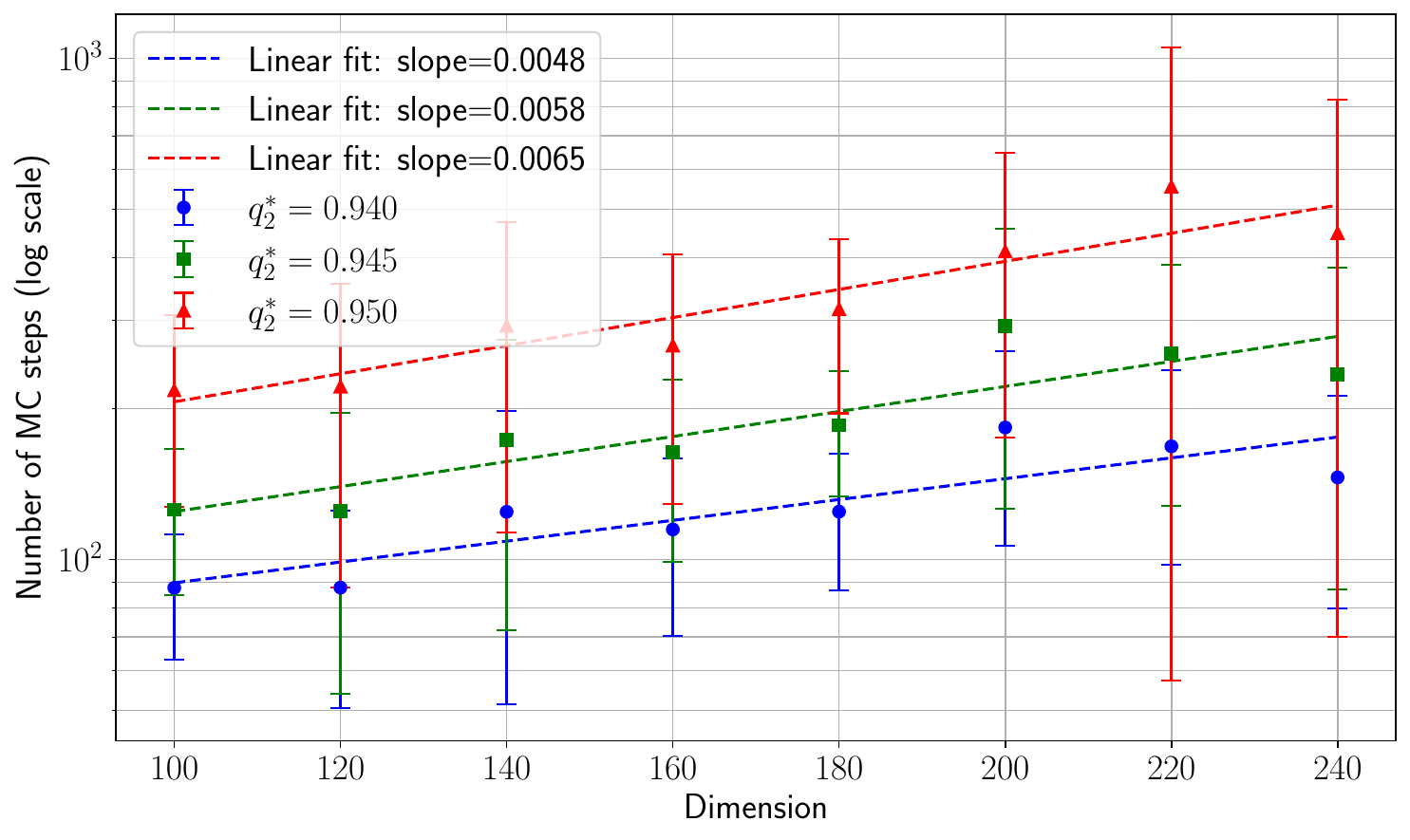}
\includegraphics[width=.49\linewidth,trim={0 0 0 0},clip]{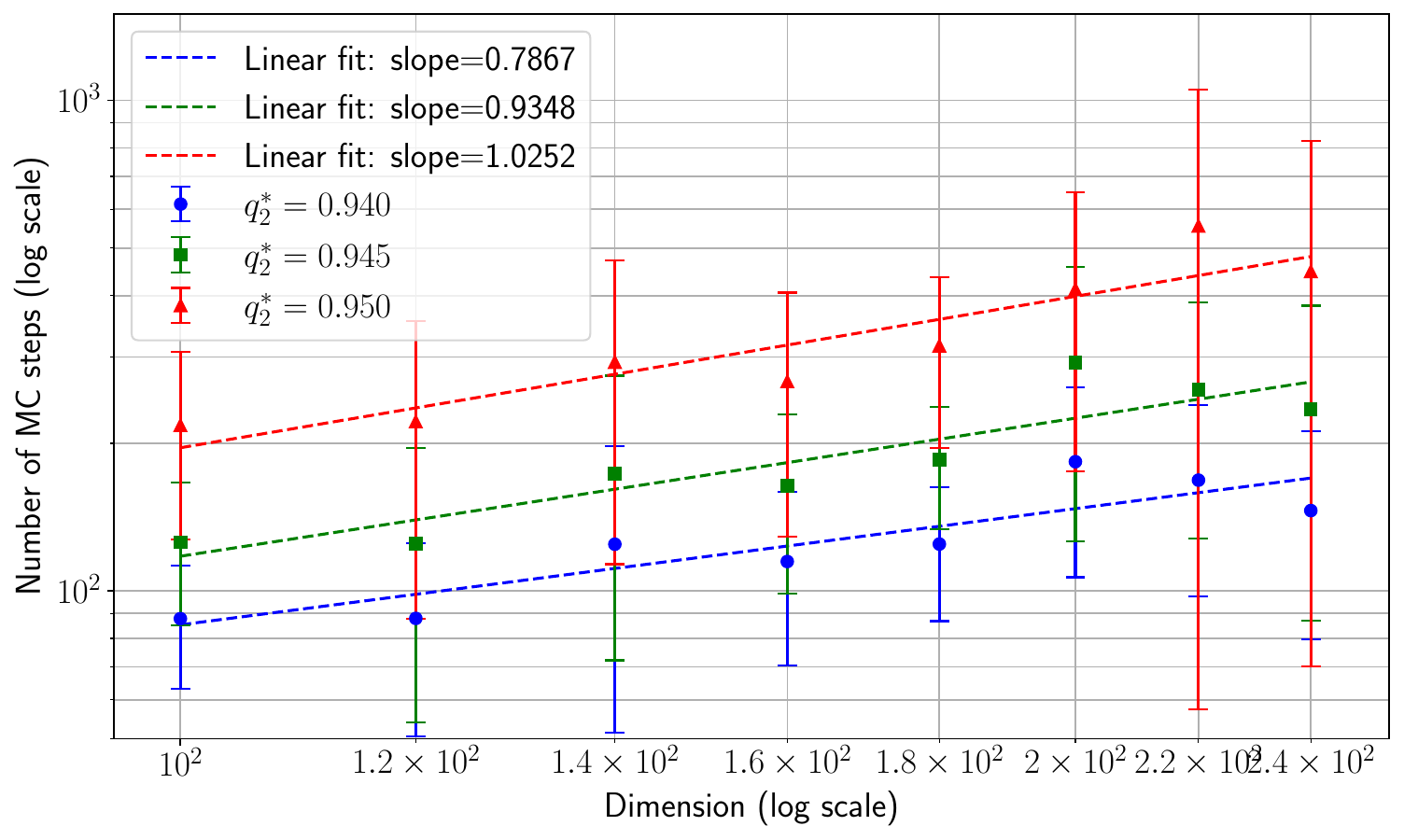}
}
\vspace{-10pt}
    \caption{
    Semilog (\textbf{Left}) and log-log (\textbf{Right}) plots of the number of Hamiltonian Monte Carlo steps needed to achieve an overlap $q_2^*>q_2^{\rm uni}$, that certifies the universal solution is outperformed. 
    The dataset was generated from a teacher with polynomial activation $\sigma_3 = \He_2/\sqrt 2 + \He_3/6$ and parameters $\Delta = 0.1$ for the linear readout, $\gamma=0.5$ and $\alpha=1.0>\alpha_{\rm sp}$ ($=0.26,0.30,0.02$ for homogeneous, Rademacher and Gaussian readouts respectively).
    Student weights are sampled using HMC (initialised uninformatively) with $4000$ iterations for homogeneous readouts (\textbf{Top row}, for which $q_2^{\rm uni}=0.883$), or $2000$ iterations for Rademacher (\textbf{Centre row}, with $q_2^{\rm uni}=0.868$) and Gaussian readouts (\textbf{Bottom row}, for which $q_2^{\rm uni}=0.903$). Each iteration is adaptative (with initial step size of $0.01$) and uses $10$ leapfrog steps. $q_2^{\rm sp}=0.941, 0.948, 0.963$ in the three cases. The readouts are kept fixed during training. 
    Points are obtained averaging over 10 teacher/data instances with error bars representing the standard deviation.
    }
    \label{fig:hardness_HMC}
\end{center}
\vskip -0.3in
\end{figure}

As before, we test homogeneous, Rademacher and Gaussian readouts, getting to the same conclusions: while for homogeneous and Rademacher readouts exponential time is more compatible with the observations, the experiments remain inconclusive for Gaussian readouts (see Fig.~\ref{fig:hardness_HMC}). We report in Fig.~\ref{fig:hardness_HMC_runs} the values of the overlap $q_2$ measured along the HMC runs for different dimensions. {Note that, with HMC steps, all $q_2$ curves saturate to a value that is off by $\approx 1\%$ w.r.t.\ that predicted by our theory for the selected values of $\alpha,\gamma$ and $\Delta$. Whether this is a finite size effect, or an effect not taken into account by the current theory is an interesting question requiring further investigation, see App.~\ref{app:structured_S2} for possible directions.}

\begin{table}[pb]
    \centering
    \begin{tabular}{lc|c|c|c|c|c|c|}
            &        & \multicolumn{3}{c|}{$\chi^2$ exponential fit} & \multicolumn{3}{c|}{$\chi^2$ power law fit}\\
        Readouts    & & \multicolumn{3}{c|}{} & \multicolumn{3}{c|}{} \\
        \hline
         Homogeneous & ($q_2^*\in \{ 0.903, 0.906, 0.909\}$) &   $\bm{2.22}$ & $\bm{1.47}$ & $\bm{1.14}$ &$8.01$ &$7.25$ &$6.35$  \\
         Rademacher& ($q_2^*\in \{0.897 ,0.904 ,0.911 \}$)& $\bm{1.88}$ & $\bm{2.12}$ & $\bm{1.70}$ &$8.10$ &$7.70$ &$8.57$ \\
         Gaussian& ($q_2^*\in \{0.940 ,0.945 ,0.950 \}$)& $0.66$&$\bm{0.44}$&$\bm{0.26}$  & $\bm{0.62}$&$0.53$ &$0.39$\\
    \end{tabular}
    \caption{$\chi^2$ test for exponential and power-law fits for the time needed by Hamiltonian Monte Carlo to reach the thresholds $q_2^*$, for various priors on the readouts. For a given row, we report three values of the $\chi^2$ test per hypothesis, corresponding with the thresholds $q_2^*$ on the left, in the order given. Fits are displayed in \figurename~\ref{fig:hardness_HMC}. Smaller values of $\chi^2$ (in bold, for given threshold and readouts) indicate a better compatibility with the hypothesis.}
    \label{tab:HMC}
\end{table}

\end{document}